\documentclass[a4paper, 11pt]{article}
\usepackage[left=1in,right=1in, top=2.0cm,bottom=2.0cm]{geometry}
\usepackage{amsmath, amsfonts, amssymb, amsthm}
\usepackage{multicol, multirow}
\usepackage{graphicx}
\usepackage{enumerate}
\usepackage[utf8]{inputenc}
\usepackage[inline,shortlabels]{enumitem}
\usepackage{dsfont}
\usepackage{authblk}
\usepackage[usenames,dvipsnames]{color}
\usepackage[colorlinks=true,linkcolor=blue, citecolor=blue, urlcolor=blue]{hyperref}
\usepackage[table, dvipsnames]{xcolor} 
\usepackage{epsfig, epstopdf}
\usepackage{caption}
\usepackage[textfont=footnotesize,justification=centering]{subcaption}
\usepackage{appendix}
\usepackage{float}
\usepackage{comment}
\usepackage[multiple]{footmisc}
\usepackage[sort]{natbib}
\usepackage{hyperref}
\usepackage{stackrel}
\usepackage[normalem]{ulem}
\usepackage{csquotes}
\usepackage{bm}
\usepackage{booktabs}
\usepackage{tabularx}
\usepackage{ltablex}
\newtheorem*{theorem-non}{Theorem}
\numberwithin{equation}{section}
\theoremstyle{definition}

\newcommand{\blueparagraph}[1]{{\color{blue}#1}}

\newcommand{\rBrackets}[1]{\left( #1 \right)}
\newcommand{\sBrackets}[1]{\left[ #1 \right]}
\newcommand{\cBrackets}[1]{\left\{ #1 \right\}}
\newcommand{\xbold}{\bm{x}}
\newcommand{\abold}{\bm{a}}
\newcommand{\Ybold}{\bm{Y}}
\newcommand{\abbar}{\overline{\bm{a}}}

\newcommand{\Xbold}{\bm{X}}
\newcommand{\sbold}{\bm{s}}

\newcommand{\betab}{\bm{\beta}}
\newcommand{\thetab}{\bm{\theta}}
\newcommand{\betabhat}{\widehat{\bm{\beta}}}
\newcommand{\betabstar}{\bm{\beta}^{\ast}}
\newcommand{\ASet}{\mathcal{A}}
\newcommand{\SSet}{\mathcal{S}}

\newcommand{\SSettilde}{\mathcal{S}}
\newcommand{\fixparen}[1]{\left( #1 \right)}

\title{Synthetic data for ratemaking: imputation-based methods \\ vs adversarial networks and autoencoders}

\author[1,\thanks{\href{yevhen.havrylenko@unil.ch}{yevhen.havrylenko@unil.ch}, corresponding author}]{Yevhen Havrylenko}
\author[2,\thanks{\href{meelis.kaarik@ut.ee}{meelis.kaarik@ut.ee}}]{Meelis K\"{a}\"{ a}rik}
\author[2,\thanks{\href{artur.tuttar@ut.ee}{artur.tuttar@ut.ee}}]{Artur Tuttar}

\affil[1]{Department of Actuarial Science, Faculty of Business and Economics, University of Lausanne, Switzerland}
\affil[2]{Institute of Mathematics and Statistics, Faculty of Science and Technology, University of Tartu, Estonia}

\allowdisplaybreaks

\date{}

\begin{document}
\maketitle

\begin{abstract}
Actuarial ratemaking depends on high-quality data, yet access to such data is often limited by the cost of obtaining new data, privacy concerns, etc. In this paper, we explore synthetic-data generation as a potential solution to these issues. In addition to generative methods previously studied in the actuarial literature, we explore and benchmark another class of approaches based on Multivariate Imputation by Chained Equations (MICE). In a comparative study using an open-source dataset, MICE-based models are evaluated against other generative models, such as Variational Autoencoders and Conditional Tabular Generative Adversarial Networks. We assess how well synthetic data preserves the original marginal distributions of variables as well as the multivariate relationships among covariates. The consistency between Generalized Linear Models (GLMs) trained on synthetic data with GLMs trained on the original data is also investigated. Furthermore, we assess the ease of use of each generative approach and study the impact of generically augmenting original data with synthetic data on the { estimation of GLMs for predicting claim counts.} Our results highlight the potential of MICE-based methods in creating high-fidelity tabular data while offering lower implementation complexity compared to deep generative models.

\end{abstract}

{\textbf{\\Keywords}: synthetic actuarial data, data augmentation, multivariate imputation by chained equations, generative adversarial networks, autoencoders}

\section{Introduction}\label{sec:introduction}
\textbf{Motivation.} High-quality actuarial data is essential for actuaries in various business lines of insurance and reinsurance companies. It helps quantify risks more accurately, and thus maintain the competitiveness of actuarial pricing and reserving models. Furthermore, researchers in actuarial science need realistic data to test their novel methodologies and compare them with existing ones. Even if insurers and reinsurers may have enough data for some products or business lines, the companies rarely publicly share their data due to concerns related to data privacy, security, and their competitive advantage. As a result, the number of open-access actuarial datasets is quite small (see, e.g., \cite{Gabrielli2018}, \cite{Cote2020}, \cite{SeguraGisbert2024}). Furthermore, the companies themselves may also face a lack of data, e.g., when they enter a new market or release a new (re-)insurance product.

{To increase the amount of actuarial data, one can generate synthetic data, which can be used either as a standalone dataset or as an enlargement of an existing dataset. In the former case, synthetic data can be publicly shared for educational and research purposes (e.g., actuarial training, validation and benchmarking of novel models and algorithms), provided that it preserves the relevant statistical properties of the original data and does not expose sensitive information. Another example of using standalone synthetic data arises when an insurance company shares synthetic data with a software developer to demonstrate the functionality and utility of a software solution before deciding whether to purchase it.

As mentioned above, synthetic data can also be used to enlarge an existing real dataset, a process commonly referred to as data augmentation. The resulting augmented dataset may be more diverse and/or more balanced, which can be beneficial for downstream predictive modeling. For example, to improve the performance of classification models (e.g., fraud-detection models), one can generate additional synthetic observations from an underrepresented class, as is done in techniques such as SMOTE. The topic of data augmentation for improving the performance of regression models on imbalanced structured datasets is relatively underexplored, according to the comprehensive survey \cite{stocksieker2025survey}. For datasets with imbalanced continuous responses (e.g., an insurance claim amount),  \cite{stocksieker2024dataaugwithvae} proposes a targeted augmentation method that preferentially generates observations in sparsely represented regions of the response distribution. Since the actuarial literature on insurance pricing and reserving contains relatively little evidence regarding the benefits of data augmentation, we include in our case study a small exploratory experiment that investigates how increasing the proportion of synthetic observations in an augmented dataset affects the estimation of coefficients in a GLM fitted to that dataset. This generic augmentation experiment should be viewed as a first, deliberately simple step toward a broader investigation of data augmentation in actuarial ratemaking. A natural next step, which goes beyond the scope of this article, would be to study the impact of targeted augmentation in an insurance context, particularly for non-continuous response variables such as claim counts.}

{
In this paper, we focus on synthetic data generation methods that require minimal distributional assumptions and are suitable for tabular datasets with both numerical as well as categorical variables and where temporal independence among data points can be assumed. Methods of this kind} gained much popularity due to the advancement of deep generative models such as generative adversarial networks (GANs) and variational autoencoders (VAEs). To our knowledge, only three papers investigated the use of such models in the context of ratemaking: \cite{Kuo2020},  \cite{Jamotton2024}, and \cite{Cote2020}. We discuss these articles in more detail in the literature overview below and mention for now only aspects that strongly motivate us to conduct further research on this topic. First, in our experience, the data generation approaches studied in those three papers seem to require quite some customization for each input dataset, which is why their ease of use in the daily work of pricing actuaries seems to be limited. Second, {those papers do not examine the effect of adding increasing amounts of synthetic data to the original training data}. 

\textbf{Literature overview.}
\cite{Kuo2020} pioneered the use of conditional tabular GANs for generating actuarial datasets, focusing on ratemaking and lapse modeling. In his experiments with the \textit{freMTPL2freq} dataset, the researcher showed that \textbf{g}eneralized \textbf{l}inear \textbf{m}odels (GLMs) trained on synthetic data achieved performance comparable to GLMs trained on real data, and the distributions of both continuous and categorical variables in the synthetic datasets closely mirrored the real datasets. However, the relative effects (how much each level of a categorical variable affects the predicted outcome relative to a reference level) of GLM models constructed using synthetic datasets do not show the same patterns as those of models constructed using real datasets. Because of the observed instability in the parameters of a GLM trained on synthetic data, developing a rating plan based on it might not be the best approach.

Inspired by \cite{Kuo2020}, \cite{Cote2020} investigate three GAN-based models for generating synthetic insurance data:  Multi-Categorical Wasserstein GAN with Gradient Penalty (MC-WGAN-GP), Conditional Tabular GAN (CTGAN), and Mixed Numerical and Categorical Differentially Private GAN (MNCDP-GAN). In a case study using the \textit{freMTPLfreq} data, they find that MC-WGAN-GP performs best in replicating univariate distributions and the observed claim frequency for each category of categorical variables, followed by CTGAN, and then MNCDP-GAN. In assessing the consistency of GLMs fitted on real versus synthetic data, CTGAN-based GLMs most closely resemble the real-data-based GLM in terms of average estimated GLM coefficients. However, MC-WGAN-GP based GLM predictions are closest to the real-data-based GLM predictions when using Mean Absolute Error and Mean Squared Error as evaluation metrics. The authors conclude that neither of the tested models performs well across all key performance indicators (KPIs) with a high level of training data confidentiality.

As an alternative to GAN-based models, \cite{Jamotton2024} propose a Variational Autoencoder (VAE) model for the generation of insurance data. Their approach features a quantile transformation for continuous variables to address multimodality. In their VAE, the researchers use a reconstruction loss function that consists of two parts:  cross-entropy for categorical variables and MSE for continuous variables. In a case study using a dataset from the homepage of the book \cite{ohlsson2010non}, the researchers find that the marginal distributions of the real datasets and those of the synthetic dataset are almost the same, as supported by visual inspection, Kolmogorov-Smirnov tests for continuous variables, and Chi-Squared tests for categorical variables. Based on Pearson and Spearman correlations for continuous variables and contingency scores based on contingency tables for categorical variables, the authors conclude that synthetic data preserves pairwise relationships among variables reasonably well. Training a GLM on synthetic data generated by their VAE and noticing that the respective Poisson deviance is almost the same as the one for a GLM trained on the original data, the scientists conclude that synthetic data can substitute real data in actuarial analyses.

Fundamentally different data generation methods are those based on multivariate imputation of data. While less explored in actuarial science, these methods are well-established in statistical disclosure control. The conceptual foundation was laid by \cite{Rubin1993}, who proposed treating the creation of synthetic data as a missing data problem: synthetic values are drawn from the posterior predictive distribution of the original data. To obtain the predictive distribution of the original data, it is common to use one of the two approaches: Sequential Regression (SR) or Multivariate Imputation by Chained Equations (MICE).

The SR approach approximates the joint multivariate distribution of data by explicitly factorizing it into a strictly ordered chain of univariate conditional densities (i.e., 1st variable, then 2nd variable given the 1st variable, and so on). 
\cite{Reiter2005} demonstrates that classification and regression trees (CART) can be effectively integrated into the SR framework. Building on that paper, \cite{Caiola2010randomforests} show that using Random Forests (RFs) as imputation models has multiple benefits for creating synthetic datasets using SR: the synthesizer requires minimal fine-tuning, naturally handles variables of any type, scales well to high-dimensional data, and automatically captures non-linear relationships. This synthetic data generation methodology is implemented in the \textsf{R}-package \textit{synthpop}, see \cite{Nowok2016synthpop}. SR approach reduces the synthesis task to a series of unidirectional regression steps according to a user-defined \enquote{synthesis order} (\textit{visit.sequence} parameter in \textit{synthpop} package), e.g., first \textit{Age}, then \textit{Income} given \textit{Age}, then \textit{Family Status} given \textit{Income} and \textit{Age}, and so on). As mentioned in \cite{Caiola2010randomforests} and \cite{drechsler2011empirical}, there is no mathematical theory underpinning the ordering of variables for generation, so different choices of the sequence could produce synthetic datasets of different quality. Once the user fixes the synthesis sequence, earlier variables can influence the generation of the later ones, but not the other way around, which is why the dependence structure among synthetically generated variables may not be close enough to the true dependence structure and may lead to biased estimation of multivariate relationships \citep{Reiter2005}. Since ratemaking data can contain variables with endogenous, bidirectional dependencies (e.g., the interplay between coverage limits, deductibles, and vehicle value) that lack a natural synthesis order, we do not explore this method in our paper and focus on MICE, {which does not require a fixed synthesis order.}

Developed at the end of the nineties, MICE operates by modeling the conditional density of each variable given all other variables, which is why it is also called the Fully Conditional Specification approach. This is an important difference to the SR methodology in which every single variable is imputed/generated conditional on a model trained on only the variables located before the current variable in the visiting sequence. Similarly to SR, MICE handles diverse variable types without explicitly defining a high-dimensional joint distribution, c.f., \cite{Drechsler2011}, but it offers more flexibility and is easier to use in practice because it does not require a fixed ordering of variables for data generation. {As outlined in Subsection 4.5.2 of \cite{vanBuuren2018}, MICE can be conceptually interpreted as a Markov Chain Monte Carlo method. By cyclically drawing from univariate conditional distributions across multiple iterations, the underlying Markov chain aims to stabilize, thereby yielding imputations that are empirically useful and may reflect the multidimensional dependence structure even when a compatible joint distribution is not theoretically guaranteed.} This approach to data imputation is implemented in the \textsf{R}-package \textit{mice}, see \cite{vanBuuren2011}. As with SR, MICE can also be used for synthetic data generation, which is demonstrated by \cite{Volker2021} for a medicine-related dataset with $9$ variables and about $750$ observations. The researchers highlighted that MICE with CART as the imputation model produces indistinguishable synthetic records that support unbiased, valid statistical inferences, and offer a user-friendly solution for securely sharing and analyzing sensitive data. Although SR and MICE methodologies for synthetic data generation have been gaining popularity and credibility in the medical and social sciences, they seem to have been largely overlooked by the actuarial community in favor of deep generative models described at the beginning of our literature overview.

A related idea of generating artificial data using imputation appeared in \cite{Neves2022}, where the authors propose a so-called \textit{tabulator} framework. In their approach, they propose to randomly (not on variable-by-variable basis) select data entries to be set to missing and then be imputed by a so-called back-end engine for imputation. As a model driving this imputation engine, the authors propose to use one of the three methods that can be seen as modifications of Generative Adversarial Imputation Networks (GAINs, originally proposed by \cite{Yoon2018}): Slim GAIN (SGAIN), Wasserstein Slim GAIN with Clipping Penalty (WSGAIN-CP), Wasserstein Slim GAIN with Gradient Penalty (WSGAIN-GP). According to the researchers, \textit{tabulator} has a comparable performance (tested on 10 datasets from various fields like healthcare, economics, and computer science) with CTGAN that, in turn, is better than medical Generative Adversarial Network (medGAN), Variational Encoder Enhanced Generative Adversarial Network (VEEGAN), and Table Generative Adversarial Network (table-GAN). The researchers conclude that the \textit{tabulator} preserves the dependence structure of the original dataset better than CTGAN, but both approaches struggle when generating synthetic data from original imbalanced data. It is also stated that synthetic data from \textit{tabulator} improves the performance of regression/classification tasks only marginally. 

Recent comprehensive reviews of the literature on synthetic data generation {discuss evidence that for tabular data deep generative models do not consistently outperform traditional synthesis methods, in particular sequential conditional approaches based on CART}. \cite{Reiter2023survey} observes that statistical agencies, such as the U.S. Census Bureau, continue to rely on sequential conditional modeling for major data products, citing CART as a \enquote{primary engine} or \enquote{leading contender} for synthesis tasks. Furthermore, \cite{Drechsler2024survey} report that, despite the great success of GANs in computer science, {the comparative evaluations reviewed therein found} that sequential regression based on CART frequently offers the highest utility but also the highest disclosure risk.

Readers interested in the literature on insurance data generation beyond traditional ratemaking are referred to \cite{So2021} and \cite{So2022} for telematics data, to \cite{Gabrielli2018}, \cite{Avanzi2021}, \cite{Avanzi2023}, \cite{Bauwelinckx2024} for reserving data, to \cite{Campo2023} for auto-insurance fraud-detection data, and to Chapter 10 in \cite{Wuethrich2025} for a brief introduction to generative modeling tools for actuaries. We refer readers to \cite{Fonseca2023} and \cite{Brophy2023} for a systematic review of the literature on the GAN-based generation of tabular data and on the GAN-based generation of time series data, respectively. For an overview comparing deep generative models, including but not limited to VAEs and GANs, we direct readers to \cite{BondTaylor2022}.

\textbf{Our contributions.} 
First, we benchmark the utility and practicality of several amputation-imputation-based approaches to data generation for actuarial ratemaking. These techniques are based on Multivariate Imputation by Chained-Equations (MICE) with Random Forests (RFs) as imputation models. According to our tests, these approaches can be more suitable for \enquote{out-of-the-box} usage in practice than the generative models studied in \cite{Kuo2020}, \cite{Cote2020}, \cite{Jamotton2024}, while having a comparable performance. 

Second, using a publicly available French Motor Third-Part Liability dataset called \textit{freMTPL2freq} \citep{Dutang2019casdatasets} as a training dataset, we compare $10$ approaches to ratemaking data generation, including the existing approaches \citep{Kuo2020, Cote2020, Jamotton2024}, their modifications that address the challenge of GANs in dealing with high-cardinality categorical variables, and the newly introduced amputation-imputation-based methods. We analyze the quality of synthetic datasets by studying two aspects: the similarity between the distribution of the original data and that of synthetic data, the ability of methods to capture structural dependence among covariates and the response variable. {In addition, we examine how well the relationships among variables are preserved as an increasing proportion of synthetic observations is added to the original training data. As mentioned earlier, the augmentation experiment represents only one possible application of synthetic data generation and should not be interpreted as the study's primary objective.}

Our main results are threefold. First, we illustrate that the MICE-RF  approach is a competitive method for generation of synthetic ratemaking data in comparison to deep generative models such as a VAE and a CTGAN. The important practical advantage of the MICE-RF algorithm is its ease of use, given the existing packages in $\textsf{R}$ and the versatility of RFs in dealing with variables of different types without much pre-processing. Second, we show that modifying a CTGAN with an AE for categorical-variable pre-processing improves the quality of synthetic high-cardinality categorical variables but deteriorates the performance of this hybrid approach in terms of other criteria.
In general, MICE-RF-based approaches are the best-performing in terms of most performance metrics we consider in our experiments. Third, our case studies indicate that for our experimental datasets based on the \textit{freMTPL2freq} {and Swedish motorcycle data, generic data augmentation does not improve the estimation of GLM coefficients.}

\textbf{Structure of the paper.} In Section \ref{sec:general_setup}, we present the general setup for tabular data generation in the context of ratemaking. In Section \ref{sec:methodology}, we describe the main components of data generation approaches compared in this paper. Section \ref{sec:case_studies} is devoted to our comparative study. Section \ref{sec:conclusion} concludes the paper. Appendix \ref{app:summary_AE} contains additional information on the hybrid approach combining CTGAN with AEs. {In Appendix \ref{app:summary_metrics}, we provide an overview of all metrics that we use in the tables comparing all tested data generation methods. Appendix \ref{app:comparison_of_marginals} contains plots that indicate how well synthetic data preserves the claim frequency distribution across various categories/bins of variables in comparison to the respective claim frequencies in the original dataset. Finally, in Appendix \ref{app:case_study_2}, we summarize our findings from the second case study.}

\section{General setup}\label{sec:general_setup}
First, we introduce notation and briefly discuss some data pre-processing techniques commonly used before fitting/training any statistical learning models. Second, we summarize the basics of GLMs for predicting insurance claim frequencies. Third, we explain what is meant by generating synthetic data and state two research questions that are answered in the subsequent sections.

We denote a tabular dataset by $\mathcal{S} :=\cBrackets{\sbold_i}_{i = 1,\dots, n_{ROD}}$, where $n_{ROD}$ is the number of \textbf{r}ows in the \textbf{o}riginal \textbf{d}ata (number of observations), $\sbold_i = (\xbold_i, v_i, y_i)$ is a tuple with three elements such that $\xbold_i = (x_{i;1}, \dots, x_{i;j}, \dots, x_{i;n_C})^\top$ is a $n_C$-dimensional vector of observed values of covariates ({predictors}), $n_C \in \mathbb{N}$, $^\top$ is the transpose operator, $v_i \in (0, 1]$ is an exposure-to-risk (proportion of a year during which the observation $i$ was tracked) and $y_i$ is an observed value of a one-dimensional response variable (dependent variable) for an observation $i = 1,\dots, n_{ROD}$.
A tuple $\sbold_i$ can be interpreted as an insurance policy snippet.
For simplicity, we may omit the index $i$ indicating the ordinal number of the observation, when it is not essential for interpretation, and just use $(\xbold, v, y)$. The vector $\xbold$ can have both categorical and numerical components. We denote the number of \textbf{c}ategorical \textbf{c}ovariates in $\xbold$ by $n_{CC}$ and the number of \textbf{n}umerical \textbf{c}ovariates in $\xbold$ by $n_{NC} = n_C - n_{CC}$. 

Before fitting/training any model on the raw data, it is usually pre-processed to a suitable format, depending on the model to be used later. After pre-processing, a variable can be represented by multiple elements, which are referred to as features. An example of pre-processing of a categorical variable is one-hot encoding. Consider a categorical variable $x_j$, $j = 1, \dots, n_{CC}$, which has $m_j$ different categories (labels) $\cBrackets{a_1^{j},\dots, a_{m_j}^{j}}$. The one-hot encoding of this variable means representing it as an $m_j$-dimensional vector containing only $0$ and $1$ as follows:
\begin{equation}\label{eq:one-hot-encoding}
    x_j \mapsto \xbold_j^{cat} = \rBrackets{x_{j_1}, \dots, x_{j_{m_j}}}^{\top} = \rBrackets{\mathds{1}_{\cBrackets{x_{j} = a_1^{j}}}, \dots, \mathds{1}_{\cBrackets{x_{j} = a_{m_j}^{j}}}}^{\top} \in \cBrackets{0, 1}^{m_j},
\end{equation}
where $\mathds{1}_{\cBrackets{\cdot}}$ is an indicator function that is equal to $1$ if the condition in curly brackets is true and $0$ otherwise.  One-hot encoding is often used (implicitly or explicitly) before training advanced statistical learning models, including those mentioned in the next section. Another way of encoding categorical variables is dummy encoding, which yields a similar representation as \eqref{eq:one-hot-encoding}, but one of the categories is omitted. This omitted category (label) is then called a reference one and is represented by an $(m_j-1)$-dimensional vector containing only $0$. Dummy encoding is frequently used for GLMs and GAMs. {It is important to stress that most of these pre-processing steps should not alter the structure of the data such that there is a loss of information, as that could affect the training and performance of the given model.}

Unless specified otherwise, we assume that for data generation algorithms, categorical variables are pre-processed via one-hot encoding and for GLMs via dummy encoding. We denote by $n_{PCF}$ the number of \textbf{p}re-processed \textbf{c}ategorical \textbf{f}eatures. 

Similarly, statistical learning models tend to perform better when numerical variables are also pre-processed, which should be explicitly done by the user before training a model or is implicitly done in a respective package implementing the model. For example, a numerical variable $x_k$ (as well as exposure $v$ and response variable $y$) can be transformed via a min-max scaler to $\widetilde{x}_k \in [-1,1]$, $k = 1, \dots, n_{NC} =: n_{PNF}$, where $PNF$ stands for \textbf{p}re-processed \textbf{n}umerical \textbf{f}eatures.

Thus, without loss of generality (w.l.o.g.) and with a slight abuse of notation, we can write the vector of pre-processed features as follows:
\begin{equation*}
    \xbold = \rBrackets{(\xbold^{cat})^{\top}, (\xbold^{num})^{\top}}^{\top}:=\rBrackets{(\xbold_{1}^{cat})^\top,\dots, (\xbold_{n_{CC}}^{cat})^\top, \widetilde{x}_{1}, \dots, \widetilde{x}_{n_{PNF}}}^{\top} \in \mathbb{R}^{n_{PF}},
\end{equation*}
where $n_{PF} =  n_{PCF} + n_{PNF}$ stands for the number of \textbf{p}re-processed \textbf{f}eatures.

To avoid loaded notation, we still use $\SSet$ to denote the dataset after pre-processing.  
An actuary can use a subset of $\SSet_{train} \subseteq \SSet$ (where subsetting is done among tuples not within their distinct elements) to train a generative model that, once trained, can be used to generate a synthetic dataset $\SSet^{A} = \cBrackets{{\sbold}^{A}_i}_{i = 1, \dots, n_{RSD}}$, where the superscript $A$ refers to the \textbf{a}pproach chosen for generative modeling, $n_{RSD}$ is the number of needed \textbf{r}ows of \textbf{s}ynthetic \textbf{d}ata and ${\sbold}^{A}_i = \rBrackets{{\xbold}^{A}_i, {v}_i^{A}, {y}_i^{A}}$. A good approach $A$ should ensure that the distribution $F_{\sbold}$ of a real ({possibly} pre-processed) data tuple is close to the distribution $F_{{\sbold}^A}$ of a synthetic ({correspondingly} pre-processed) data tuple. This implies that the data generation approach should learn the structural link between $(\xbold , v)$ and $y$. 

In ratemaking, it is common to model the link between $y$ and $(\xbold, v)$ via a generalized linear model (GLM). GLMs were introduced in \cite{Nelder1972} and became very popular in insurance, where they are routinely used in various business lines such as claim management, underwriting, etc. In our paper, we will only discuss the basics of a Poisson GLM, which is frequently used to predict insurance claim counts and which we use in numerical studies. For a deep dive into the theory and practice of GLMs in insurance readers are referred to Chapter 4 of \cite{Denuit2019MLABookVolume1} and Chapter 5 of \cite{Wuethrich2023}. 

Let $Y$ be a random variable representing a response variable (e.g., number of insurance claims per year) and $y$ be its realization (observation). Let $\bm{X}$ be a vector of (random) features in a format suitable for a GLM (e.g., categorical variables are represented via dummy encoding, the first feature always equals $1$ if a GLM shall contain an intercept, etc.) and $\xbold$ be its realization (observation). Let $v \in (0, 1]$ denote a known exposure. Then a Poisson GLM with a so-called log-link function and without interactions among covariates assumes that:
\begin{equation*}
    [Y| \bm{X} = \xbold, v] \sim \text{Poisson}\rBrackets{v \cdot \exp\rBrackets{\betab^\top \xbold}},
\end{equation*}
where $\betab := \rBrackets{\beta_0, \dots, \beta_{n_{PF}}}^\top$ is the vector of GLM parameters. Taking an expectation 
and applying a logarithmic function, we get:
\begin{equation} \label{eq:Poisson_GLM_no_interactions}
    \ln \Bigl(\underbrace{\mathbb{E}[Y| \bm{X} = \xbold, v]}_{=y,\text{ target}} \Bigr)= \underbrace{\ln(v)}_{\text{offset}} + \underbrace{\betab^\top \xbold}_{\text{linear predictor}},
\end{equation}
where $\mathbb{E}[\cdot | \cdot ]$ denotes the conditional expectation operator.

A Poisson GLM specified by \eqref{eq:Poisson_GLM_no_interactions} has no interactions. Depending on the data, it may be beneficial to include interactions in a GLM. For simplicity, let us consider a pairwise interaction. An interaction between two features $x_j$ and $x_k$ (elements of $\xbold$) is modeled as an additional term $I\rBrackets{x_j, x_k}$ in the right-hand side of \eqref{eq:Poisson_GLM_no_interactions} such that $I\rBrackets{x_j, x_k}$ cannot be represented as the sum of two univariate functions, i.e., $I\rBrackets{x_j, x_k} \neq I_1(x_j) + I_2(x_k)$, $j \in \cBrackets{1, \dots,n_{PF}}, k \in \cBrackets{1, \dots,n_{PF}}$, $j \neq k$. A common choice is
\begin{equation}\label{eq:interaction_term}
    I\rBrackets{x_j, x_{k}} = \beta_{j,k} x_j x_k.
\end{equation}
In \eqref{eq:interaction_term}, $x_j$ (and/or $x_k$) can be just a feature representing one category. To consider an interaction with an entire categorical variable, a sum of different \eqref{eq:interaction_term} (one for each category in the respective categorical variable, except for the dummy one) must be added to the right-hand side of \eqref{eq:Poisson_GLM_no_interactions}. For more information on modeling interactions in GLMs, see, e.g., \cite{fox2018r}, or Section 4.2.5 in \cite{Denuit2019MLABookVolume1}.

Let us denote the true value of $\betab$ by $\betabstar$. In practice, $\betabstar$ is not known and must be estimated from collected data. Using a training data $\SSet_{train}$, an actuary can obtain an estimate of $\betabstar$ by a so-called maximum likelihood estimation (MLE), which is equivalent to minimizing Poisson deviance in the case of a Poisson GLM. When comparing different GLM models, the one with a lower Poisson deviance on the respective test set $\SSet_{test}:=\SSet \setminus \SSet_{train}$ is considered a better-performing one. An alternative performance measure is the Akaike Information Criterion (AIC), which takes into account the number of parameters of a GLM and is based only on $\SSet_{train}$.

The predictive performance of a GLM depends on various aspects including but not limited to the choice of covariates used in a GLM and the quality as well as quantity of the observed data. Addressing the former aspect, an actuary should manually select explanatory variables (elements of $\xbold$) or obtain the list of important variables for a GLM using so-called variable-selection algorithms, such as backward, forward, or mixed variable selection (VS) methods {(see, e.g., Chapter 15 in \cite{draper1998} or Section 13.2.2 in \cite{fox2016})}. Let us denote by $\betabhat := \betabhat\rBrackets{\SSet_{train}}$ the estimate of $\betabstar$ when an actuary correctly specifies all variables with significant effects in a GLM, i.e., when GLM is trained with all elements of $\xbold$ that correspond to non-zero elements of $\betabstar$. Let  $\widehat{\betabhat\,} := \widehat{\betabhat\,}\!\rBrackets{\SSet_{train}}$ denote the estimate of $\betabstar$ when an actuary uses some VS method.

To address the aspect of \textit{data quantity} and \textit{quality}, an actuary can generate a synthetic dataset $\SSettilde^{A}$ of a desirable \textit{size} and reasonable \textit{quality}. In this situation, two natural questions arise: one is related to the usage of synthetic data as a standalone set, and the other is related to working with the original dataset that is augmented with a part of the synthetic dataset. First, how well does $\SSettilde^{A}$ preserve (from training data $\SSettilde_{train}$) the marginal distributions of variables as well as the multivariate relationships among covariates, and how consistent is a GLM trained on $\SSettilde^{A}$ with a GLM fitted on $\SSettilde_{train}$? {Second, how does training GLMs on an augmented dataset $\SSettilde_{train} \cup \SSettilde^{A}$ affect the estimation of $\betabstar$; more specifically, are $\betabhat(\SSettilde_{train} \cup \SSettilde^{A})$ and $\widehat{\betabhat\,}\!(\SSettilde_{train} \cup \SSettilde^{A})$ closer to the true $\betabstar$ than the corresponding estimates based only on $\SSettilde_{train}$?} These are the two main research questions that we aim to answer in this paper.

\section{Methodology}\label{sec:methodology}
In this section, we give an overview of the main building blocks of different synthetic data generators explored in this paper. First, we start with the imputation framework MICE and explain how it can be used for synthetic data generation. Second, we briefly explain a CTGAN model. Third, we summarize deterministic and variational autoencoders (AEs).

\subsection{Multivariate imputation by chained equations}\label{subsec:MICE_theory}
According to \cite{vanBuuren2018}, multivariate imputation by chained equations (MICE) is one of the best approaches to dealing with missing data. MICE is also known as fully conditional specification, sequential regression multivariate imputation, or simply chained equations.

MICE commonly operates under the assumption that data is missing at random, which means that the probability of a value being missing depends only on observed values and not on unobserved values. Avoiding explicit assumptions on the joint distribution of the data, MICE works by specifying a set of conditional distributions for each variable with missing data, given the other variables in the data. The algorithm then iteratively imputes the missing values by drawing from these conditional distributions. {So, 
conceptually, MICE algorithm is a Markov Chain Monte Carlo (MCMC) method and is sometimes called \enquote{ordered pseudo-Gibbs sampler}, see, e.g., \cite{gelman2004}. If the conditional densities are so-called compatible, then it is guaranteed that MICE method is a valid Gibbs sampler.  When the compatibility condition is not satisfied, MICE still produces sensible imputations, but there is no theoretical guarantee that the resulting imputed dataset corresponds to a mathematically valid joint distribution. According to \cite{gelman2004}: \enquote{...having a joint distribution in the imputation is less important than incorporating information from other variables and unique features of the dataset (e.g., zero/nonzero features in income components, bounds, skip patterns, nonlinearity, interactions). Conditional modeling allows enormous flexibility in dealing with practical problems.} }
In practice, the assumption about the existence of the joint distribution is often implicitly made, see, e.g., \cite{Drechsler2011}.

Borrowing notation from \cite{Drechsler2011}, we summarize the MICE algorithm:
\begin{enumerate}
  \item For every missing value in a dataset, impute it using a simple technique and treat these initial imputed values as \enquote{placeholders}.
  \item For a variable $Y_j$ set its \enquote{placeholders} back to missing.
  \item {Fit an imputation model for the observed values of $Y_j$ conditional on the currently filled-in values of all other variables, and perform an estimation task to capture model uncertainty.}
  \item Replace missing values of $Y_{j}$ with {a random draw from the conditional posterior distribution formulated during the estimation in Step 3.}
  \item Repeat Steps 2-4 for each $j$ with missing data to complete one cycle.
  \item Iterate Steps 2-5 for several cycles to update imputations until the convergence or the maximal number $n_{MI}$ of MICE iterations is reached.
  \item Repeat Steps 1-6 {independently in parallel to generate the desired number $n_{IS}$ of multiply imputed datasets.} 
\end{enumerate}

\begin{figure}[!ht]
    \centering
    \includegraphics[width=1\linewidth]{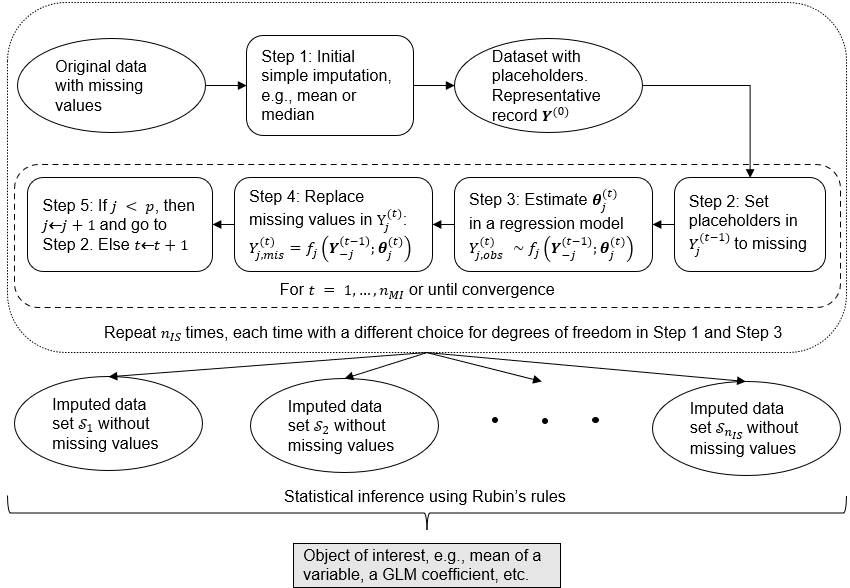}
    \caption{Schematic representation of MICE. $\Ybold^{(t - 1)}_{-j} := \rBrackets{Y^{(t)}_1,\dots,Y^{(t)}_{j - 1}, Y^{(t-1)}_{j+1}, \dots, Y^{(t-1)}_p}$, $Y^{(t)}_{j,mis}$ denotes the elements of the column corresponding to $Y^{(t)}_{j}$ that must be imputed $\Bigl($i.e., missing in the column corresponding to $Y^{(0)}_{j}\Bigr)$, $Y^{(t)}_{j,obs}$ denotes the elements in the respective column that were originally observed in $Y^{(0)}_{j}$, $f_j$ is an imputation model (e.g., a GLM, a random forest, etc.), $\bm{\theta}^{(t)}_j$ is the vector of parameters of $f_j$ in MICE iteration $t$.}
    \label{fig:MICE_scheme}
\end{figure}

{In Step 3 one can flexibly choose a model, i.e., it can be parametric or non-parametric. Depending on the chosen model, the estimation task involves either drawing parameters from a mathematical posterior distribution of model parameters or generating an empirical approximation of the posterior via resampling techniques such as bootstrapping. In Step 7, each independent run generates variation by utilizing new random draws in Steps 1 and 3.} The examples of factors contributing to the variability of the imputed datasets are: initial imputation of the missing data in Step 1 (mean, median, random draw from observed values, etc.), specification of the imputation model in Step 3 (e.g., bootstrap sampling if using random forests). Once multiple imputed datasets have been created, inference can be done regarding a quantity of interest (e.g., the mean of a variable, the correlation coefficient between two variables). The so-called Rubin's rules provide a statistical foundation for such inference. For a scheme that illustrates how MICE works, see Figure \ref{fig:MICE_scheme}.

{Having provided an overview of MICE, we comment on the convergence its underlying Markov chain (MC). According to standard MCMC theory, an MC converges if the following three properties are satisfied by the MC: irreducibility, aperiodicity, positive recurrence. As outlined in Subsection 4.5.2 of \cite{vanBuuren2018}, the first desired property is usually satisfied since the user of MICE has a huge flexibility to explore various regions of the state space. The second desired property (aperiodicity) may be violated when the imputation models are inconsistent and lead to a \enquote{ping-pong problem}. However, to prevent the MC from periodicity one can add randomness to the creation of imputed values, which is the case for random forests as an imputation model. The third desired property is positive recurrence. In \cite{vanBuuren2018} it is stated that the estimation of imputation model's parameters empirically leads to the absence or a mild presence of non-recurrence. 
    
The convergence of MICE method is usually evaluated by tracking lower-dimensional summary statistics (means, standard deviations, and cross-variable associations), but there are not enough theoretical results on this topic. One could apply the standard theory of MCMC methods, but only in a limited way for two reasons. First, unless the specified conditional models are compatible, there is no mathematical guarantee that the underlying MC in MICE targets a coherent joint posterior distribution, which is usually assumed in the literature on MCMC methods. Second, the formal tests for convergence require substantially more iterations than the relatively small number commonly used in MICE applications. In practice, MICE is often run for 5-10 iterations, which can be sufficient for empirical stabilization in many applications, as discussed in Subsections 4.5.2 and 4.5.5 of \cite{vanBuuren2018} and other literature sources. }
    
{Before we discuss how to use the MICE algorithm for generating synthetic datasets, we briefly mention its flexibility in applications. First, practitioners can explicitly enforce real-world data restrictions on a variable-by-variable basis. These constraints can be unconditional bounds on variables (e.g., the age of a driver cannot be less than the minimal age required by the respective legislation) or involve linear constraints (e.g., the age of a driver must be greater than or equal to the age of the youngest driver). Second, MICE naturally accommodates more complex inter-variable logic, such as hierarchical dependencies (e.g., if the variable indicating the presence of the second driver equals \enquote{false} then skip imputing variables related to the second driver) or compositional dependencies (e.g., total premium must be equal to the sum of liability premium and collision premium), through mechanisms like passive imputation. Deterministic relationships and derived fields are calculated algebraically from the most recently imputed base variables, preventing the generation of logically impossible combinations. Although imposing hard bounds on variables and deterministic relations between variables may result in theoretically incompatible conditional distributions, it provides the pragmatic control needed to generate realistic, logically consistent synthetic data. For more details on MICE practicality, interested readers are referred to Sections 3.4 and 3.5 in \cite{vanBuuren2011}, Section 3.2 in \cite{Drechsler2011}, and Section 5.4 in \cite{vanBuuren2018}. }

As mentioned earlier, the MICE framework can be used to generate synthetic datasets. For simplicity, we assume that the training data has no missing values; otherwise, a 2-step approach is needed, where missing values are imputed in the first step before generation (second step). The idea is to iteratively select parts of the original data and then replace them with synthetic values drawn from the posterior distribution learned by MICE.

There are different ways to sequentially choose chunks of data for replacement (imputation). One way is to choose a full column in each iteration, cf. \cite{Volker2021}. This case has two peculiarities. First, the imputation model is trained on the observed data, which is then replaced by the model prediction. Second, the MICE algorithm converges immediately, and each draw will be a direct draw from the posterior distribution, because a so-called monotone missingness pattern is satisfied for column-wise replacement/imputation (see Section 3.1.2 in \cite{Drechsler2011}). Another approach, which in general does not satisfy the monotone missingness pattern, is at each amputation-imputation step to randomly select  cells whose values must be replaced. The values in these cells are deleted, and treated as missing, and the MICE algorithm is executed, requiring more than one iteration to converge.

In our case studies, we test several ways to iteratively set a portion of data to missing and then apply MICE to impute. One of them is inspired by the \textit{tabulator} approach of \cite{Neves2022}, where the authors suggest setting data to missing in chunks of approximately equal size and to impute missing data using a GAIN. Instead of a GAIN, we will use MICE as the imputation engine.

Similarly to the MICE-based imputation, there are Rubin rules for analyzing (e.g., computing a point estimate, an estimate for the variance) unknown statistical quantities from fully synthetic datasets, cf. \cite{Reiter2003} and Section 6.1 in \cite{Drechsler2011}. 

In our case studies, we use random forests (RFs) as the imputation model. This imputation model and many others are implemented in the \textsf{R}-package \textbf{mice}, cf. \cite{vanBuuren2011}.  RFs were introduced in \cite{Breiman2001}. It is a non-parametric model that can capture complex relationships among variables without the need to specify a particular regression model. This makes RFs well-suited for imputing missing data in datasets with non-linear interactions among variables. Another important advantage of RFs is that they can often be used out-of-the-box and can be used rather easily to handle heterogeneous data. For more information on RFs in actuarial context, see, e.g., Chapter 4 in \cite{Denuit2020MLABookVolume2} or Chapter 6 in \cite{Wuethrich2025}.

\subsection{Conditional Tabular GAN}
In this subsection, which is based on \cite{Kuo2020} and Sections $2$ and $3$ of \cite{Cote2020}, we describe how a GAN works and explain how a CTGAN extends it.

A GAN consists of two neural networks (NNs): a generator $G$ and a discriminator $D$. The generator creates synthetic data samples, while the discriminator evaluates their authenticity. Through this adversarial process, the generator learns to produce increasingly realistic synthetic data that can fool the discriminator.

Let us denote by $\bm{\theta}_{G}$ and $\bm{\theta}_D$ the vectors of parameters of $G$ and $D$ respectively. $G$ takes as input random noise $\bm{Z} \sim F_{\bm{Z}}$, where $F_{\bm{Z}}$ is its cumulative distribution function, and maps the input to the space of the data to be generated. Usually, $F_{\bm{Z}}$ is a multivariate standard normal distribution and the dimension of $\bm{Z}$ is lower than that of the data point to be generated. The generator's output is a fake data point $G(\bm{Z}; \bm{\theta}_{G})$, whose distribution we denote by $F_{G}$. The discriminator's output $D(\bm{S}; \bm{\theta}_{D})$ is a score between $0$ and $1$, which can be interpreted as the probability of the input data point $\bm{S}$ to be a real data point. 

A GAN-training process (search for optimal values of parameters $\theta_{G}$ and $\theta_{D}$) is a min-max game in the following sense. On the one hand, the discriminator is trained by maximizing the expected score $\mathbb{E}_{\bm{S}}\sBrackets{D(\bm{S}; \bm{\theta}_{D})}$ of a real data point as well as minimizing (w.r.t. $\bm{\theta}_{D}$) the expected score of a synthetic data point $\mathbb{E}_{\bm{Z}}\sBrackets{D(G(\bm{Z}; \bm{\theta}_{G}); \bm{\theta}_{D})}$. On the other hand, the generator is trained by maximizing the discriminator's expected score $\mathbb{E}_{\bm{Z}}\sBrackets{D(G(\bm{Z}; \bm{\theta}_{G}); \bm{\theta}_{D})}$ on a synthetic data point $\bm{Z}$. So the objective function of a GAN combines both goals and is given by:
\begin{equation}\label{eq:GAN_objective_function}
    \min_{\thetab_G} \max_{\thetab_D} \rBrackets{ \mathbb{E}_{\bm{S}}\sBrackets{\ln \rBrackets{ D(\bm{S}; \bm{\theta}_{D})} } +  \mathbb{E}_{\bm{Z}}\sBrackets{\ln \rBrackets{ 1 - D(G(\bm{Z}; \bm{\theta}_{G}); \bm{\theta}_{D}) } } }.
\end{equation}

To improve the convergence of optimization routines for \eqref{eq:GAN_objective_function}, researchers have proposed modifications to \eqref{eq:GAN_objective_function} such as using Wasserstein-$1$ distance between the true distribution $F_{\Xbold}$ and the synthetic distribution $F_{G}$, enforcing additional constraints, adding penalties, etc.

Proposed in \cite{Xu2019}, CTGAN builds upon the GAN framework by incorporating several innovations to address the challenges of generating synthetic tabular data: mode-specific normalization, conditional generation of data, and handling of mixed data. Next we summarize an exemplary architecture of CTGAN, as in \cite{Cote2020}.

As per \cite{Kuo2020}, a CTGAN generates individual records by randomly selecting a variable (e.g., fuel type) and then choosing a value (e.g., diesel) based on the actual data frequency. The algorithm finds a matching training data row for this value and generates the remaining variables conditioned on the initial selection. Both the generated and true records are evaluated by the discriminator. Figure 3 in \cite{Cote2020} schematically illustrates this process and Figure 2 in \cite{Cote2020} visually exemplifies the architecture of the generator and discriminator. The discriminator (blue) and the generator (orange) each have two fully-connected layers to capture relationships between columns. The generator, which supports categorical variables, employs skip connections; see \cite{Xu2019} for extensions to continuous variables. CTGAN also uses the Packed Generative Adversarial Network (PacGAN) framework in the discriminator to mitigate a so-called mode collapse. This peculiarity means that the model focuses on a small subset of the real data and, thus, has a bias towards the generation of samples that are not sufficiently diverse.

Although a CTGAN tends to be substantially better than a GAN for generating high-cardinality categorical variables, this type of variables can still be challenging for a CTGAN, depending on the characteristics of the real data, such as the degree of cardinality and the frequency of each category. Aiming to improve the performance of a CTGAN for high-cardinality categorical variables, we try encoding them via non-linear autoencoders (AEs) before training a CTGAN. 

\subsection{Autoencoders}{\label{subsec:AE}}
The original aim of AEs is learning a compressed representation of data (encoding) and then reconstructing the original input (decoding). Structurally, an AE can be thought of as a pair of connected feedforward NNs. So principal components analysis approach can be interpreted as a linear AE, i.e., the activation functions of both NNs are linear. \cite{Hinton2006} demonstrated the power of non-linear AEs for learning better, lower-dimensional representations than linear AEs. In our paper, we use two different types of AEs due to their different application -- deterministic AEs and variational AEs.

Our application of deterministic AEs is the transformation of high-cardinality categorical variables to low-dimensional numerical vectors (instead of the one-hot encoding) before fitting them to GAN-based models for data generation. The benefits of using such AEs for training NNs in insurance predictive modeling have been studied in \cite{Delong2023}, which is our main reference for this application of non-linear AEs.

Consider a vector of features $\abold \in \mathbb{R}^{p}$, $p \in \mathbb{N}$.  An AE is a representation learning model that consists of two functions -- a so-called \textit{encoder} $\varphi: \mathbb{R}^{p} \rightarrow \mathbb{R}^{l}$ and a \textit{decoder} $\psi: \mathbb{R}^{l} \to \mathbb{R}^{p}$, $l \in \mathbb{N}$.

The mapping $\abold \mapsto \varphi(\abold)$ yields a $l$-dimensional representation of the $p$-dimensional vector. The mapping $\abbar \mapsto \psi(\abbar)$ tries to reconstruct the $p$-dimensional vector $\abold$ from its $l$-dimensional representation $\abbar = \varphi(\abold)$. The so-called reconstruction function of an AE is defined as
\begin{equation*}
    \pi = \psi \circ \varphi: \mathbb{R}^{p} \to \mathbb{R}^{p}.
\end{equation*}

Let $L: \mathbb{R}^{p} \times \mathbb{R}^{p}  \rightarrow \mathbb{R}$ be a function that measures the error (difference) between $\abold$ and $\pi(\abold)$. Then for a training dataset $(\abold_i)_{i=1}^{n}$, the objective of an AE is to find the functions $\varphi$ and $\psi$ that minimize the so-called reconstruction error defined as follows:
\begin{equation}\label{eq:AE_L_function}
   \frac{1}{n} \sum_{i=1}^{n} L(\pi(\abold_i), \abold_i).
\end{equation}

An AE is trained in an unsupervised manner. If an AE achieves a small reconstruction error, it is said that the encoder $\varphi$ extracts the most important information from a multi-dimensional vector of features $\abold$. In our application, $\abold$ will be a vector of features coming from a one-hot encoded categorical covariate, i.e., $\abold = \xbold_j^{cat}$ with $\xbold_j^{cat}$ defined as per \eqref{eq:one-hot-encoding}, $j \in \cBrackets{1, \dots, n_{CC}}$. For each categorical variable $\xbold_j^{cat}$, a separate AE is trained to reduce the dimensionality $p_j = m_j$ of  $\xbold_j^{cat}$ to $l_j < p_j$,  $j \in \cBrackets{1, \dots, n_{CC}}$. Afterwards, this representation $\overline{\xbold}_j^{cat} = \varphi\rBrackets{{\xbold}_j^{cat}}$, instead of ${\xbold}_j^{cat}$, is used as input for CTGAN to generate data. The approach of using a separate AE for each categorical variable is called \textit{Separate AEs} in \cite{Delong2023}. We use the same architecture of separate AEs, as proposed by \cite{Delong2023}. In particular, the reconstruction function $\pi$ is an NN with one hidden layer. The output-layer neurons have softmax activation functions and return the probabilities that the reconstructed feature takes a particular category. The category with the highest predicted probability is the one predicted for the reconstructed feature. $L$ in \eqref{eq:AE_L_function} is a cross-entropy loss function.

The approach described above is applied to all categorical features in the dataset. An example of the architecture of a neural network used in this paper to build the autoencoder of type \textit{Separate AEs} for categorical features (with 2 and 3 categories) is presented in Figure 2 in \cite{Delong2023}. See their Section 3.2 for further details on \textit{Separate AEs}.

Instead of using a separate AE for each categorical variable, we could have one joint AE for all categorical variables simultaneously. This approach is called \textit{Joint AE} in \cite{Delong2023}. According to Figure 4 there, for a large number of epochs (e.g., $100$ for the \textit{freMTPL2freq} data) and a sufficiently high dimension of the representation of all categorical features (e.g., $15$ for the \textit{freMTPL2freq} data), the reconstruction error power measured with the cosine similarity is very similar for \textit{Separate AEs} and \textit{Joint AE}. 

The second type of AEs used in our paper is variational autoencoders (VAEs). They were introduced by \cite{Kingma2013} and can be seen as a Bayesian extension to deterministic AEs.  VAEs learn a probabilistic representation of data points in a latent space, which is why these models can be used for synthetic data generation in addition to representation learning. In our case studies, we also consider the VAE-based approach proposed by \cite{Jamotton2024}, who also tested their methodology on a dataset with about $62000$ observations. For an overview of VAEs and details about their methodology, we refer the reader to their article.

\section{Case studies}\label{sec:case_studies}

This section is dedicated to our comparative study, which is based on the \textit{freMTPL2freq} dataset. In Subsection \ref{subsec:experiment_setting}, we provide the details about our method training pipelines, the mechanics of data augmentation, and the evaluation pipelines for generation methods. In Subsection \ref{subsec:experiment_evaluation}, we specify all the methods tested and state explicit formulas for modeling the structural dependence between observed covariates and simulated claim counts that represent \enquote{ground truth} in tested scenarios. The results are discussed in Subsection \ref{subsec:experiment_results}. 

\subsection{General setting and evaluation metrics}\label{subsec:experiment_setting}

Our aim with this case study is to answer the research questions stated at the end of Section \ref{sec:general_setup} and, in doing so, utilize pre-existing implementations of synthetic data generation methods with minimal user interventions for data preparation or method training. This aspect is significant in practice since, in our experience from the industry, actuaries seldom have sufficient time to implement and fine-tune their specific algorithms and methods. Since we want to identify the best \enquote{off-the-shelf} data generation approach, we do not do any pre-processing of variables when preparing data in this section. Any processing is done by the generation method implementation. We execute our case study on the high-performance computation cluster at the University of Tartu and can provide our code upon request.

In our experiments, we use the \textit{freMTPL2freq} dataset to give us a baseline of what a realistic dataset might look like. It has $678013$ observations with $9$ explanatory variables ($5$ categorical and $4$ numeric), $1$ exposure variable and $1$ response variable (the number of claims). This dataset has been used in multiple actuarial data science articles, including \cite{Kuo2020}, where it is used for training a CTGAN to generate synthetic data. {The exploratory analysis of the above-mentioned dataset can be found in Section 13.1 in \cite{Wuethrich2023}. According to Table 13.3 there, the Pearson's correlation and the Spearman's rho for the numeric variables are low, except for the pair of variables driver age and bonus malus. \cite{Wuethrich2023} also find a considerable association among categorical variables the vehicle brand and region as well as some association between vehicle brand and both vehicle power and vehicle age. The \textit{freMTPL2freq} dataset is the de facto benchmark in modern actuarial data science, and introducing non-standard datasets would reduce comparability with existing literature.} 

Although we have access to the original values of the observed response variable $y_i$ in the \textit{freMTPL2freq} dataset, we do not know the true relation between the observed value $y_i$ and the corresponding observed covariates $\xbold_i$. To test whether the generation methods learn the relation correctly, we simulate $Y_i \sim Poisson\rBrackets{\exp\rBrackets{f(\xbold_i)}}$, where $f(\xbold_i)$ is a known function of covariates, and use the realized value as $y_i$. Thus, we implicitly assume that all observations have the same exposure $1$. We deem this assumption acceptable, since, in practice, the rate is set using the values for annual policies (e.g., yearly predicted number of claims), which are scaled to fit the duration desired by the client.

We test two specific forms of ${f}$. First, we assume that ${f}$ is linear in all covariates $x_j$, $j\in \{0,\dots,d\}$, where $d+1$ is the number of terms (including the intercept) in the model. This yields ${f}_{lin}(\xbold) = \displaystyle \sum_{j = 0}^{d} \beta_j^* x_j$, where coefficients $\beta_j^* \in \mathbb{R}$ are known and serve as the \enquote{true} values defining the relationship between the response variable and the covariates in pre-processed form (cf. Section \ref{sec:general_setup}). From now on, ${f}_{lin}$ is referred to as the \textit{ linear formula}.

In the second formula we use the previously defined ${f}_{lin}$ and additionally assume a set of interactions among covariates, namely ${f}_{int} (\xbold) = {f}_{lin}(\xbold) + 
    \sum_{\{j,k\}\in \mathcal{I}}\beta_{j;k} x_j x_k$,
where the interaction part is defined as in Equation (\ref{eq:interaction_term}) and $\mathcal{I}$ is the set of index pairs for the interactions of interest. Although we use double indices for notational clarity, one can construct a simple mapping from double indices to single indices. The values ${f}_{lin}(\xbold_i)$ and ${f}_{int}(\xbold_i)$ are then used as the means of the respective Poisson distributions to simulate the number of claims $y_i$, $i\in \cBrackets{1, \dots, n_{ROD}}$.

So, in our experiments, we have two datasets, each of which contains simulated number claims for which we know their \enquote{true} structural dependence on original covariates. In the sequel, we refer to this artificially generated number of claims as the response variable and use the $90\%/10\%$ training/test split, i.e., the number of \textbf{r}ows in the \textbf{tr}ain set is $n_{RTR} = \lfloor n_{ROD} \cdot 0.9\rfloor$. 

Let now $\ASet = \cBrackets{A_1, \dots, A_{n_A}}$ be the set of algorithms (approaches) for generation of synthetic data, $n_{A} \in \mathbb{N}$ is the number of algorithms for synthetic data generation. For each approach $A$, we denote the generation method for this algorithm by $GM_{A}$. 

For each algorithm, although the technical specifics of training can vary, the underlying pipeline we use is the same. We supply each algorithm the training data $\SSet_{train}$ with the artificially generated response variable and without altering any of the covariate variable values. Then each of the methods utilizes the supplied training data in accordance with its implementation and, thus, trains the generation method $GM_{A}$. However, for some methods AEs (see Section \ref{subsec:AE}) are utilized to transform categorical variables. This requires additional steps in the training pipeline, which are outlined in Appendix \ref{app:summary_AE}.

Each generation method is used to generate $n_{E}$ (referred to as the \textit{number of experiments}) synthetic datasets, where the number of rows for a single synthetic dataset is the same as in the training data. Denote a generated synthetic dataset as $\SSet^{A;k}$, where $k \in\{1,\dots,n_{E}\}$ and $A \in \ASet$.

Using the described pipeline, synthetic datasets $\SSet^{A;k}$ are generated for each algorithm $A \in \ASet$ and each $k \in \{1,\dots,n_{E}\}$. Then, for each synthetic dataset, we compute two types of metrics: the evaluation of the dataset metrics and the evaluation of model metrics.

With dataset metrics, we want to assess whether the proportion of data in some subset of synthetic data is the same or close to the proportion of data in the original training data. Statistical tests like Kolmogorov-Smirnov or Anderson-Darling for numeric variables and $\chi^2$-test for categorical variables could be used. However, if the data size is large, then these tests are very sensitive to the slightest differences between synthetic data and original training data. Therefore, we use other numeric metrics to see how well the proportions of data are captured across all variables and pair-wise combinations of variables.

For dataset evaluation, we compare the proportions of specific values in the original and synthetic data. The comparison of data is done on discretized forms of the original variables.

Denote the original variable $j$ before pre-processing by $x_j^{orig}$. If $x_j^{orig}$ is a categorical variable, then no additional steps need to be taken, since after pre-processing, the variable will be discretized into a vector of binary variables $\xbold_j^{cat}$. 

If $x_j^{orig}$ is numeric, then it is discretized into $n_{bin_j}$ bins, where bin cut-off values are selected based on the quantiles of the numeric variable values in the original training data. The quantiles are used to ensure that the proportion of observations in each bin is the same or very close. Note that if a value appears in several quantiles (e.g., it is very prevalent in the data), it has its own bin. For each bin, then, a binary variable (feature) is added, and the vector of these features is denoted by $\xbold_j^{cat}$. Additionally, we denote by $\mathcal{I}_j$ the set of indices for pre-processed variables corresponding to $\xbold_j^{cat}$ .

For each feature $x_k$ from $\xbold_j^{cat}$ in training data, we calculate the ratio of observations belonging to that bin or category as $r\fixparen{x_k} = \frac{\sum_{i = 1}^{n_{RTR}} x_{i;k}}{n_{RTR}}$, where $k \in \mathcal{I}_j$. Similarly, we obtain $\widehat{r}\fixparen{x_u}$ for any synthetic data.

These ratios {are used to calculate Mean Absolute Error ($MAE$) and Mean Absolute Percentage Error ($MAPE$) for the original variable $x_j^{orig}$:}
\begin{equation*}
    MAE(x_j^{orig}) = \frac{1}{\left| \mathcal{I}_j\right|}\displaystyle\sum_{k \in \mathcal{I}_j} \left| r\fixparen{x_k} - \widehat{r}\fixparen{x_k}\right| \quad \text{and} \quad MAPE(x_j^{orig}) = \frac{1}{\left| \mathcal{I}_j\right|}\displaystyle\sum_{k \in \mathcal{I}_j} \frac{\left| r\fixparen{x_k} - \widehat{r}\fixparen{x_k}\right|}{\left|\widehat{r}\fixparen{x_k}\right|},
\end{equation*}
\noindent where $\left| \mathcal{I}_j\right|$ denotes the number of elements in set $\mathcal{I}_j$.

For pairwise data proportion assessment, we extend the above metrics to pairs of variables. Given two (original) variables $x_a^{orig}$ and $x_b^{orig}$, the ratio for any two features $x_k$ and $x_u$ with $k \in \mathcal{I}_a$ and $u\in \mathcal{I}_b$ in the training data is computed as $r\fixparen{x_k,x_u} = \frac{\sum_{i = 1}^{n_{RTR}} x_{i;k} \cdot x_{i;u}}{n_{RTR}}$. For synthetic data, $\widehat{r}\fixparen{x_k,x_u}$ can be obtained in the same way. The metrics in this case are defined analogously to the single-variable case, except the summation is done over both sets of indices $\mathcal{I}_a$ and $\mathcal{I}_b$.

Lastly, for numerical variables, it is possible to assess if the correlations among them are retained by the data generation method. We calculate the pairwise correlations $\rho(x_k,x_u)$ for all numeric variable pairs $(x_k,x_u)$ in the training data. For synthetic data, we obtain the correlations in a similar way, and then calculate $MAE$ and $MAPE$ for correlations analogously to ratios above.

We use the synthetic datasets $\SSet^{A;k}$ and the training data $\SSet_{train}$ as the reference for dataset metrics evaluation. For model metrics, the scope is expanded to better understand what effect \emph{data augmentation} has on predictive model performance.

Under the term \emph{data augmentation}, we mean the creation of $m$ new datasets for each synthetic dataset $\SSet^{A;k}$. More precisely, given a synthetic dataset $\SSet^{A;k}$, we split it (row-wise) into $m$ disjoint (approximately) equally sized parts $\SSet_l^{A;k}$, $l \in \{1,\dots,m\},$ and append the increasing collection of these synthetic data parts to the original training data. 
In order to describe all possible combinations of training and synthetic data, we introduce a new structure {tuple} ${\color{blue}\bm{\tau}}=(t,L)$, where $t$ is the indicator stating whether the training set is included ($t=1$) or not ($t=0$), and $L$ is the number of included synthetic data parts.
With this setup, we then get 
\begin{equation*}
    \SSet^{A;k;(0,L)} = \bigcup_{l = 1}^{L} \SSet_l^{A;k}\text{ and }\SSet^{A;k;(1,L)} = \SSet_{train}\bigcup\SSet^{A;k;(0,L)},
\end{equation*}
where $L \in \{1,\dots,m\}$, $\SSet^{A;k} = \bigcup_{l = 1}^{m} \SSet_l^{A;k}$ and $\SSet_j^{A;k} \bigcap \SSet_l^{A;k} = \emptyset$,  $\forall j \neq l$.
By construction we have $\SSet^{A;k;(1,0)} =\SSet_{train}$ and $\SSet^{A;k;(0,m)} =\SSet^{A;k}$.

For the evaluation of model metrics, the underlying models need to be trained. We focus on using GLMs as the predictive model type. For each dataset $\SSet^{A;k;\bm{\tau}}$, two predictive GLM models are derived. One model, where the formula used to build the model is the same as the corresponding true model formula (i.e., \emph{linear formula} or \emph{interaction formula}), and one which is derived using a bidirectional step-wise variable selection using $AIC$. Note that for the \emph{interaction formula}, variable selection included explicitly defined columns for each of the interactions between the different variables, as this was the easiest way to include interactions in the step-wise search. The values related to the trained GLM models are denoted as follows:
\begin{itemize}
    \item $\betab^\ast = \rBrackets{\beta_0^\ast,\beta_1^\ast,\dots,\beta_d^\ast}^\top$ -- the vector of true model coefficients as defined in the respective response variable generation formula (i.e., \emph{linear formula} or \emph{interaction formula}).
    \item $\widehat{\betab} = \left(\widehat{\beta_0},\widehat{\beta_1},\dots,\widehat{\beta}_d\right)^\top$ -- the vector of estimated model coefficients based on pre-specified true model structure, using the GLM trained on the training dataset $\SSet_{train}$.
    \item $\widehat{\betab}^{A;k} = \left(\widehat{\beta}_{0}^{A;k},\widehat{\beta}_{1}^{A;k} ,\dots,\widehat{\beta}_{d}^{A;k} \right)^\top$ -- the vector of estimated model coefficients based on pre-specified true model structure, using the GLM trained only on the synthetic dataset $\SSet^{A;k}$.
    \item $\widehat{\betab}^{A;k;\bm{\tau}} = \left(\widehat{\beta}_{0}^{A;k;\bm{\tau}},\widehat{\beta}_{1}^{A;k;\bm{\tau}} ,\dots,\widehat{\beta}_{d}^{A;k;\bm{\tau}} \right)^\top$ -- the vector of estimated model coefficients based on pre-specified true model structure, using the GLM trained on the dataset specified by the data structure parameter $s$.
    \item $MSE\left(\widehat{\beta}_{j}^\bullet\right)$ -- mean squared error of the estimated coefficient $\widehat{\beta}_{j}^\bullet$, where "$\bullet$" represents any algorithm/data structure combination.
\end{itemize}

As these values can be obtained for every trained GLM model, we use two metrics to assess the effect of synthetic data on the estimated values of model coefficients.

First, to measure the distance between the true GLM coefficients and their estimate based on the true structure of a GLM and a synthetic dataset generated by an approach $A$ and specified by the structure tuple $\bm{\tau}$, we use the following metric:
\begin{equation}\label{eq:coefficient_distance_metric}
    M_1\fixparen{A,\bm{\tau}} = \frac{1}{d}\frac{1}{n_E} \sum_{k = 1}^{n_E}
    \sum_{j = 0}^{d} \frac{\left(\beta_j^\ast - \widehat{\beta}^{A;k;\bm{\tau}}_j\right)^2}{MSE\left(\widehat{\beta}_j\right)}
    .
\end{equation}
In (\ref{eq:coefficient_distance_metric}), {${MSE\left(\widehat{\beta}_j\right)}$ denotes the mean squared error of the coefficient estimator based on the original training data, approximated by its  estimated variance}. This ensures that the distance values can be compared with each other.

The second metric we use is defined as follows:
\begin{equation}\label{eq:SD_increase}
    M_2\fixparen{A,\bm{\tau}} = \frac{1}{d}\sum_{j = 1}^{d} \frac{MSE\fixparen{\widehat{\beta}_j} + \frac{1}{n_E}\sum_{k = 1}^{n_E}\fixparen{\widehat{\beta}_j-\widehat{\beta}^{A;k;\bm{\tau}}_j}^2}{MSE\fixparen{\widehat{\beta}_j}}.
\end{equation}
Assuming that the estimation of coefficients is unbiased, the metric defined by Formula \eqref{eq:SD_increase} is motivated by the following equality:
    \begin{align*}
     \mathbb{E}\sBrackets{\rBrackets{\beta_j^\ast-\widehat{\beta}_{j}^{\bullet}  }^2 } & =  \mathbb{E}\sBrackets{\rBrackets{\beta_j^\ast  -\widehat{\beta}_{j} + \widehat{\beta}_{j} - \widehat{\beta}_{j}^{\bullet}}^2 }  =  \mathbb{E}\sBrackets{\rBrackets{
        \beta_j^\ast-\widehat{\beta}_{j}}^2}+
        \mathbb{E}\sBrackets{\rBrackets{\widehat{\beta}_{j} -\widehat{\beta}_{j}^{\bullet} }^2}.
    \end{align*}

By construction, $M_2\fixparen{A,\bm{\tau}} = 1 + \delta$, where $\delta \geq 0$. In an ideal situation, $\delta$ is very small, thus meaning that the increase in coefficient value uncertainty is small when synthetic data is used.

We also test all models based on model goodness-of-fit metrics on the unseen dataset $\SSet_{test}$. Denote the observed and predicted values of the response variable by $\mathbf{y}$ and $\widehat{\mathbf{y}}$, respectively.
    {Poisson deviance 
       $DEV_{Po}\fixparen{\mathbf{y},\mathbf{\widehat{y}}} = 
            \frac{2}{n}\sum_{i = 1}^{n}\fixparen{y_i\ln\fixparen{\frac{y_i}{\widehat{y}_i}}-\fixparen{y_i - \widehat{y}_i}}$}
    and 
        $RMSE\fixparen{\mathbf{y},\mathbf{\widehat{y}}} = \sqrt{\frac{1}{n} \sum_{i =1}^{n}\fixparen{y_i-\widehat{y}_i}^2}$
are the model goodness-of-fit metrics that we use.

Since we are interested in the underlying performance of each approach $A$, all metrics will be averaged across $n_E$ different experiments related to $A$. The number of experiments $n_E$ should be as large as possible to reduce uncertainty in the evaluation of the algorithm.

Last but not least, we subjectively assess the ease of use of each method. This aspect is often ignored by other studies, as they mainly focus on the performance of the methods.

\subsection{Tested methods and formulas}\label{subsec:experiment_evaluation}

In our case study, we test various methods that can be grouped into four groups: GAN-related methods, MICE-related methods, hybrid methods and other methods. In the description of the methods, the monospaced shorthand used in the results is shown inside the parentheses. Where possible, we use the \enquote{off-the-shelf} implementation of a method.

The first group of methods we test are GAN-based, namely:
\begin{itemize}
    \item CTGAN (\texttt{CTGAN}) -- implementation of \cite{Xu2019} available in the \href{https://docs.sdv.dev/sdv}{sdv} library for Python;
    \item CTGAN with AEs (\texttt{CTGAN\_WITH\_AE}) -- a CTGAN enhanced by AEs for encoding-decoding categorical variables;
    \item Wasserstein GAN with gradient penalty (\texttt{COTE\_MC\_WGAN\_GP}) -- implementation of the method graciously provided by \cite{Cote2020};
    \item GAN-based tabulator (\texttt{JCS\_TABULATOR}) -- tabulator as proposed and implemented by \cite{Neves2022}.
\end{itemize}

The second group of synthetic-data generators we test are the imputation-amputation methods based on MICE.  One of the considered approaches is the \textit{tabulator} proposed by \cite{Neves2022}, but we use MICE with RFs as the imputation engine instead of the GAIN-based models suggested there. In our MICE-based tabulator, we have $5$ sequential amputation-imputation iterations, each of which sets random $20\%$ of cells to missing and imputes them using MICE-RF to generate synthetic data points. The other two methods have a different amputation logic. First, we create a copy of the training data, where we set $75\%$ of cell values to missing. Second, we append this data with missing values to the original training data and apply MICE-RF to impute the missing values. The data with imputed cells is treated as generated data. {In this way, we obtain generated data where 25\% of the cell values remain unchanged and 75\% are synthetic. Although the resulting dataset is partially synthetic, the missing-completely-at-random (MCAR) selection of cells for replacement is expected to reduce the probability that a generated row exactly matches a row in the original training data.\footnote{{This approach does not simply refill the removed values with their original values. The MCAR structure determines which values are removed, while MICE generates new values for the resulting missing entries using fitted imputation models. In the RF imputation algorithm, each imputed value is randomly sampled from observed donor values associated with the relevant terminal nodes; see Algorithm A.1 in \cite{doove2014recursive}.}} However, this procedure does not provide a formal privacy guarantee or rule out other forms of disclosure risk.}


Hereinafter, we refer to this approach as the \enquote{MICE partially synthetic method}. The third approach extends the MICE partially synthetic method by the second iteration, in which the cells that were not imputed in the first iteration (the remaining quarter of the original data) are set to missing and then imputed via MICE. This results in a fully synthetic dataset. This method is referred to as \enquote{MICE fully synthetic method}. For each of the approaches described in this paragraph, we use only one synthetic dataset for evaluating data fidelity and data utility. This is different from the original idea of MICE for statistical inference (see Subsection \ref{subsec:MICE_theory}, Figure \ref{fig:MICE_scheme}), where multiple different imputed datasets are created and then combined to study a statistical object with the help of Rubin's rules. In summary, the MICE-based methods we explore and benchmark in our case study are:
\begin{itemize}
    \item MICE partially synthetic method (\texttt{MICE\_PART\_SYN}) -- the method described above and implemented using the \textsf{R}-package \href{https://cran.r-project.org/package=mice}{\texttt{mice}};
    \item MICE fully synthetic method (\texttt{MICE\_FULL\_SYN}) -- the modification of the partially synthetic method, where in an additional step the remaining $25\%$ of the original data is imputed via MICE;
    \item MICE tabulator (\texttt{MICE\_TABULATOR}) -- the tabulator setup from \cite{Neves2022} with $5$ iterations and MICE with RFs as the imputation engine.
    \item MICE as per Volker-Vink (\texttt{MICE\_VV}) -- amputation-imputation procedure utilizing MICE column-wise as described in \cite{Volker2021}.
\end{itemize}

{For all MICE-based approaches, we used simple random draw for imputation initialization and $5$ iterations of model training with left-to-right visiting sequence (the number of artificial claim counts variable being the last). For the RF imputation method, we use $10$ trees and the default parameters of the \texttt{ranger} package. Among those parameters, the most relevant is the unlimited maximum tree depth, with a minimum node size of $1$ or $5$ (depending on the column type). We did not conduct a separate sensitivity analysis regarding these implementation choices. Consequently, the reported results are conditional on the specifications used.}

Before we move on to other synthetic data generators in our case study, we clarify the different types of \enquote{iterations} in MICE-based methods. There are amputation-imputation iterations and there are MICE iterations. The former iteration type means that, first, a not yet imputed part of the training data is set to missing and, second, the respective missing values are imputed (using MICE-RF in our case study). A MICE iteration should be understood as a sequence of actions in Steps 2-5 in Figure \ref{fig:MICE_scheme} done one single time, i.e., once for each variable creating missingness, training an RF that uses other variables to predict the variable with missingness, and replacing introduced missing values by RF predictions.

For example, in our case study, the MICE-tabulator approach has $5$ amputation-imputation iterations, the MICE partially synthetic method has $1$ amputation-imputation step, and the MICE fully synthetic method has $2$ amputation-imputation steps. In the MICE-based generation as per \cite{Volker2021}, there is no explicit creation of missingness before training imputation models. Instead, for each column, an imputation model is trained without randomly deleting observed values. Then each observed value is set to missing and then replaced by the respective prediction from the trained imputation model. The resulting imputed column is then used as the predictor for further imputation in the chain of columns. Since a single pass over the original dataset is performed only once to obtain a fully synthetic dataset, there are in total $n_C$ amputation-imputation iterations.

After testing the above-described groups of methods, we noticed that GAN-based approaches learned the univariate distributions of numeric variables worse than MICE-based methods and had challenges with high-cardinality categorical variables. Due to this finding, we decided to consider two hybrid models. One of them combines a CTGAN and a MICE component, namely a CTGAN first generates synthetic data and then $75\%$ of the synthetic numeric variable values are set to missing and then imputed using MICE. Note that also for this CTGAN-MICE model, the data with removed cell values was combined with the original training data to impute the missing numerical column values. The second hybrid model that we tested enhances CTGAN-MICE by using {separate} AEs to encode categorical variables before applying CTGAN-MICE and then using clustering with AEs to decode synthetic AE-encoded categorical variables. So, the hybrid approaches we study are: 
\begin{itemize}
    \item CTGAN with MICE (\texttt{CTGAN\_MICE}) -- a CTGAN enhanced by MICE for numeric variables;
    \item CTGAN with AEs and MICE (\texttt{CTGAN\_WITH\_AE\_MICE}) -- a CTGAN enhanced by AEs for categorical variables and MICE for numeric variables.
\end{itemize}

{Both \texttt{CTGAN\_WITH\_AE} and \texttt{CTGAN\_WITH\_AE\_MICE} use separate AEs to encode categorical variables. Separate AEs encode each categorical variable independently and therefore do not directly exploit dependencies among categorical variables. However, as mentioned in Section \ref{subsec:AE}, \cite{Delong2023} find that separate and joint AEs achieve similar reconstruction performance after sufficiently extensive training. Moreover, in our pipeline, the separately encoded categorical variables are subsequently provided jointly to CTGAN, which can learn dependencies among their encoded representations. We did not directly compare separate and joint AEs within the complete synthetic-data generation pipeline and leave such a comparison for future research.}

Lastly, we test the method proposed in \cite{Jamotton2024}, which uses variational autoencoders (VAEs) for data generation. We use the implementation graciously shared with us by the authors of \cite{Jamotton2024}. The monospaced shorthand for this method is \texttt{VAE\_JAMOTTON}.

As described in Subsection \ref{subsec:experiment_setting}, artificial claim counts are generated for this case study in such a way that we know the true relationship between the response variable and the descriptive variables. The following formula for a linear relationship is used:
\begin{align}\label{eval:linear_formula}
\begin{split}
     f_{lin}(\xbold) &= -3 + 0.0075\cdot (\text{VEHICLE\_AGE}-5) + 0.0075\cdot (\text{DRIVER\_AGE}-35) \\
     &+ 0.0075\cdot (\text{BONUS\_MALUS}-100) + 0.15\cdot \mathds{1}_{\fixparen{\text{AREA} \in \text{\{A,C,E\}}}}\\
     &- 0.5\cdot \mathds{1}_{\fixparen{\text{AREA} \in \text{\{B,F\}}}} - 0.5\cdot \mathds{1}_{\fixparen{\text{VEHICLE\_POWER} \in \text{\{4,5,6\}}}}\\
     &-0.4\cdot \mathds{1}_{\fixparen{\text{VEHICLE\_POWER} \in \text{\{7,8,9\}}}}+0.15\cdot \mathds{1}_{\fixparen{\text{VEHICLE\_POWER} \in \text{\{12,13,14\}}}}\\
     &+0.3\cdot \mathds{1}_{\fixparen{\text{VEHICLE\_POWER} \in \text{\{15\}}}} - 0.5\cdot \mathds{1}_{\fixparen{\text{VEHICLE\_BRAND} \in \text{\{B3,B4,B5\}}}}\\
     &+0.2\cdot \mathds{1}_{\fixparen{\text{VEHICLE\_BRAND} \in \text{\{B10,B11\}}}}+0.6\cdot \mathds{1}_{\fixparen{\text{VEHICLE\_BRAND} \in \text{\{B13,B14\}}}}\\
     &+0.45\cdot\mathds{1}_{\fixparen{\text{VEHICLE\_GAS} \in \text{\{Diesel\}}}}.
\end{split}
\end{align}
Note that the DENSITY and REGION variables available in the \textit{freMTPL2freq}  dataset are not used in generating the claim counts. This is done for the purpose of variable selection testing.

For the interaction formula, two interactions between numeric and categorical variables and two interactions between two categorical variables are added on top of the linear formula:
\begin{align}\label{eval:interaction_formula}
     f_{int}(\xbold) &= f_{lin}(\xbold) + 0.0015\cdot\text{BONUS\_MALUS}\cdot \mathds{1}_{\fixparen{\text{AREA} \in \text{\{A,B,C\}}}} \notag \\
     &-0.003 \cdot\text{BONUS\_MALUS}\cdot \mathds{1}_{\fixparen{\text{AREA} \in \text{\{D,E\}}}} \notag \\
     &+0.015 \cdot\text{VEHICLE\_AGE}\cdot \mathds{1}_{\fixparen{\text{VEHICLE\_POWER} \in \text{\{10,11,12,13,14,15\}}}} \notag \\
     &-0.015 \cdot\text{VEHICLE\_AGE}\cdot \mathds{1}_{\fixparen{\text{VEHICLE\_POWER} \in \text{\{4,5\}}}} \notag \\
     &+0.15\cdot\mathds{1}_{\fixparen{\text{VEHICLE\_GAS} \in \text{\{Diesel\}}}}\cdot\mathds{1}_{\fixparen{\text{VEHICLE\_POWER} \in \text{\{4,5,6,7\}}}} \\
     &-0.25\cdot\mathds{1}_{\fixparen{\text{VEHICLE\_GAS} \in \text{\{Regular\}}}}\cdot\mathds{1}_{\fixparen{\text{VEHICLE\_POWER} \in \text{\{10,11,12,13,14,15\}}}}  \notag \\
     &+0.4\cdot\mathds{1}_{\fixparen{\text{AREA} \in \text{\{A,B,D\}}}}\cdot\mathds{1}_{\fixparen{\text{VEHICLE\_BRAND} \in \text{\{B1,B4,B10\}}}} \notag \\
     &+0.2\cdot\mathds{1}_{\fixparen{\text{AREA} \in \text{\{C,D,E\}}}}\cdot\mathds{1}_{\fixparen{\text{VEHICLE\_BRAND} \in \text{\{B2,B6,B11,B12\}}}} \notag \\
     &-0.6\cdot\mathds{1}_{\fixparen{\text{AREA} \in \text{\{A,C,E,F\}}}}\cdot\mathds{1}_{\fixparen{\text{VEHICLE\_BRAND} \in \text{\{B3,B5,B13,B14\}}}} \notag \\
     &-0.3\cdot\mathds{1}_{\fixparen{\text{AREA} \in \text{\{F\}}}}\cdot\mathds{1}_{\fixparen{\text{VEHICLE\_BRAND} \in \text{\{B1,B2,B12\}}}}. \notag
\end{align}
The claim counts simulated using \eqref{eval:linear_formula} and \eqref{eval:interaction_formula} lead to the average claim frequency of $5.37\%$ and $6.04\%$, respectively. These numbers are close to the values observed in practice {in our region (Estonian Motor Insurance Bureau, 2025; statistics available in Estonian: \url{https://www.lkf.ee/et/statistika})}.

We acknowledge that the chosen linear and interaction formulas follow the setup of generalized linear models, which is the core basis of insurance ratemaking in practice. {All data generation methods considered in this case study are non-linear models. Nevertheless, the GLM-based response-generating mechanisms restrict the complexity of the relationships tested here, and the comparative performance of the methods may differ under more complex dependence structures.} The topic of more complex dependence structures is left as an avenue for further research.

\subsection{Case study results}\label{subsec:experiment_results}
The case study showcases how different methods for data generation perform across a variety of different performance metrics defined in the previous sections. We do not claim that one particular method is best suited for synthetic data generation, but we assess the strengths and weaknesses of different methods.

With our case study, we evaluated a total of $11$ different approaches to synthetic data generation. However, we display the results only for $10$ of them, as the tested tabulator by \cite{Neves2022} did not generate any number of claims. {Upon investigation, we did not find a clear reason why the implementation provided by \cite{Neves2022} should fail, but the dataset generated by GAN-based imputation failed to adequately generate both the predictor variables and response variables. The tabulator approach using MICE as the imputation engine did work, so we think the problem is related to the GAN-based imputation model used in the provided implementation.}

In total, we have two sets of results, one for the case linear formula (\ref{eval:linear_formula}), and one for the case where the interaction formula (\ref{eval:interaction_formula}) is used. The cases differ in complexity and inter-variable relationships used to generate the artificial number of claims.

For both cases, the methods are assessed based on the model-specific and data-specific metrics. Model-specific metrics are coefficient difference metrics $M_1$ (\ref{eq:coefficient_distance_metric}) and $M_2$ (\ref{eq:SD_increase}), the number of correctly selected (using the variable selection method) variables {(present in claim generation formulas (\ref{eval:linear_formula}) or (\ref{eval:interaction_formula}))} the number of incorrect variables selected {(additional variables not present in the claim generation formulas (\ref{eval:linear_formula}) or (\ref{eval:interaction_formula}))}, and lastly Poisson deviance and RMSE of the trained GLM (model with only correct variables). For the interactions case, the number of missing main effects (case where the model has the interaction term but not the corresponding main effect) is also shown. 

Data-specific metrics are MAE/MAPE for numeric, categorical and pair-wise combinations of columns and MAE/MAPE for correlation structure within numeric columns {and categorical (Cramér's V)} of the generated data. Lastly, our subjective ranking of the method's ease of use is also provided.

Note that every synthetic dataset is uniquely generated 5 times ($n_E = 5$) and all of the following results are presented as simple averages across all generated datasets for a given method (except for $M_1$ and $M_2$ metrics, since these incorporate $n_E$). For data-specific metrics, each column is taken as equally important, so the metrics reflect the average of all columns or combinations. {Additionally, the marginal distribution of each variable for both dependence structure settings can be seen in the Appendix \ref{app:comparison_of_marginals}. Because only five repetitions were conducted and run-level variability was not systematically analyzed, the reported results should not be interpreted as establishing differences in the stability of the considered generators.}

The final results for the evaluation of methods are presented in Tables \ref{results:linear_table} and \ref{results:interaction_table}. Each cell shows the rank of the method with respect to the metric shown in the column. Additionally, the table is color-matched based on the rank of the method, and in the case of ties, with respect to the metric, all cells are given the same highest rank available. Note that in all cases, the "training" method is shown as the reference value for different metrics {and corresponds to the metric evaluated using only the original training data. Additionally, the methods are ordered in the tables according to the following groups: MICE-based methods, CTGAN-based methods, the VAE-based method, and WGAN-based method.}

Before looking at the results, it is important to note that in our case study, all generators were trained using the University of Tartu high-performance computation cluster (\cite{UTHPC}). Because this cluster does not inherently have hardware stability, we were unable to directly compare fitting and generation times. However, we observed that \texttt{MICE}-based and \texttt{CTGAN}-based methods usually took the least time (around 3 hours for model fitting and data generation), while more complex custom implementations, such as \texttt{COTE\_MC\_WGAN\_GP}, took significantly longer for model fitting and generation (about 10 hours in total).

\begin{table}[!ht]
\centering
\resizebox{\textwidth}{!}{%
\begin{tabular}{|l|r|r|r|r|r|r|r|r|r|r|r|r|r|r|r|r|r|}
\hline
\textbf{Generation   method}   & \multicolumn{1}{l|}{\textbf{M1}}                                                      & \multicolumn{1}{l|}{\textbf{M2}}                                                      & \multicolumn{1}{l|}{\textbf{\begin{tabular}[c]{@{}l@{}}Correct \\ variables \\ selected\end{tabular}}} & \multicolumn{1}{l|}{\textbf{\begin{tabular}[c]{@{}l@{}}Incorrect \\ variables \\ selected\end{tabular}}} & \multicolumn{1}{l|}{\textbf{\begin{tabular}[c]{@{}l@{}}Fit \\ Poisson \\ deviance\end{tabular}}} & \multicolumn{1}{l|}{\textbf{\begin{tabular}[c]{@{}l@{}}Fit\\ RMSE\end{tabular}}}    & \multicolumn{1}{l|}{\textbf{\begin{tabular}[c]{@{}l@{}}Pair-wise \\ MAE\end{tabular}}} & \multicolumn{1}{l|}{\textbf{\begin{tabular}[c]{@{}l@{}}Pair-Wise \\ MAPE\end{tabular}}} & \multicolumn{1}{l|}{\textbf{\begin{tabular}[c]{@{}l@{}}Correlation \\ MAE\end{tabular}}} & \multicolumn{1}{l|}{\textbf{\begin{tabular}[c]{@{}l@{}}Correlation \\ MAPE\end{tabular}}} & \multicolumn{1}{l|}{\textbf{\begin{tabular}[c]{@{}l@{}}Categorical \\ column \\ MAE\end{tabular}}} & \multicolumn{1}{l|}{\textbf{\begin{tabular}[c]{@{}l@{}}Categorical \\ column \\ MAPE\end{tabular}}} & \multicolumn{1}{l|}{\textbf{\begin{tabular}[c]{@{}l@{}}Numeric \\ column \\ MAE\end{tabular}}} & \multicolumn{1}{l|}{\textbf{\begin{tabular}[c]{@{}l@{}}Numeric \\ column \\ MAPE\end{tabular}}} & \multicolumn{1}{l|}{\textbf{\begin{tabular}[c]{@{}l@{}}Cramér's V\\ MAE\end{tabular}}} & \multicolumn{1}{l|}{\textbf{\begin{tabular}[c]{@{}l@{}}Cramér's V \\ MAPE\end{tabular}}} & \multicolumn{1}{l|}{\textbf{\begin{tabular}[c]{@{}l@{}}Ease \\ of use \\ ranking\end{tabular}}} \\ \hline
\textbf{training}              & \cellcolor[HTML]{63BE7B}\begin{tabular}[c]{@{}r@{}}1\\      (3.15668)\end{tabular}    & \cellcolor[HTML]{63BE7B}\begin{tabular}[c]{@{}r@{}}1\\      (1)\end{tabular}          & \cellcolor[HTML]{63BE7B}\begin{tabular}[c]{@{}r@{}}1\\      (7)\end{tabular}                           & \cellcolor[HTML]{63BE7B}\begin{tabular}[c]{@{}r@{}}1\\      (0)\end{tabular}                             & \cellcolor[HTML]{63BE7B}\begin{tabular}[c]{@{}r@{}}1\\      (0.31917)\end{tabular}               & \cellcolor[HTML]{63BE7B}\begin{tabular}[c]{@{}r@{}}1\\      (0.23744)\end{tabular}  & \cellcolor[HTML]{63BE7B}\begin{tabular}[c]{@{}r@{}}1\\      (0)\end{tabular}           & \cellcolor[HTML]{63BE7B}\begin{tabular}[c]{@{}r@{}}1\\      (0)\end{tabular}            & \cellcolor[HTML]{63BE7B}\begin{tabular}[c]{@{}r@{}}1\\      (0)\end{tabular}             & \cellcolor[HTML]{63BE7B}\begin{tabular}[c]{@{}r@{}}1\\      (0)\end{tabular}              & \cellcolor[HTML]{63BE7B}\begin{tabular}[c]{@{}r@{}}1\\      (0)\end{tabular}                       & \cellcolor[HTML]{63BE7B}\begin{tabular}[c]{@{}r@{}}1\\      (0)\end{tabular}                        & \cellcolor[HTML]{63BE7B}\begin{tabular}[c]{@{}r@{}}1\\      (0)\end{tabular}                   & \cellcolor[HTML]{63BE7B}\begin{tabular}[c]{@{}r@{}}1\\      (0)\end{tabular}                    & \cellcolor[HTML]{63BE7B}\begin{tabular}[c]{@{}r@{}}1\\      (0)\end{tabular}           & \cellcolor[HTML]{63BE7B}\begin{tabular}[c]{@{}r@{}}1\\      (0)\end{tabular}             & \cellcolor[HTML]{63BE7B}1                                                                       \\ \hline
\textbf{MICE\_PART\_SYN}       & \cellcolor[HTML]{82C77C}\begin{tabular}[c]{@{}r@{}}2\\      (33.93813)\end{tabular}   & \cellcolor[HTML]{82C77C}\begin{tabular}[c]{@{}r@{}}2\\      (33.93479)\end{tabular}   & \cellcolor[HTML]{63BE7B}\begin{tabular}[c]{@{}r@{}}1\\      (7)\end{tabular}                           & \cellcolor[HTML]{DFE282}\begin{tabular}[c]{@{}r@{}}5\\      (1.2)\end{tabular}                           & \cellcolor[HTML]{82C77C}\begin{tabular}[c]{@{}r@{}}2\\      (0.32019)\end{tabular}               & \cellcolor[HTML]{82C77C}\begin{tabular}[c]{@{}r@{}}2\\      (0.23759)\end{tabular}  & \cellcolor[HTML]{82C77C}\begin{tabular}[c]{@{}r@{}}2\\      (0.00061)\end{tabular}     & \cellcolor[HTML]{A1D07E}\begin{tabular}[c]{@{}r@{}}3\\      (0.33907)\end{tabular}      & \cellcolor[HTML]{C0D980}\begin{tabular}[c]{@{}r@{}}4\\      (0.00945)\end{tabular}       & \cellcolor[HTML]{82C77C}\begin{tabular}[c]{@{}r@{}}2\\      (0.27654)\end{tabular}        & \cellcolor[HTML]{DFE282}\begin{tabular}[c]{@{}r@{}}5\\      (0.00185)\end{tabular}                 & \cellcolor[HTML]{A1D07E}\begin{tabular}[c]{@{}r@{}}3\\      (0.02183)\end{tabular}                  & \cellcolor[HTML]{A1D07E}\begin{tabular}[c]{@{}r@{}}3\\      (0.00138)\end{tabular}             & \cellcolor[HTML]{A1D07E}\begin{tabular}[c]{@{}r@{}}3\\      (0.27231)\end{tabular}              & \cellcolor[HTML]{82C77C}\begin{tabular}[c]{@{}r@{}}2\\      (0.01315)\end{tabular}     & \cellcolor[HTML]{DFE282}\begin{tabular}[c]{@{}r@{}}5\\      (0.11307)\end{tabular}       & \cellcolor[HTML]{97CD7E}2                                                                       \\ \hline
\textbf{MICE\_FULL\_SYN}       & \cellcolor[HTML]{A1D07E}\begin{tabular}[c]{@{}r@{}}3\\      (41.03633)\end{tabular}   & \cellcolor[HTML]{A1D07E}\begin{tabular}[c]{@{}r@{}}3\\      (41.52358)\end{tabular}   & \cellcolor[HTML]{63BE7B}\begin{tabular}[c]{@{}r@{}}1\\      (7)\end{tabular}                           & \cellcolor[HTML]{C0D980}\begin{tabular}[c]{@{}r@{}}4\\      (1)\end{tabular}                             & \cellcolor[HTML]{A1D07E}\begin{tabular}[c]{@{}r@{}}3\\      (0.32031)\end{tabular}               & \cellcolor[HTML]{A1D07E}\begin{tabular}[c]{@{}r@{}}3\\      (0.23759)\end{tabular}  & \cellcolor[HTML]{A1D07E}\begin{tabular}[c]{@{}r@{}}3\\      (0.00072)\end{tabular}     & \cellcolor[HTML]{DFE282}\begin{tabular}[c]{@{}r@{}}5\\      (0.44582)\end{tabular}      & \cellcolor[HTML]{DFE282}\begin{tabular}[c]{@{}r@{}}5\\      (0.01031)\end{tabular}       & \cellcolor[HTML]{A1D07E}\begin{tabular}[c]{@{}r@{}}3\\      (0.35079)\end{tabular}        & \cellcolor[HTML]{FFEB84}\begin{tabular}[c]{@{}r@{}}6\\      (0.00232)\end{tabular}                 & \cellcolor[HTML]{C0D980}\begin{tabular}[c]{@{}r@{}}4\\      (0.02677)\end{tabular}                  & \cellcolor[HTML]{C0D980}\begin{tabular}[c]{@{}r@{}}4\\      (0.00176)\end{tabular}             & \cellcolor[HTML]{FFEB84}\begin{tabular}[c]{@{}r@{}}6\\      (0.37186)\end{tabular}              & \cellcolor[HTML]{DFE282}\begin{tabular}[c]{@{}r@{}}5\\      (0.01381)\end{tabular}     & \cellcolor[HTML]{FFEB84}\begin{tabular}[c]{@{}r@{}}6\\      (0.11956)\end{tabular}       & \cellcolor[HTML]{FFEB84}4                                                                       \\ \hline
\textbf{MICE\_VV}              & \cellcolor[HTML]{FED280}\begin{tabular}[c]{@{}r@{}}7\\      (128.96419)\end{tabular}  & \cellcolor[HTML]{FDB87B}\begin{tabular}[c]{@{}r@{}}8\\      (135.24457)\end{tabular}  & \cellcolor[HTML]{FFEB84}\begin{tabular}[c]{@{}r@{}}6\\      (6.8)\end{tabular}                         & \cellcolor[HTML]{82C77C}\begin{tabular}[c]{@{}r@{}}2\\      (0.4)\end{tabular}                           & \cellcolor[HTML]{DFE282}\begin{tabular}[c]{@{}r@{}}5\\      (0.32533)\end{tabular}               & \cellcolor[HTML]{FED280}\begin{tabular}[c]{@{}r@{}}7\\      (0.23833)\end{tabular}  & \cellcolor[HTML]{DFE282}\begin{tabular}[c]{@{}r@{}}5\\      (0.00135)\end{tabular}     & \cellcolor[HTML]{82C77C}\begin{tabular}[c]{@{}r@{}}2\\      (0.20606)\end{tabular}      & \cellcolor[HTML]{FA8370}\begin{tabular}[c]{@{}r@{}}10\\      (0.02837)\end{tabular}      & \cellcolor[HTML]{FDB87B}\begin{tabular}[c]{@{}r@{}}8\\      (1.02168)\end{tabular}        & \cellcolor[HTML]{FED280}\begin{tabular}[c]{@{}r@{}}7\\      (0.00339)\end{tabular}                 & \cellcolor[HTML]{DFE282}\begin{tabular}[c]{@{}r@{}}5\\      (0.02955)\end{tabular}                  & \cellcolor[HTML]{DFE282}\begin{tabular}[c]{@{}r@{}}5\\      (0.00189)\end{tabular}             & \cellcolor[HTML]{82C77C}\begin{tabular}[c]{@{}r@{}}2\\      (0.07052)\end{tabular}              & \cellcolor[HTML]{FB9D75}\begin{tabular}[c]{@{}r@{}}9\\      (0.0372)\end{tabular}      & \cellcolor[HTML]{FED280}\begin{tabular}[c]{@{}r@{}}7\\      (0.22138)\end{tabular}       & \cellcolor[HTML]{97CD7E}2                                                                       \\ \hline
\textbf{MICE\_TABULATOR}       & \cellcolor[HTML]{C0D980}\begin{tabular}[c]{@{}r@{}}4\\      (112.49103)\end{tabular}  & \cellcolor[HTML]{DFE282}\begin{tabular}[c]{@{}r@{}}5\\      (114.40835)\end{tabular}  & \cellcolor[HTML]{FA8370}\begin{tabular}[c]{@{}r@{}}10\\      (6)\end{tabular}                          & \cellcolor[HTML]{A1D07E}\begin{tabular}[c]{@{}r@{}}3\\      (0.8)\end{tabular}                           & \cellcolor[HTML]{C0D980}\begin{tabular}[c]{@{}r@{}}4\\      (0.32393)\end{tabular}               & \cellcolor[HTML]{C0D980}\begin{tabular}[c]{@{}r@{}}4\\      (0.23812)\end{tabular}  & \cellcolor[HTML]{C0D980}\begin{tabular}[c]{@{}r@{}}4\\      (0.00107)\end{tabular}     & \cellcolor[HTML]{FFEB84}\begin{tabular}[c]{@{}r@{}}6\\      (0.4497)\end{tabular}       & \cellcolor[HTML]{FB9D75}\begin{tabular}[c]{@{}r@{}}9\\      (0.0249)\end{tabular}        & \cellcolor[HTML]{FFEB84}\begin{tabular}[c]{@{}r@{}}6\\      (0.67136)\end{tabular}        & \cellcolor[HTML]{82C77C}\begin{tabular}[c]{@{}r@{}}2\\      (0.00081)\end{tabular}                 & \cellcolor[HTML]{82C77C}\begin{tabular}[c]{@{}r@{}}2\\      (0.01022)\end{tabular}                  & \cellcolor[HTML]{82C77C}\begin{tabular}[c]{@{}r@{}}2\\      (0.001)\end{tabular}               & \cellcolor[HTML]{DFE282}\begin{tabular}[c]{@{}r@{}}5\\      (0.36295)\end{tabular}              & \cellcolor[HTML]{FA8370}\begin{tabular}[c]{@{}r@{}}10\\      (0.03845)\end{tabular}    & \cellcolor[HTML]{FDB87B}\begin{tabular}[c]{@{}r@{}}8\\      (0.24778)\end{tabular}       & \cellcolor[HTML]{FFEB84}4                                                                       \\ \hline
\textbf{CTGAN}                 & \cellcolor[HTML]{F8696B}\begin{tabular}[c]{@{}r@{}}11\\      (383.62816)\end{tabular} & \cellcolor[HTML]{F8696B}\begin{tabular}[c]{@{}r@{}}11\\      (393.44698)\end{tabular} & \cellcolor[HTML]{63BE7B}\begin{tabular}[c]{@{}r@{}}1\\      (7)\end{tabular}                           & \cellcolor[HTML]{FED280}\begin{tabular}[c]{@{}r@{}}7\\      (1.4)\end{tabular}                           & \cellcolor[HTML]{FA8370}\begin{tabular}[c]{@{}r@{}}10\\      (0.34049)\end{tabular}              & \cellcolor[HTML]{FA8370}\begin{tabular}[c]{@{}r@{}}10\\      (0.23956)\end{tabular} & \cellcolor[HTML]{FB9D75}\begin{tabular}[c]{@{}r@{}}9\\      (0.00292)\end{tabular}     & \cellcolor[HTML]{FB9D75}\begin{tabular}[c]{@{}r@{}}9\\      (8.9032)\end{tabular}       & \cellcolor[HTML]{FED280}\begin{tabular}[c]{@{}r@{}}7\\      (0.01508)\end{tabular}       & \cellcolor[HTML]{FA8370}\begin{tabular}[c]{@{}r@{}}10\\      (1.38351)\end{tabular}       & \cellcolor[HTML]{FA8370}\begin{tabular}[c]{@{}r@{}}10\\      (0.00678)\end{tabular}                & \cellcolor[HTML]{FA8370}\begin{tabular}[c]{@{}r@{}}10\\      (0.11463)\end{tabular}                 & \cellcolor[HTML]{FDB87B}\begin{tabular}[c]{@{}r@{}}8\\      (0.01167)\end{tabular}             & \cellcolor[HTML]{FB9D75}\begin{tabular}[c]{@{}r@{}}9\\      (42.48769)\end{tabular}             & \cellcolor[HTML]{FED280}\begin{tabular}[c]{@{}r@{}}7\\      (0.02787)\end{tabular}     & \cellcolor[HTML]{FA8370}\begin{tabular}[c]{@{}r@{}}10\\      (0.33897)\end{tabular}      & \cellcolor[HTML]{FFEB84}4                                                                       \\ \hline
\textbf{CTGAN\_MICE}           & \cellcolor[HTML]{DFE282}\begin{tabular}[c]{@{}r@{}}5\\      (113.48165)\end{tabular}  & \cellcolor[HTML]{C0D980}\begin{tabular}[c]{@{}r@{}}4\\      (111.53667)\end{tabular}  & \cellcolor[HTML]{F8696B}\begin{tabular}[c]{@{}r@{}}11\\      (5.6)\end{tabular}                        & \cellcolor[HTML]{FB9D75}\begin{tabular}[c]{@{}r@{}}9\\      (2)\end{tabular}                             & \cellcolor[HTML]{FED280}\begin{tabular}[c]{@{}r@{}}7\\      (0.32572)\end{tabular}               & \cellcolor[HTML]{FFEB84}\begin{tabular}[c]{@{}r@{}}6\\      (0.23832)\end{tabular}  & \cellcolor[HTML]{FDB87B}\begin{tabular}[c]{@{}r@{}}8\\      (0.00253)\end{tabular}     & \cellcolor[HTML]{FED280}\begin{tabular}[c]{@{}r@{}}7\\      (3.96253)\end{tabular}      & \cellcolor[HTML]{82C77C}\begin{tabular}[c]{@{}r@{}}2\\      (0.00543)\end{tabular}       & \cellcolor[HTML]{C0D980}\begin{tabular}[c]{@{}r@{}}4\\      (0.44177)\end{tabular}        & \cellcolor[HTML]{FA8370}\begin{tabular}[c]{@{}r@{}}10\\      (0.00678)\end{tabular}                & \cellcolor[HTML]{FA8370}\begin{tabular}[c]{@{}r@{}}10\\      (0.11463)\end{tabular}                 & \cellcolor[HTML]{FFEB84}\begin{tabular}[c]{@{}r@{}}6\\      (0.00563)\end{tabular}             & \cellcolor[HTML]{FED280}\begin{tabular}[c]{@{}r@{}}7\\      (18.16422)\end{tabular}             & \cellcolor[HTML]{FED280}\begin{tabular}[c]{@{}r@{}}7\\      (0.02787)\end{tabular}     & \cellcolor[HTML]{FA8370}\begin{tabular}[c]{@{}r@{}}10\\      (0.33897)\end{tabular}      & \cellcolor[HTML]{FCB47A}7                                                                       \\ \hline
\textbf{CTGAN\_WITH\_AE}       & \cellcolor[HTML]{FDB87B}\begin{tabular}[c]{@{}r@{}}8\\      (136.08475)\end{tabular}  & \cellcolor[HTML]{FFEB84}\begin{tabular}[c]{@{}r@{}}6\\      (125.10637)\end{tabular}  & \cellcolor[HTML]{FFEB84}\begin{tabular}[c]{@{}r@{}}6\\      (6.8)\end{tabular}                         & \cellcolor[HTML]{FB9D75}\begin{tabular}[c]{@{}r@{}}9\\      (2)\end{tabular}                             & \cellcolor[HTML]{F8696B}\begin{tabular}[c]{@{}r@{}}11\\      (0.34465)\end{tabular}              & \cellcolor[HTML]{F8696B}\begin{tabular}[c]{@{}r@{}}11\\      (0.2422)\end{tabular}  & \cellcolor[HTML]{FED280}\begin{tabular}[c]{@{}r@{}}7\\      (0.00244)\end{tabular}     & \cellcolor[HTML]{FA8370}\begin{tabular}[c]{@{}r@{}}10\\      (17.68936)\end{tabular}    & \cellcolor[HTML]{FDB87B}\begin{tabular}[c]{@{}r@{}}8\\      (0.01736)\end{tabular}       & \cellcolor[HTML]{F8696B}\begin{tabular}[c]{@{}r@{}}11\\      (1.74036)\end{tabular}       & \cellcolor[HTML]{A1D07E}\begin{tabular}[c]{@{}r@{}}3\\      (0.0016)\end{tabular}                  & \cellcolor[HTML]{FFEB84}\begin{tabular}[c]{@{}r@{}}6\\      (0.05566)\end{tabular}                  & \cellcolor[HTML]{FA8370}\begin{tabular}[c]{@{}r@{}}9\\      (0.01395)\end{tabular}             & \cellcolor[HTML]{FA8370}\begin{tabular}[c]{@{}r@{}}10\\      (83.98836)\end{tabular}            & \cellcolor[HTML]{A1D07E}\begin{tabular}[c]{@{}r@{}}3\\      (0.01338)\end{tabular}     & \cellcolor[HTML]{A1D07E}\begin{tabular}[c]{@{}r@{}}3\\      (0.10637)\end{tabular}       & \cellcolor[HTML]{FCB47A}7                                                                       \\ \hline
\textbf{CTGAN\_WITH\_AE\_MICE} & \cellcolor[HTML]{FA8370}\begin{tabular}[c]{@{}r@{}}10\\      (153.73562)\end{tabular} & \cellcolor[HTML]{FB9D75}\begin{tabular}[c]{@{}r@{}}9\\      (143.03862)\end{tabular}  & \cellcolor[HTML]{FFEB84}\begin{tabular}[c]{@{}r@{}}6\\      (6.8)\end{tabular}                         & \cellcolor[HTML]{FED280}\begin{tabular}[c]{@{}r@{}}7\\      (1.4)\end{tabular}                           & \cellcolor[HTML]{FB9D75}\begin{tabular}[c]{@{}r@{}}9\\      (0.33246)\end{tabular}               & \cellcolor[HTML]{FB9D75}\begin{tabular}[c]{@{}r@{}}9\\      (0.23933)\end{tabular}  & \cellcolor[HTML]{FFEB84}\begin{tabular}[c]{@{}r@{}}6\\      (0.00199)\end{tabular}     & \cellcolor[HTML]{FDB87B}\begin{tabular}[c]{@{}r@{}}8\\      (8.25317)\end{tabular}      & \cellcolor[HTML]{A1D07E}\begin{tabular}[c]{@{}r@{}}3\\      (0.00569)\end{tabular}       & \cellcolor[HTML]{FED280}\begin{tabular}[c]{@{}r@{}}7\\      (0.71934)\end{tabular}        & \cellcolor[HTML]{A1D07E}\begin{tabular}[c]{@{}r@{}}3\\      (0.0016)\end{tabular}                  & \cellcolor[HTML]{FFEB84}\begin{tabular}[c]{@{}r@{}}6\\      (0.05566)\end{tabular}                  & \cellcolor[HTML]{FED280}\begin{tabular}[c]{@{}r@{}}7\\      (0.00703)\end{tabular}             & \cellcolor[HTML]{FDB87B}\begin{tabular}[c]{@{}r@{}}8\\      (38.72062)\end{tabular}             & \cellcolor[HTML]{A1D07E}\begin{tabular}[c]{@{}r@{}}3\\      (0.01338)\end{tabular}     & \cellcolor[HTML]{A1D07E}\begin{tabular}[c]{@{}r@{}}3\\      (0.10637)\end{tabular}       & \cellcolor[HTML]{FA8F73}9                                                                       \\ \hline
\textbf{VAE\_JAMOTTON}         & \cellcolor[HTML]{FFEB84}\begin{tabular}[c]{@{}r@{}}6\\      (126.94602)\end{tabular}  & \cellcolor[HTML]{FED280}\begin{tabular}[c]{@{}r@{}}7\\      (129.11846)\end{tabular}  & \cellcolor[HTML]{FB9D75}\begin{tabular}[c]{@{}r@{}}9\\      (6.6)\end{tabular}                         & \cellcolor[HTML]{DFE282}\begin{tabular}[c]{@{}r@{}}5\\      (1.2)\end{tabular}                           & \cellcolor[HTML]{FFEB84}\begin{tabular}[c]{@{}r@{}}6\\      (0.32534)\end{tabular}               & \cellcolor[HTML]{DFE282}\begin{tabular}[c]{@{}r@{}}5\\      (0.23828)\end{tabular}  & \cellcolor[HTML]{FA8370}\begin{tabular}[c]{@{}r@{}}10\\      (0.00303)\end{tabular}    & \cellcolor[HTML]{C0D980}\begin{tabular}[c]{@{}r@{}}4\\      (0.40646)\end{tabular}      & \cellcolor[HTML]{F8696B}\begin{tabular}[c]{@{}r@{}}11\\      (0.03651)\end{tabular}      & \cellcolor[HTML]{FB9D75}\begin{tabular}[c]{@{}r@{}}9\\      (1.07215)\end{tabular}        & \cellcolor[HTML]{FDB87B}\begin{tabular}[c]{@{}r@{}}8\\      (0.00429)\end{tabular}                 & \cellcolor[HTML]{FDB87B}\begin{tabular}[c]{@{}r@{}}8\\      (0.0809)\end{tabular}                   & \cellcolor[HTML]{FA8370}\begin{tabular}[c]{@{}r@{}}10\\      (0.01424)\end{tabular}            & \cellcolor[HTML]{C0D980}\begin{tabular}[c]{@{}r@{}}4\\      (0.32234)\end{tabular}              & \cellcolor[HTML]{F8696B}\begin{tabular}[c]{@{}r@{}}11\\      (0.04399)\end{tabular}    & \cellcolor[HTML]{FB9D75}\begin{tabular}[c]{@{}r@{}}9\\      (0.33535)\end{tabular}       & \cellcolor[HTML]{F97C6F}10                                                                      \\ \hline
\textbf{COTE\_MC\_WGAN\_GP}    & \cellcolor[HTML]{FB9D75}\begin{tabular}[c]{@{}r@{}}9\\      (152.74518)\end{tabular}  & \cellcolor[HTML]{FA8370}\begin{tabular}[c]{@{}r@{}}10\\      (155.80504)\end{tabular} & \cellcolor[HTML]{63BE7B}\begin{tabular}[c]{@{}r@{}}1\\      (7)\end{tabular}                           & \cellcolor[HTML]{FB9D75}\begin{tabular}[c]{@{}r@{}}9\\      (2)\end{tabular}                             & \cellcolor[HTML]{FDB87B}\begin{tabular}[c]{@{}r@{}}8\\      (0.32906)\end{tabular}               & \cellcolor[HTML]{FDB87B}\begin{tabular}[c]{@{}r@{}}8\\      (0.23892)\end{tabular}  & \cellcolor[HTML]{F8696B}\begin{tabular}[c]{@{}r@{}}11\\      (0.00412)\end{tabular}    & \cellcolor[HTML]{F8696B}\begin{tabular}[c]{@{}r@{}}11\\      (82.61544)\end{tabular}    & \cellcolor[HTML]{FFEB84}\begin{tabular}[c]{@{}r@{}}6\\      (0.01052)\end{tabular}       & \cellcolor[HTML]{DFE282}\begin{tabular}[c]{@{}r@{}}5\\      (0.48478)\end{tabular}        & \cellcolor[HTML]{FB9D75}\begin{tabular}[c]{@{}r@{}}9\\      (0.00444)\end{tabular}                 & \cellcolor[HTML]{FB9D75}\begin{tabular}[c]{@{}r@{}}9\\      (0.08287)\end{tabular}                  & \cellcolor[HTML]{F8696B}\begin{tabular}[c]{@{}r@{}}11\\      (0.02235)\end{tabular}            & \cellcolor[HTML]{F8696B}\begin{tabular}[c]{@{}r@{}}11\\      (445.05061)\end{tabular}           & \cellcolor[HTML]{FFEB84}\begin{tabular}[c]{@{}r@{}}6\\      (0.01466)\end{tabular}     & \cellcolor[HTML]{82C77C}\begin{tabular}[c]{@{}r@{}}2\\      (0.09655)\end{tabular}       & \cellcolor[HTML]{F8696B}11                                                                      \\ \hline
\end{tabular}%
}
\caption{Summary of evaluation results across all tested algorithms when the variable dependence is based on linear formula \eqref{eval:linear_formula}. \blueparagraph{Meaning of those metrics is summarized in Appendix \ref{app:summary_metrics}.}}
\label{results:linear_table}
\end{table}

Looking at Table \ref{results:linear_table}, we can see that in no case does any method outperform training data. The differences between training data and synthetic data can be quite large for example for metric $M_1$ or $M_2$. However, {the comparatively large coefficient discrepancies are accompanied by only small descriptive differences in test-set Poisson deviance and RMSE}.

{
The coefficient-based measures ($M_1$ and $M_2$) and the predictive measures (Poisson deviance and RMSE) evaluate different aspects of the synthetic data quality. $M_1$ and $M_2$ measure coefficient recovery, whereas Poisson deviance and RMSE evaluate prediction accuracy. Because coefficient discrepancies in $M_1$ and $M_2$ are normalized by the relatively small estimated coefficient MSEs, even modest absolute discrepancies can produce large metric values. Moreover, predictive performance depends on the combined effect of the coefficient discrepancies on the linear predictor. These discrepancies may partly offset each other or influence infrequent categories and/or interactions. Thus, good predictive performance does not necessarily imply good preservation of the underlying regression structure.
}

We can see that the best-performing synthetic data generation methods are \texttt{MICE\_PART\_SYN} and \texttt{MICE\_FULL\_SYN}. They show almost the same accuracy in terms of goodness-of-fit metrics, but \texttt{MICE\_PART\_SYN} is better when it comes to the accuracy of model coefficients. The worst performing methods are  \texttt{CTGAN}, \texttt{CTGAN\_WITH\_AE\_MICE}, \texttt{COTE\_MC\_WGAN\_GP} and \texttt{CTGAN\_WITH\_AE}. These methods have higher model coefficient metric values and worse accuracy in model predictions.

Almost all methods show consistency in the correct terms using the variable selection method. However, for all methods, the variable selection included additional variables not present in the true model. Here, the worst-performing method was \texttt{CTGAN\_MICE}, where automatic variable selection failed to include 1.4 correct variables, on average, and included, on average, 2 additional incorrect variables (the worst possible result). The best variable selection results were achieved by models trained on data using the \texttt{MICE\_PART\_SYN} and \texttt{MICE\_FULL\_SYN} methods, which always selected the correct variables and included only 1 additional incorrect variable on average. Notably, \texttt{MICE\_VV} included, on average, only $0.4$ incorrect variables.

Similar observations can be seen for data-specific metrics. Here again, \texttt{MICE\_PART\_SYN} overall performs the best across all data metrics. From Table \ref{results:linear_table}, we conclude that different methods have different niches. For example, it seems \texttt{MICE\_TABULATOR} is much better at categorical column generation {when it comes to proportion of categorical data}, followed closely by \texttt{CTGAN\_WITH\_AE} and \texttt{CTGAN\_WITH\_AE\_MICE} (since they share categorical variables). {However, we can see that based on  Cramér's V correlation errors, the \texttt{MICE\_TABULATOR} is the worst at preserving the dependence structure for categorical variables, while \texttt{CTGAN\_WITH\_AE} and \texttt{CTGAN\_WITH\_AE\_MICE} are performing well in that regard.}  This suggests that the use of autoencoders improves the generation of categorical columns. Additionally, we observe that adding MICE to CTGAN-based methods slightly improves the generation of numeric columns, as both \texttt{CTGAN\_MICE} and \texttt{CTGAN\_WITH\_AE\_MICE} perform better than their non-augmented counterparts. WGAN by \cite{Cote2020} shows the worst overall performance on data-specific metrics.

We also qualitatively assess the usability of each tested generator based on three criteria: intuitiveness (conceptual clarity), operational complexity (training and usage pipelines), and deployment overhead (environment setup, debugging, and code adaptation).

We find MICE-based methods to be the most accessible, primarily due to their streamlined implementation in the \textsf{R} package \texttt{mice}, which requires minimal setup beyond standard installation. Notably, \texttt{mice} outputs by default synthetic data without exporting the trained imputation models.

In contrast, CTGAN-based methods require additional setup time and more extensive data pre-processing (e.g., matrix fragmentation, type casting). For practical reasons, these experiments were conducted in Python, as this environment was more compatible with our computational workflow than the available \textsf{R} implementation.

Incorporating AEs, while conceptually straightforward, increases implementation complexity due to the need for supplementary data preparation, neural network specification, and post-hoc clustering to recover categorical labels.

Finally, the custom implementations kindly provided by the authors of \cite{Jamotton2024} and \cite{Cote2020} represent highly specialized, advanced architectures. Consequently, integrating these custom-built models into a standardized, comparative computational framework naturally necessitates a higher degree of code adaptation and environment configuration.

\begin{table}[!ht]
\centering
\resizebox{\textwidth}{!}{%
\begin{tabular}{|l|r|r|r|r|r|r|r|r|r|r|r|r|r|r|r|r|r|l|}
\hline
\textbf{Generation   method}   & \multicolumn{1}{l|}{\textbf{M1}}                                                     & \multicolumn{1}{l|}{\textbf{M2}}                                                      & \multicolumn{1}{l|}{\textbf{\begin{tabular}[c]{@{}l@{}}Correct \\ variables \\ selected\end{tabular}}} & \multicolumn{1}{l|}{\textbf{\begin{tabular}[c]{@{}l@{}}Incorrect \\ variables \\ selected\end{tabular}}} & \multicolumn{1}{l|}{\textbf{\begin{tabular}[c]{@{}l@{}}Missing \\ main effect \\ variable\end{tabular}}} & \multicolumn{1}{l|}{\textbf{\begin{tabular}[c]{@{}l@{}}Fit \\ Poisson \\ deviance\end{tabular}}} & \multicolumn{1}{l|}{\textbf{\begin{tabular}[c]{@{}l@{}}Fit\\ RMSE\end{tabular}}}    & \multicolumn{1}{l|}{\textbf{\begin{tabular}[c]{@{}l@{}}Pair-wise \\ MAE\end{tabular}}} & \multicolumn{1}{l|}{\textbf{\begin{tabular}[c]{@{}l@{}}Pair-Wise \\ MAPE\end{tabular}}} & \multicolumn{1}{l|}{\textbf{\begin{tabular}[c]{@{}l@{}}Correlation \\ MAE\end{tabular}}} & \multicolumn{1}{l|}{\textbf{\begin{tabular}[c]{@{}l@{}}Correlation \\ MAPE\end{tabular}}} & \multicolumn{1}{l|}{\textbf{\begin{tabular}[c]{@{}l@{}}Categorical \\ column \\ MAE\end{tabular}}} & \multicolumn{1}{l|}{\textbf{\begin{tabular}[c]{@{}l@{}}Categorical \\ column \\ MAPE\end{tabular}}} & \multicolumn{1}{l|}{\textbf{\begin{tabular}[c]{@{}l@{}}Numeric \\ column \\ MAE\end{tabular}}} & \multicolumn{1}{l|}{\textbf{\begin{tabular}[c]{@{}l@{}}Numeric \\ column \\ MAPE\end{tabular}}} & \multicolumn{1}{l|}{\textbf{\begin{tabular}[c]{@{}l@{}}Cramér's V\\ MAE\end{tabular}}} & \multicolumn{1}{l|}{\textbf{\begin{tabular}[c]{@{}l@{}}Cramér's V \\ MAPE\end{tabular}}} & \textbf{\begin{tabular}[c]{@{}l@{}}Ease \\ of use \\ ranking\end{tabular}} \\ \hline
\textbf{training}              & \cellcolor[HTML]{63BE7B}\begin{tabular}[c]{@{}r@{}}1\\      (0.86047)\end{tabular}   & \cellcolor[HTML]{63BE7B}\begin{tabular}[c]{@{}r@{}}1\\      (1)\end{tabular}          & \cellcolor[HTML]{63BE7B}\begin{tabular}[c]{@{}r@{}}1\\      (16)\end{tabular}                          & \cellcolor[HTML]{63BE7B}\begin{tabular}[c]{@{}r@{}}1\\      (0)\end{tabular}                             & \cellcolor[HTML]{63BE7B}\begin{tabular}[c]{@{}r@{}}1\\      (0)\end{tabular}                             & \cellcolor[HTML]{63BE7B}\begin{tabular}[c]{@{}r@{}}1\\      (0.3394)\end{tabular}                & \cellcolor[HTML]{63BE7B}\begin{tabular}[c]{@{}r@{}}1\\      (0.25286)\end{tabular}  & \cellcolor[HTML]{63BE7B}\begin{tabular}[c]{@{}r@{}}1\\      (0)\end{tabular}           & \cellcolor[HTML]{63BE7B}\begin{tabular}[c]{@{}r@{}}1\\      (0)\end{tabular}            & \cellcolor[HTML]{63BE7B}\begin{tabular}[c]{@{}r@{}}1\\      (0)\end{tabular}             & \cellcolor[HTML]{63BE7B}\begin{tabular}[c]{@{}r@{}}1\\      (0)\end{tabular}              & \cellcolor[HTML]{63BE7B}\begin{tabular}[c]{@{}r@{}}1\\      (0)\end{tabular}                       & \cellcolor[HTML]{63BE7B}\begin{tabular}[c]{@{}r@{}}1\\      (0)\end{tabular}                        & \cellcolor[HTML]{63BE7B}\begin{tabular}[c]{@{}r@{}}1\\      (0)\end{tabular}                   & \cellcolor[HTML]{63BE7B}\begin{tabular}[c]{@{}r@{}}1\\      (0)\end{tabular}                    & \cellcolor[HTML]{63BE7B}\begin{tabular}[c]{@{}r@{}}1\\      (0)\end{tabular}           & \cellcolor[HTML]{63BE7B}\begin{tabular}[c]{@{}r@{}}1\\      (0)\end{tabular}             & \cellcolor[HTML]{63BE7B}1                                                  \\ \hline
\textbf{MICE\_PART\_SYN}       & \cellcolor[HTML]{82C77C}\begin{tabular}[c]{@{}r@{}}2\\      (13.62087)\end{tabular}  & \cellcolor[HTML]{82C77C}\begin{tabular}[c]{@{}r@{}}2\\      (17.6173)\end{tabular}    & \cellcolor[HTML]{A1D07E}\begin{tabular}[c]{@{}r@{}}3\\      (15.4)\end{tabular}                        & \cellcolor[HTML]{FED280}\begin{tabular}[c]{@{}r@{}}7\\      (1.8)\end{tabular}                           & \cellcolor[HTML]{63BE7B}\begin{tabular}[c]{@{}r@{}}1\\      (0)\end{tabular}                             & \cellcolor[HTML]{82C77C}\begin{tabular}[c]{@{}r@{}}2\\      (0.34127)\end{tabular}               & \cellcolor[HTML]{82C77C}\begin{tabular}[c]{@{}r@{}}2\\      (0.25318)\end{tabular}  & \cellcolor[HTML]{82C77C}\begin{tabular}[c]{@{}r@{}}2\\      (0.00061)\end{tabular}     & \cellcolor[HTML]{C0D980}\begin{tabular}[c]{@{}r@{}}4\\      (0.34177)\end{tabular}      & \cellcolor[HTML]{C0D980}\begin{tabular}[c]{@{}r@{}}4\\      (0.00889)\end{tabular}       & \cellcolor[HTML]{82C77C}\begin{tabular}[c]{@{}r@{}}2\\      (0.30741)\end{tabular}        & \cellcolor[HTML]{A1D07E}\begin{tabular}[c]{@{}r@{}}3\\      (0.002)\end{tabular}                   & \cellcolor[HTML]{A1D07E}\begin{tabular}[c]{@{}r@{}}3\\      (0.02169)\end{tabular}                  & \cellcolor[HTML]{A1D07E}\begin{tabular}[c]{@{}r@{}}3\\      (0.00139)\end{tabular}             & \cellcolor[HTML]{C0D980}\begin{tabular}[c]{@{}r@{}}4\\      (0.29525)\end{tabular}              & \cellcolor[HTML]{C0D980}\begin{tabular}[c]{@{}r@{}}4\\      (0.01313)\end{tabular}     & \cellcolor[HTML]{DFE282}\begin{tabular}[c]{@{}r@{}}5\\      (0.11139)\end{tabular}       & \cellcolor[HTML]{85C87D}2                                                  \\ \hline
\textbf{MICE\_FULL\_SYN}       & \cellcolor[HTML]{A1D07E}\begin{tabular}[c]{@{}r@{}}3\\      (16.69526)\end{tabular}  & \cellcolor[HTML]{A1D07E}\begin{tabular}[c]{@{}r@{}}3\\      (21.04755)\end{tabular}   & \cellcolor[HTML]{82C77C}\begin{tabular}[c]{@{}r@{}}2\\      (15.6)\end{tabular}                        & \cellcolor[HTML]{DFE282}\begin{tabular}[c]{@{}r@{}}5\\      (1.6)\end{tabular}                           & \cellcolor[HTML]{FFEB84}\begin{tabular}[c]{@{}r@{}}6\\      (0.4)\end{tabular}                           & \cellcolor[HTML]{A1D07E}\begin{tabular}[c]{@{}r@{}}3\\      (0.34166)\end{tabular}               & \cellcolor[HTML]{A1D07E}\begin{tabular}[c]{@{}r@{}}3\\      (0.25323)\end{tabular}  & \cellcolor[HTML]{A1D07E}\begin{tabular}[c]{@{}r@{}}3\\      (0.00073)\end{tabular}     & \cellcolor[HTML]{DFE282}\begin{tabular}[c]{@{}r@{}}5\\      (0.45389)\end{tabular}      & \cellcolor[HTML]{DFE282}\begin{tabular}[c]{@{}r@{}}5\\      (0.00967)\end{tabular}       & \cellcolor[HTML]{C0D980}\begin{tabular}[c]{@{}r@{}}4\\      (0.38692)\end{tabular}        & \cellcolor[HTML]{C0D980}\begin{tabular}[c]{@{}r@{}}4\\      (0.00242)\end{tabular}                 & \cellcolor[HTML]{C0D980}\begin{tabular}[c]{@{}r@{}}4\\      (0.02565)\end{tabular}                  & \cellcolor[HTML]{C0D980}\begin{tabular}[c]{@{}r@{}}4\\      (0.00175)\end{tabular}             & \cellcolor[HTML]{FFEB84}\begin{tabular}[c]{@{}r@{}}6\\      (0.41926)\end{tabular}              & \cellcolor[HTML]{DFE282}\begin{tabular}[c]{@{}r@{}}5\\      (0.01413)\end{tabular}     & \cellcolor[HTML]{FFEB84}\begin{tabular}[c]{@{}r@{}}6\\      (0.11972)\end{tabular}       & \multicolumn{1}{r|}{\cellcolor[HTML]{C0D980}4}                             \\ \hline
\textbf{MICE\_VV}              & \cellcolor[HTML]{FB9D75}\begin{tabular}[c]{@{}r@{}}9\\      (60.33524)\end{tabular}  & \cellcolor[HTML]{FB9D75}\begin{tabular}[c]{@{}r@{}}9\\      (66.83869)\end{tabular}   & \cellcolor[HTML]{FFEB84}\begin{tabular}[c]{@{}r@{}}6\\      (10.2)\end{tabular}                        & \cellcolor[HTML]{A1D07E}\begin{tabular}[c]{@{}r@{}}3\\      (0.8)\end{tabular}                           & \cellcolor[HTML]{FFEB84}\begin{tabular}[c]{@{}r@{}}6\\      (0.4)\end{tabular}                           & \cellcolor[HTML]{FFEB84}\begin{tabular}[c]{@{}r@{}}6\\      (0.35077)\end{tabular}               & \cellcolor[HTML]{FFEB84}\begin{tabular}[c]{@{}r@{}}6\\      (0.25461)\end{tabular}  & \cellcolor[HTML]{DFE282}\begin{tabular}[c]{@{}r@{}}5\\      (0.00143)\end{tabular}     & \cellcolor[HTML]{82C77C}\begin{tabular}[c]{@{}r@{}}2\\      (0.20707)\end{tabular}      & \cellcolor[HTML]{F8696B}\begin{tabular}[c]{@{}r@{}}11\\      (0.02914)\end{tabular}      & \cellcolor[HTML]{FDB87B}\begin{tabular}[c]{@{}r@{}}8\\      (1.13441)\end{tabular}        & \cellcolor[HTML]{DFE282}\begin{tabular}[c]{@{}r@{}}5\\      (0.00377)\end{tabular}                 & \cellcolor[HTML]{DFE282}\begin{tabular}[c]{@{}r@{}}5\\      (0.0319)\end{tabular}                   & \cellcolor[HTML]{DFE282}\begin{tabular}[c]{@{}r@{}}5\\      (0.00183)\end{tabular}             & \cellcolor[HTML]{82C77C}\begin{tabular}[c]{@{}r@{}}2\\      (0.05385)\end{tabular}              & \cellcolor[HTML]{FA8370}\begin{tabular}[c]{@{}r@{}}10\\      (0.03828)\end{tabular}    & \cellcolor[HTML]{FED280}\begin{tabular}[c]{@{}r@{}}7\\      (0.22635)\end{tabular}       & \cellcolor[HTML]{85C87D}2                                                  \\ \hline
\textbf{MICE\_TABULATOR}       & \cellcolor[HTML]{FDB87B}\begin{tabular}[c]{@{}r@{}}8\\      (49.90837)\end{tabular}  & \cellcolor[HTML]{FDB87B}\begin{tabular}[c]{@{}r@{}}8\\      (55.7844)\end{tabular}    & \cellcolor[HTML]{FDB87B}\begin{tabular}[c]{@{}r@{}}8\\      (9.4)\end{tabular}                         & \cellcolor[HTML]{FDB87B}\begin{tabular}[c]{@{}r@{}}8\\      (2)\end{tabular}                             & \cellcolor[HTML]{F8696B}\begin{tabular}[c]{@{}r@{}}11\\      (1.2)\end{tabular}                          & \cellcolor[HTML]{DFE282}\begin{tabular}[c]{@{}r@{}}5\\      (0.34909)\end{tabular}               & \cellcolor[HTML]{DFE282}\begin{tabular}[c]{@{}r@{}}5\\      (0.25434)\end{tabular}  & \cellcolor[HTML]{C0D980}\begin{tabular}[c]{@{}r@{}}4\\      (0.00111)\end{tabular}     & \cellcolor[HTML]{FFEB84}\begin{tabular}[c]{@{}r@{}}6\\      (0.4616)\end{tabular}       & \cellcolor[HTML]{FA8370}\begin{tabular}[c]{@{}r@{}}10\\      (0.02481)\end{tabular}      & \cellcolor[HTML]{FFEB84}\begin{tabular}[c]{@{}r@{}}6\\      (0.49791)\end{tabular}        & \cellcolor[HTML]{82C77C}\begin{tabular}[c]{@{}r@{}}2\\      (0.00071)\end{tabular}                 & \cellcolor[HTML]{82C77C}\begin{tabular}[c]{@{}r@{}}2\\      (0.00955)\end{tabular}                  & \cellcolor[HTML]{82C77C}\begin{tabular}[c]{@{}r@{}}2\\      (0.00102)\end{tabular}             & \cellcolor[HTML]{DFE282}\begin{tabular}[c]{@{}r@{}}5\\      (0.3482)\end{tabular}               & \cellcolor[HTML]{F8696B}\begin{tabular}[c]{@{}r@{}}11\\      (0.03896)\end{tabular}    & \cellcolor[HTML]{FDB87B}\begin{tabular}[c]{@{}r@{}}8\\      (0.24946)\end{tabular}       & \cellcolor[HTML]{CBDC81}4                                                  \\ \hline
\textbf{CTGAN}                 & \cellcolor[HTML]{F8696B}\begin{tabular}[c]{@{}r@{}}11\\      (97.22979)\end{tabular} & \cellcolor[HTML]{FA8370}\begin{tabular}[c]{@{}r@{}}10\\      (97.45528)\end{tabular}  & \cellcolor[HTML]{C0D980}\begin{tabular}[c]{@{}r@{}}4\\      (12)\end{tabular}                          & \cellcolor[HTML]{DFE282}\begin{tabular}[c]{@{}r@{}}5\\      (1.6)\end{tabular}                           & \cellcolor[HTML]{A1D07E}\begin{tabular}[c]{@{}r@{}}3\\      (0.2)\end{tabular}                           & \cellcolor[HTML]{F8696B}\begin{tabular}[c]{@{}r@{}}11\\      (0.37112)\end{tabular}              & \cellcolor[HTML]{FA8370}\begin{tabular}[c]{@{}r@{}}10\\      (0.25564)\end{tabular} & \cellcolor[HTML]{FA8370}\begin{tabular}[c]{@{}r@{}}10\\      (0.00278)\end{tabular}    & \cellcolor[HTML]{FDB87B}\begin{tabular}[c]{@{}r@{}}8\\      (6.29387)\end{tabular}      & \cellcolor[HTML]{FED280}\begin{tabular}[c]{@{}r@{}}7\\      (0.01237)\end{tabular}       & \cellcolor[HTML]{F8696B}\begin{tabular}[c]{@{}r@{}}11\\      (2.07503)\end{tabular}       & \cellcolor[HTML]{FA8370}\begin{tabular}[c]{@{}r@{}}10\\      (0.0101)\end{tabular}                 & \cellcolor[HTML]{FA8370}\begin{tabular}[c]{@{}r@{}}10\\      (0.09023)\end{tabular}                 & \cellcolor[HTML]{FA8370}\begin{tabular}[c]{@{}r@{}}10\\      (0.01193)\end{tabular}            & \cellcolor[HTML]{FDB87B}\begin{tabular}[c]{@{}r@{}}8\\      (30.54492)\end{tabular}             & \cellcolor[HTML]{FED280}\begin{tabular}[c]{@{}r@{}}7\\      (0.02553)\end{tabular}     & \cellcolor[HTML]{FA8370}\begin{tabular}[c]{@{}r@{}}10\\      (0.34411)\end{tabular}      & \cellcolor[HTML]{CBDC81}4                                                  \\ \hline
\textbf{CTGAN\_MICE}           & \cellcolor[HTML]{FED280}\begin{tabular}[c]{@{}r@{}}7\\      (49.87283)\end{tabular}  & \cellcolor[HTML]{FFEB84}\begin{tabular}[c]{@{}r@{}}6\\      (53.21122)\end{tabular}   & \cellcolor[HTML]{FB9D75}\begin{tabular}[c]{@{}r@{}}9\\      (8.4)\end{tabular}                         & \cellcolor[HTML]{FDB87B}\begin{tabular}[c]{@{}r@{}}8\\      (2)\end{tabular}                             & \cellcolor[HTML]{FFEB84}\begin{tabular}[c]{@{}r@{}}6\\      (0.4)\end{tabular}                           & \cellcolor[HTML]{FB9D75}\begin{tabular}[c]{@{}r@{}}9\\      (0.35463)\end{tabular}               & \cellcolor[HTML]{FB9D75}\begin{tabular}[c]{@{}r@{}}9\\      (0.25515)\end{tabular}  & \cellcolor[HTML]{FB9D75}\begin{tabular}[c]{@{}r@{}}9\\      (0.00243)\end{tabular}     & \cellcolor[HTML]{FED280}\begin{tabular}[c]{@{}r@{}}7\\      (2.93129)\end{tabular}      & \cellcolor[HTML]{82C77C}\begin{tabular}[c]{@{}r@{}}2\\      (0.00479)\end{tabular}       & \cellcolor[HTML]{A1D07E}\begin{tabular}[c]{@{}r@{}}3\\      (0.38592)\end{tabular}        & \cellcolor[HTML]{FA8370}\begin{tabular}[c]{@{}r@{}}10\\      (0.0101)\end{tabular}                 & \cellcolor[HTML]{FA8370}\begin{tabular}[c]{@{}r@{}}10\\      (0.09023)\end{tabular}                 & \cellcolor[HTML]{FED280}\begin{tabular}[c]{@{}r@{}}7\\      (0.005)\end{tabular}               & \cellcolor[HTML]{FED280}\begin{tabular}[c]{@{}r@{}}7\\      (13.22874)\end{tabular}             & \cellcolor[HTML]{FED280}\begin{tabular}[c]{@{}r@{}}7\\      (0.02553)\end{tabular}     & \cellcolor[HTML]{FA8370}\begin{tabular}[c]{@{}r@{}}10\\      (0.34411)\end{tabular}      & \cellcolor[HTML]{FEC87E}7                                                  \\ \hline
\textbf{CTGAN\_WITH\_AE}       & \cellcolor[HTML]{C0D980}\begin{tabular}[c]{@{}r@{}}4\\      (41.8)\end{tabular}      & \cellcolor[HTML]{C0D980}\begin{tabular}[c]{@{}r@{}}4\\      (46.39565)\end{tabular}   & \cellcolor[HTML]{FED280}\begin{tabular}[c]{@{}r@{}}7\\      (10)\end{tabular}                          & \cellcolor[HTML]{C0D980}\begin{tabular}[c]{@{}r@{}}4\\      (1.2)\end{tabular}                           & \cellcolor[HTML]{A1D07E}\begin{tabular}[c]{@{}r@{}}3\\      (0.2)\end{tabular}                           & \cellcolor[HTML]{C0D980}\begin{tabular}[c]{@{}r@{}}4\\      (0.34617)\end{tabular}               & \cellcolor[HTML]{C0D980}\begin{tabular}[c]{@{}r@{}}4\\      (0.2539)\end{tabular}   & \cellcolor[HTML]{FDB87B}\begin{tabular}[c]{@{}r@{}}8\\      (0.00207)\end{tabular}     & \cellcolor[HTML]{FA8370}\begin{tabular}[c]{@{}r@{}}10\\      (19.89066)\end{tabular}    & \cellcolor[HTML]{FDB87B}\begin{tabular}[c]{@{}r@{}}8\\      (0.02117)\end{tabular}       & \cellcolor[HTML]{FA8370}\begin{tabular}[c]{@{}r@{}}10\\      (1.79309)\end{tabular}       & \cellcolor[HTML]{FDB87B}\begin{tabular}[c]{@{}r@{}}8\\      (0.00529)\end{tabular}                 & \cellcolor[HTML]{FFEB84}\begin{tabular}[c]{@{}r@{}}6\\      (0.06383)\end{tabular}                  & \cellcolor[HTML]{FB9D75}\begin{tabular}[c]{@{}r@{}}9\\      (0.00852)\end{tabular}             & \cellcolor[HTML]{FA8370}\begin{tabular}[c]{@{}r@{}}10\\      (103.25122)\end{tabular}           & \cellcolor[HTML]{82C77C}\begin{tabular}[c]{@{}r@{}}2\\      (0.0114)\end{tabular}      & \cellcolor[HTML]{82C77C}\begin{tabular}[c]{@{}r@{}}2\\      (0.10076)\end{tabular}       & \cellcolor[HTML]{FEC87E}7                                                  \\ \hline
\textbf{CTGAN\_WITH\_AE\_MICE} & \cellcolor[HTML]{DFE282}\begin{tabular}[c]{@{}r@{}}5\\      (47.65917)\end{tabular}  & \cellcolor[HTML]{DFE282}\begin{tabular}[c]{@{}r@{}}5\\      (50.98831)\end{tabular}   & \cellcolor[HTML]{FA8370}\begin{tabular}[c]{@{}r@{}}10\\      (8)\end{tabular}                          & \cellcolor[HTML]{FDB87B}\begin{tabular}[c]{@{}r@{}}8\\      (2)\end{tabular}                             & \cellcolor[HTML]{FFEB84}\begin{tabular}[c]{@{}r@{}}6\\      (0.4)\end{tabular}                           & \cellcolor[HTML]{FED280}\begin{tabular}[c]{@{}r@{}}7\\      (0.35251)\end{tabular}               & \cellcolor[HTML]{FED280}\begin{tabular}[c]{@{}r@{}}7\\      (0.25483)\end{tabular}  & \cellcolor[HTML]{FFEB84}\begin{tabular}[c]{@{}r@{}}6\\      (0.00194)\end{tabular}     & \cellcolor[HTML]{FB9D75}\begin{tabular}[c]{@{}r@{}}9\\      (8.69745)\end{tabular}      & \cellcolor[HTML]{A1D07E}\begin{tabular}[c]{@{}r@{}}3\\      (0.00555)\end{tabular}       & \cellcolor[HTML]{DFE282}\begin{tabular}[c]{@{}r@{}}5\\      (0.47747)\end{tabular}        & \cellcolor[HTML]{FDB87B}\begin{tabular}[c]{@{}r@{}}8\\      (0.00529)\end{tabular}                 & \cellcolor[HTML]{FFEB84}\begin{tabular}[c]{@{}r@{}}6\\      (0.06383)\end{tabular}                  & \cellcolor[HTML]{FFEB84}\begin{tabular}[c]{@{}r@{}}6\\      (0.00387)\end{tabular}             & \cellcolor[HTML]{FB9D75}\begin{tabular}[c]{@{}r@{}}9\\      (45.25532)\end{tabular}             & \cellcolor[HTML]{82C77C}\begin{tabular}[c]{@{}r@{}}2\\      (0.0114)\end{tabular}      & \cellcolor[HTML]{82C77C}\begin{tabular}[c]{@{}r@{}}2\\      (0.10076)\end{tabular}       & \cellcolor[HTML]{FB9975}9                                                  \\ \hline
\textbf{VAE\_JAMOTTON}         & \cellcolor[HTML]{FFEB84}\begin{tabular}[c]{@{}r@{}}6\\      (49.58059)\end{tabular}  & \cellcolor[HTML]{FED280}\begin{tabular}[c]{@{}r@{}}7\\      (55.09532)\end{tabular}   & \cellcolor[HTML]{F8696B}\begin{tabular}[c]{@{}r@{}}11\\      (7.8)\end{tabular}                        & \cellcolor[HTML]{82C77C}\begin{tabular}[c]{@{}r@{}}2\\      (0.4)\end{tabular}                           & \cellcolor[HTML]{FFEB84}\begin{tabular}[c]{@{}r@{}}6\\      (0.4)\end{tabular}                           & \cellcolor[HTML]{FDB87B}\begin{tabular}[c]{@{}r@{}}8\\      (0.35376)\end{tabular}               & \cellcolor[HTML]{FDB87B}\begin{tabular}[c]{@{}r@{}}8\\      (0.25501)\end{tabular}  & \cellcolor[HTML]{FED280}\begin{tabular}[c]{@{}r@{}}7\\      (0.00203)\end{tabular}     & \cellcolor[HTML]{A1D07E}\begin{tabular}[c]{@{}r@{}}3\\      (0.31078)\end{tabular}      & \cellcolor[HTML]{FB9D75}\begin{tabular}[c]{@{}r@{}}9\\      (0.02237)\end{tabular}       & \cellcolor[HTML]{FED280}\begin{tabular}[c]{@{}r@{}}7\\      (0.95891)\end{tabular}        & \cellcolor[HTML]{FED280}\begin{tabular}[c]{@{}r@{}}7\\      (0.00523)\end{tabular}                 & \cellcolor[HTML]{FDB87B}\begin{tabular}[c]{@{}r@{}}8\\      (0.07658)\end{tabular}                  & \cellcolor[HTML]{FDB87B}\begin{tabular}[c]{@{}r@{}}8\\      (0.00784)\end{tabular}             & \cellcolor[HTML]{A1D07E}\begin{tabular}[c]{@{}r@{}}3\\      (0.21804)\end{tabular}              & \cellcolor[HTML]{FB9D75}\begin{tabular}[c]{@{}r@{}}9\\      (0.02938)\end{tabular}     & \cellcolor[HTML]{FB9D75}\begin{tabular}[c]{@{}r@{}}9\\      (0.2745)\end{tabular}        & \cellcolor[HTML]{FA8170}10                                                 \\ \hline
\textbf{COTE\_MC\_WGAN\_GP}    & \cellcolor[HTML]{FA8370}\begin{tabular}[c]{@{}r@{}}10\\      (95.74713)\end{tabular} & \cellcolor[HTML]{F8696B}\begin{tabular}[c]{@{}r@{}}11\\      (101.62427)\end{tabular} & \cellcolor[HTML]{DFE282}\begin{tabular}[c]{@{}r@{}}5\\      (11.8)\end{tabular}                        & \cellcolor[HTML]{FDB87B}\begin{tabular}[c]{@{}r@{}}8\\      (2)\end{tabular}                             & \cellcolor[HTML]{A1D07E}\begin{tabular}[c]{@{}r@{}}3\\      (0.2)\end{tabular}                           & \cellcolor[HTML]{FA8370}\begin{tabular}[c]{@{}r@{}}10\\      (0.3585)\end{tabular}               & \cellcolor[HTML]{F8696B}\begin{tabular}[c]{@{}r@{}}11\\      (0.25585)\end{tabular} & \cellcolor[HTML]{F8696B}\begin{tabular}[c]{@{}r@{}}11\\      (0.00447)\end{tabular}    & \cellcolor[HTML]{F8696B}\begin{tabular}[c]{@{}r@{}}11\\      (49.09032)\end{tabular}    & \cellcolor[HTML]{FFEB84}\begin{tabular}[c]{@{}r@{}}6\\      (0.0105)\end{tabular}        & \cellcolor[HTML]{FB9D75}\begin{tabular}[c]{@{}r@{}}9\\      (1.30688)\end{tabular}        & \cellcolor[HTML]{FFEB84}\begin{tabular}[c]{@{}r@{}}6\\      (0.00448)\end{tabular}                 & \cellcolor[HTML]{FB9D75}\begin{tabular}[c]{@{}r@{}}9\\      (0.08906)\end{tabular}                  & \cellcolor[HTML]{F8696B}\begin{tabular}[c]{@{}r@{}}11\\      (0.02414)\end{tabular}            & \cellcolor[HTML]{F8696B}\begin{tabular}[c]{@{}r@{}}11\\      (312.41321)\end{tabular}           & \cellcolor[HTML]{FFEB84}\begin{tabular}[c]{@{}r@{}}6\\      (0.01557)\end{tabular}     & \cellcolor[HTML]{C0D980}\begin{tabular}[c]{@{}r@{}}4\\      (0.1074)\end{tabular}        & \cellcolor[HTML]{F8696B}11                                                 \\ \hline
\end{tabular}%
}
\caption{Summary of evaluation results across tested algorithms when the variable dependence is based on interaction formula \eqref{eval:interaction_formula}. \blueparagraph{Meaning of those metrics is summarized in Appendix \ref{app:summary_metrics}.}}
\label{results:interaction_table}
\end{table}

Table \ref{results:interaction_table} shows very similar results to what we saw from in Table \ref{results:linear_table}. No method is able to beat the results we observe for training data. However, here we can see that the model accuracy metrics, even for the best methods, are further away than in the linear case. In other words, this indicates that the generator fidelity depends on the underlying data dependence structure.

From the model-specific results, we see that the coefficient distance metrics $M_1$ and $M_2$ have much smaller values than in the linear case. The best methods based on coefficient difference metrics are again \texttt{MICE\_PART\_SYN} and \texttt{MICE\_FULL\_SYN}, and the worst method seems to be \texttt{COTE\_MC\_WGAN\_GP} and \texttt{CTGAN}. Notably, \texttt{CTGAN\_WITH\_AE} and \texttt{CTGAN\_WITH\_AE\_MICE} show a big uplift in rankings while additional MICE-based methods lose ranks compared to the linear case. {This suggests that introducing the interaction terms specified in formula \eqref{eval:interaction_formula} reduces the relative advantage of the MICE-based methods, although they remain among the strongest overall performers across the considered metrics.}

Model goodness-of-fit metrics yield slightly different results than model coefficient metrics. MICE-based methods lead in terms of these metrics, while GAN and VAE-based methods show worse results. Notably, \texttt{CTGAN\_WITH\_AE} shows much better results when compared to other GAN-based generation methods.

The variable selection, however, shows much more varied results. Again, \texttt{MICE\_PART\_SYN} and \texttt{MICE\_FULL\_SYN} methods show the best results, where variable selection, on average, selected 15.4 and 15.6 correct terms. Beyond these two methods, the next-best method selected, on average, 12 correct terms. Variable selection for \texttt{VAE\_JAMOTTON} got the worst result, on average, selecting only 7.8 correct terms for the model. We can also see that for almost all methods, variable selection added additional incorrect terms, where only \texttt{VAE\_JAMOTTON} picked on average 0.4 additional incorrect terms. These combined facts indicate that using \texttt{VAE\_JAMOTTON} results in a model, which is not desirable in this case, {since the model selected fails to include many of the interaction terms specified in the true claim generation formula (\ref{eval:interaction_formula}).}

From the data-specific results, we observe rankings that are almost identical to those for the linear case. This is expected, since we did not introduce any additional inter-variable relationships beyond the interaction between the response and the two covariates. Here again, the use of autoencoders improves MAE/MAPE for categorical columns; however, this improvement is not sufficient to beat the results of MICE-based methods. Again, \texttt{MICE\_TABULATOR} shows very strong performance on categorical columns {proportionally but fails to capture the dependence structure underneath}. Additionally, \texttt{MICE\_VV} and \texttt{MICE\_TABULATOR} indicate a loss of the original data {numeric variable} correlation structure. {Lastly, from Table \ref{results:interaction_table}, we can see that Cramér's V results for \texttt{CTGAN\_WITH\_AE} and \texttt{CTGAN\_WITH\_AE\_MICE} (as they share categorical variables) are above those for \texttt{MICE\_PART\_SYN} and \texttt{MICE\_FULL\_SYN}. The difference is minor, but this again justifies the use of autoencoders for model variable pre-processing.}

Upon analyzing synthetic data on a more granular level, we noticed that \texttt{COTE\_MC\_WGAN\_GP} method {can generate implausible numeric values, such as observations for which the driver's age is below $18$. By construction, the considered MICE-RF approaches do not generate values outside the observed range of a numerical variable, because the RF imputation algorithm samples each imputed value from observed donor values associated with the relevant terminal nodes, cf. Algorithm A.1 in \cite{doove2014recursive}. This property may also be viewed as a limitation when the objective is to generate previously unobserved extreme values. Nevertheless, it is important that a synthetic data generator produces plausible observations. In this respect, the considered MICE-based data generators preserve the observed ranges of numerical values and, in addition, appear sufficiently flexible to accommodate business constraints when necessary, as briefly explained in Section \ref{subsec:MICE_theory}.}

{Additionally, we conducted an empirical check to calculate the exact match rate between the synthetic and original training data. The exact match rate (the proportion of rows with exactly the same covariate values as the training data) for all methods is below $1\%$, and for MICE partially and fully synthetic at $0\%$. In general, data privacy is a very broad topic and there are many different privacy metrics, see, e.g., \cite{boudewijn2023privacy}, \cite{pilgram2025consensus}. A comprehensive analysis of how well each of the considered synthetic data generators preserves privacy is beyond the scope of our paper and is one of the natural questions for future research.}

Complementing the results in Table \ref{results:linear_table} and Table \ref{results:interaction_table}, we analyzed the averaged (across all experiments) covariance matrices for numeric variables in the generated data. The resulting covariance matrices support the findings above, identifying \texttt{MICE\_PART\_SYN} and \texttt{MICE\_FULL\_SYN} methods give, on average, the closest covariance matrices to the covariance matrix for the original numerical variables. However, it is important to note that GAN-based models performed better than MICE-based methods with respect to the variance of the \texttt{density} column. In particular, \texttt{CTGAN} maintained accurate variance, whereas the MICE-based methods underestimated this dispersion measure compared to the original data.

\begin{figure}
    \centering
    \includegraphics[width=0.95\linewidth]{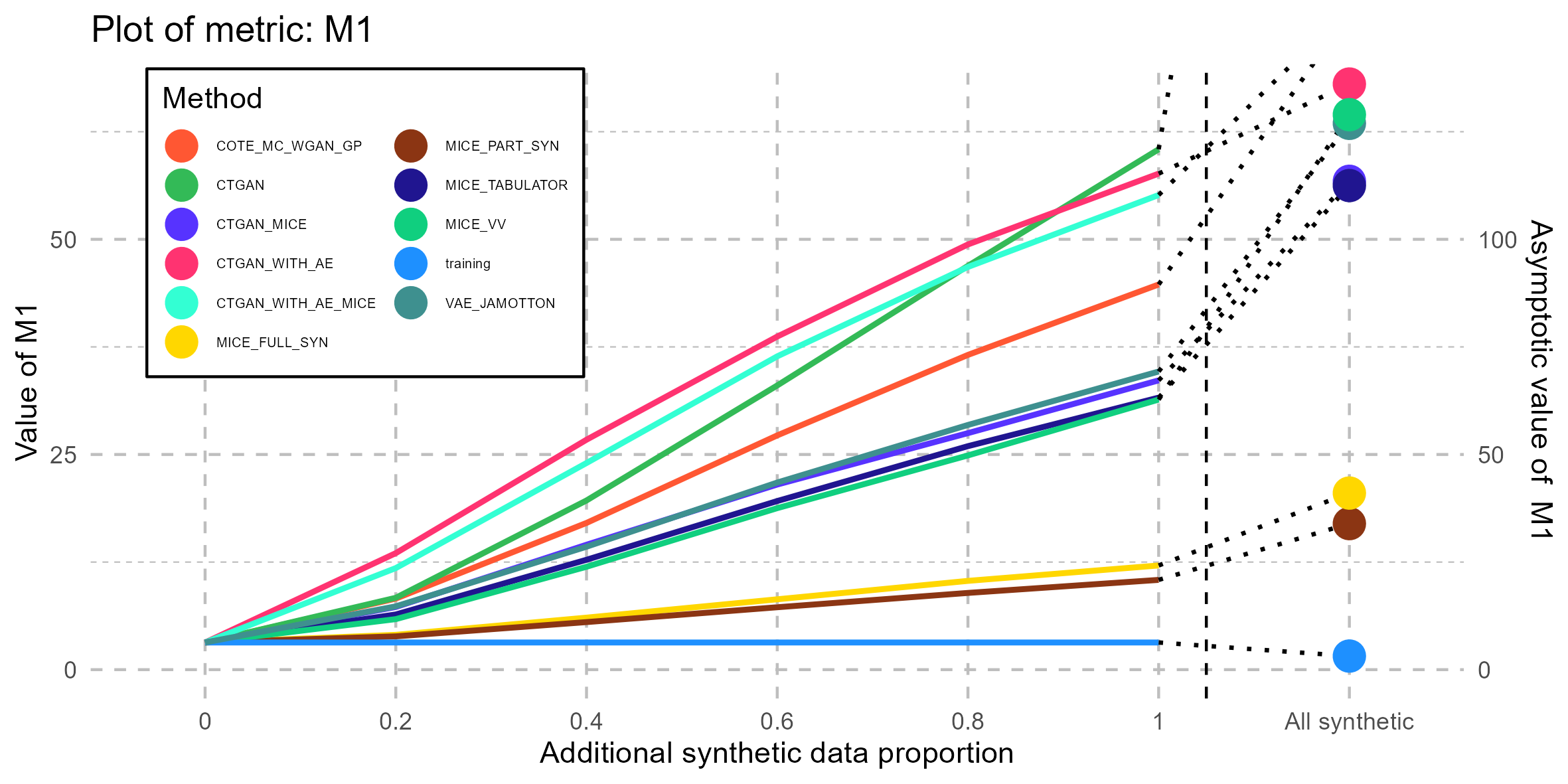}
    \caption{\small $M_1$ metric results as a function of synthetic data proportion. "All synthetic" represents the cases where only synthetic data is used and are treated as "asymptotic" estimation of the shown relationship. The "All synthetic" metric values are scaled 1:2 (right side axis) compared to the other presented values (left side axis). The dotted lines represent the corresponding asymptotic trends.}
    \label{fig:M1}
\end{figure}

Having discussed data generation fidelity, we now focus on data augmentation, {which is one possible application of synthetic data generation. The objective of this experiment is not to demonstrate improved GLM estimation. Since the synthetic observations are generated from the original training data, they do not provide additional independent information. Instead, we assess how much information is retained as the proportion of synthetic observations increases.} For this, we examine how the value of the $M_1$ metric depends on the proportion of synthetic data added. We observe that as we increase the proportion of additional synthetic data compared to the original training data, the value for the $M_1$ metric increases. 
The only case in which we observed a decrease in a metric is for the metric $M_1$ for the method \texttt{MICE\_PART\_SYN} and a synthetic data proportion of $0.2$ in the interaction case. In this case, the augmented data achieved the $M_1$ metric value of $0.78717$ while the corresponding training data metric was $0.86047$. This is the only result where this happened, it is not enough to say that augmentation of the original \textit{freMTPL2freq} data might help when it comes to better accuracy of coefficients. {Overall, the results are consistent with the fact that synthetic observations do not provide additional independent information. Therefore, systematic improvements in parameter estimation are not expected.}

The general relationship between the increase in the proportion of additional synthetic data and the metric $M_1$ value is shown in Figure \ref{fig:M1}.  As the proportion of synthetic data increases, we observe a clear upward trend in the $M_1$ metric value for the given method. Based on the way the $M_1$ metric is calculated and how GLM relies on the underlying data to compute the coefficient values, we estimate that if we keep on increasing the proportion of additional synthetic data, then we will asymptotically reach the metric values which we can compute for the case where we use only synthetic data. These pure synthetic data results are shown on the right side of the plot. {The corresponding results for $M_2$ are qualitatively similar and are therefore not shown.}

{The same case study setting was also applied to an additional Swedish motorcycle claim frequency dataset used in \cite{Jamotton2024}. Overview of this additional case study and its results can be found in Appendix \ref{app:case_study_2}.}

\section{Conclusion}\label{sec:conclusion}

Our paper demonstrates that the amputation-imputation framework based on MICE can be effectively used {for generating synthetic data in} an actuarial context. This framework is competitive with more sophisticated deep generative models, such as CTGANs or VAEs, for generating ratemaking data.

In our case studies based on the open-source \textit{freMTPL2freq} dataset {and the Swedish motorcycle dataset}, synthetic data generated using MICE with RFs as the imputation engine shows, on average, the best properties in terms of modeling efficacy and achieves high ranks in model accuracy and coefficient difference metrics. In general, several of the $10$ tested methods for synthetic data generation seem to have their niches for specific use cases.

Having subjectively evaluated the ease of use for each tested method, we identify MICE as the most user-friendly approach, since it requires the least amount of additional work to get started, which is an important practical aspect for actuaries. Custom deep generative models, such as WGAN by \cite{Cote2020} or VAE by \cite{Jamotton2024}, seem to require significantly more expertise and tuning to get good results.

{These results should be interpreted within the considered actuarial framework, namely Poisson claim frequency modeling with GLMs, rather than as a general ranking of synthetic data generation methods. For this setting, simple MICE-based approaches provide a competitive and practical alternative to more complex generative models. They require substantially less pre-processing, parameter tuning and implementation effort, while achieving comparable results across the considered evaluation criteria. The proposed evaluation framework is more generally applicable, although the quantitative results may differ for other response distributions, modeling approaches, or insurance applications.}

Our case studies also show that generic data augmentation does not generally improve the estimation of GLMs trained on the real data augmented with the synthetic one. We observe that increasing the proportion of synthetic data relative to the original training data results in approximately linear increases in metrics measuring the difference between the coefficients of two GLMs, e.g., the true GLM vs a GLM estimated on augmented data. We have observed only one case in which data augmentation improved the estimation of the true GLM coefficients.

We leave several aspects for future research. For example, it may be interesting to investigate how to integrate common business constraints into the amputation-imputation framework based on MICE {and to study the sensitivity of MICE-based synthesizers to implementation choices such as the initial imputation strategy, the number of iterations, etc.}. Another research direction is the exploration of targeted data augmentation. While ratemaking has historically relied on association measures like correlation, future research must rigorously evaluate the extent to which synthetic data generators preserve the underlying causal structure of insurance risk. Finally, future studies should systematically assess the disclosure risk inherent in each generator, specifically quantifying their vulnerability to membership inference and reconstruction attacks.

\section*{Acknowledgments}
{The authors thank two anonymous referees for their valuable comments that significantly improved the paper.}

Yevhen Havrylenko acknowledges the financial support by the PRIME program (\href{https://www.daad.de/en/studying-in-germany/scholarships/daad-funding-programmes/prime/prime-fellows-202324/}{link}) of the German Academic Exchange Service (DAAD, \href{https://ror.org/039djdh30}{https://ror.org/039djdh30}) with funds from the German Federal Ministry of Education and Research (BMBF). Yevhen Havrylenko acknowledges the support of Ulm University, the University of Copenhagen and Johannes Kepler University Linz, where he conducted parts of this research project. Meelis Käärik and Artur Tuttar acknowledge the support of the Estonian Research Council grants PRG1197 and PRG3105.

The authors of this publication express their gratitude to Raul Kangro and Märt Möls for their insights, ideas, and discussion related to the article. The authors are grateful to the participants of the 28th International Congress on Insurance: Mathematics and Economics, 1st ASTIN Bulletin conference, the LIDAM ISBA Workshop on Fairness and Machine Learning in Insurance, a research seminar at Korea University, and the 7th European Actuarial Journal conference. The authors are thankful to Magnus Käärik for assistance in refining the language and enhancing clarity.

\section*{Declaration of generative AI and AI-assisted technologies}
The authors used ChatGPT (free and EDU version) and Gemini (Flash and Pro) models to improve the language of the manuscript. After using those large language models, the authors reviewed and edited the content as needed and take full responsibility for the content of the article.

\section*{Declaration of competing interest}
The authors declare no conflict of interest.

\section*{Data availability statement}
The authors intend to upload their code and data to a publicly accessible GitHub repository upon acceptance of the article for publication.

\bibliographystyle{apalike}
\bibliography{Literature}

\newpage
\appendix
\section{Brief summary of autoencoders}\label{app:summary_AE}

In this appendix, we explain how deterministic AEs are combined with synthetic data generators (e.g., CTGAN) to potentially improve the quality of synthetic categorical variables. 

Before training a generation method $GM_A$, we perform the following steps for each categorical variable $x_j$:
\begin{enumerate}
    \item represent $x_j$ as $\xbold_j^{cat}$ via one-hot encoding;
    \item train a separate AE (see Section \ref{subsec:AE}), which finds a low-dimensional representation (encoding) $\varphi_j(\xbold_j^{cat})$ of $\xbold_j^{cat}$;
    \item in $\SSet_{train}$, replace $x_j$ by $\varphi_j(\xbold_j^{cat})$.
\end{enumerate}

Since the generation method $GM_{A}$ generates data in the same form as it is trained on, the synthetic data will have a low-dimensional numeric representation of the categorical variables. This is an issue since the meaning of the categorical variables is not apparent. To resolve this, we perform the following steps for each vector of features corresponding to a categorical variable (in its quasi-encoded representation) in the synthetic dataset:
\begin{enumerate}
    \item use Nearest Centroid Classification to identify for this vector its closest class; the centroid of each class is determined by the encoding (given by the respective AE fitted before training $GM_A$) of the respective category of the original categorical variable;
    \item replace this vector by the one-hot encoded representation of the category that corresponds to the centroid of the class to which the generated vector belongs; this one-hot encoded representation can be recovered from the centroid using the decoding part of AE.
\end{enumerate}

\newpage
\section{Summary of metrics used in the case studies}\label{app:summary_metrics}

{
In this appendix, a summary table describing each metric used in the case study is included.

\keepXColumns 

\captionof{table}{Metric Description Table}
\label{tab:metric_description}

\begin{tabularx}{\textwidth}{>{\bfseries}l X}
\toprule
\textbf{Metric} & \textbf{Description} \\
\midrule
\endhead 

\bottomrule
\endfoot 

\bottomrule
\endlastfoot 

$M_1$ & Measures the normalized distance between the true GLM coefficients and the GLM coefficients estimated using synthetic data generated by a specific approach and data structure. See equation \eqref{eq:coefficient_distance_metric}.\\
$M_2$ & Measures the relative increase in GLM coefficient estimation uncertainty when synthetic data is used instead of or in addition to real training data. See equation \eqref{eq:SD_increase}.\\
Correct variables selected & The average number of true explanatory variables (present in the underlying claim generation formula) that were correctly chosen by the automated variable selection method. \\
Incorrect variables selected & The average number of irrelevant variables (not present in the true claim generation formula) that were erroneously chosen by the automated variable selection method. \\
Missing main effects variable & Applicable in models with interaction terms; The average number of instances where an interaction term was included by variable selection without its corresponding main effect being selected. \\
Fit Poisson deviance & A model goodness-of-fit metric evaluating the accuracy of the Poisson GLM predictions on unseen test data. \\
Fit RMSE & Root Mean Squared Error, measuring the average magnitude of prediction error between observed and predicted claim counts on test data. \\
Pair-wise MAE & Mean Absolute Error measuring the difference between real and synthetic data in the observed proportions of joint categories/binned value combinations across variable pairs. \\
Pair-wise MAPE & Mean Absolute Percentage Error assessing the proportion accuracy for pairwise variable combinations between real and synthetic data. \\
\pagebreak 
Correlation MAE & Mean Absolute Error assessing how accurately synthetic data preserves correlation across all pairs of numeric variables compared to the original data. \\
Correlation MAPE & Mean Absolute Percentage Error evaluating relative errors in numeric pairwise correlations between real and synthetic datasets. \\
Categorical column MAE & Mean Absolute Error measuring the average discrepancy between real and synthetic data in the proportion/frequency of occurrence across individual categorical feature levels. \\
Categorical column MAPE & Mean Absolute Percentage Error measuring relative proportion errors for individual categorical features. \\
Numeric column MAE & Mean Absolute Error evaluating how closely synthetic binned (by quantiles) numeric features reproduce the proportion distribution of binned quantiles of original numeric columns. \\
Numeric column MAPE & Mean Absolute Percentage Error evaluating relative distributional deviations in binned quantiles of numeric features. \\
Cramér's V MAE & Mean Absolute Error measuring how accurately the pairwise categorical dependence structure (measured via Cramér's V) in real data is preserved in synthetic data. \\
Cramér's V MAPE & Mean Absolute Percentage Error assessing relative accuracy in preserving categorical association metrics (Cramér's V). \\
Ease of use ranking & A subjective assessment ranking methods based on their intuitiveness, operational complexity (pipeline setup), and deployment overhead. \\
\end{tabularx}

}

\newpage
\section{Comparison of real and synthetic marginal frequencies }\label{app:comparison_of_marginals}
{
In this appendix, we compare the marginal distributions of variables in the real dataset and in synthetic datasets.

Two plots are created for all categorical variables and all numeric variables. The plot shows the frequency for a given category of the variable (shown in symbol and color) in the training data (x-axis) versus the corresponding frequency of the category in synthetic data (y-axis). This plot is added for every synthetic data generation method. In an ideal situation, all points should lie on or close to the bisector line shown on the plots. If points deviate from the line, then the corresponding category is underrepresented (if the point is below the line) or overrepresented (if the point is above the line) in the synthetic data for the given method.

For numeric variables, all values have been binned into 8 bins using the quantile values in the same manner as described in the main text of the paper on page 20. The resulting plots show the frequency of the numeric variable bins in training and synthetic data. This was done because most numeric variables considered in the study were not strictly continuous, and some variables had point masses at specific values (e.g., the bonus-malus variable with a point mass at 50) in the training data.

\begin{figure}[h]
    \centering
    \includegraphics[width=0.95\linewidth]{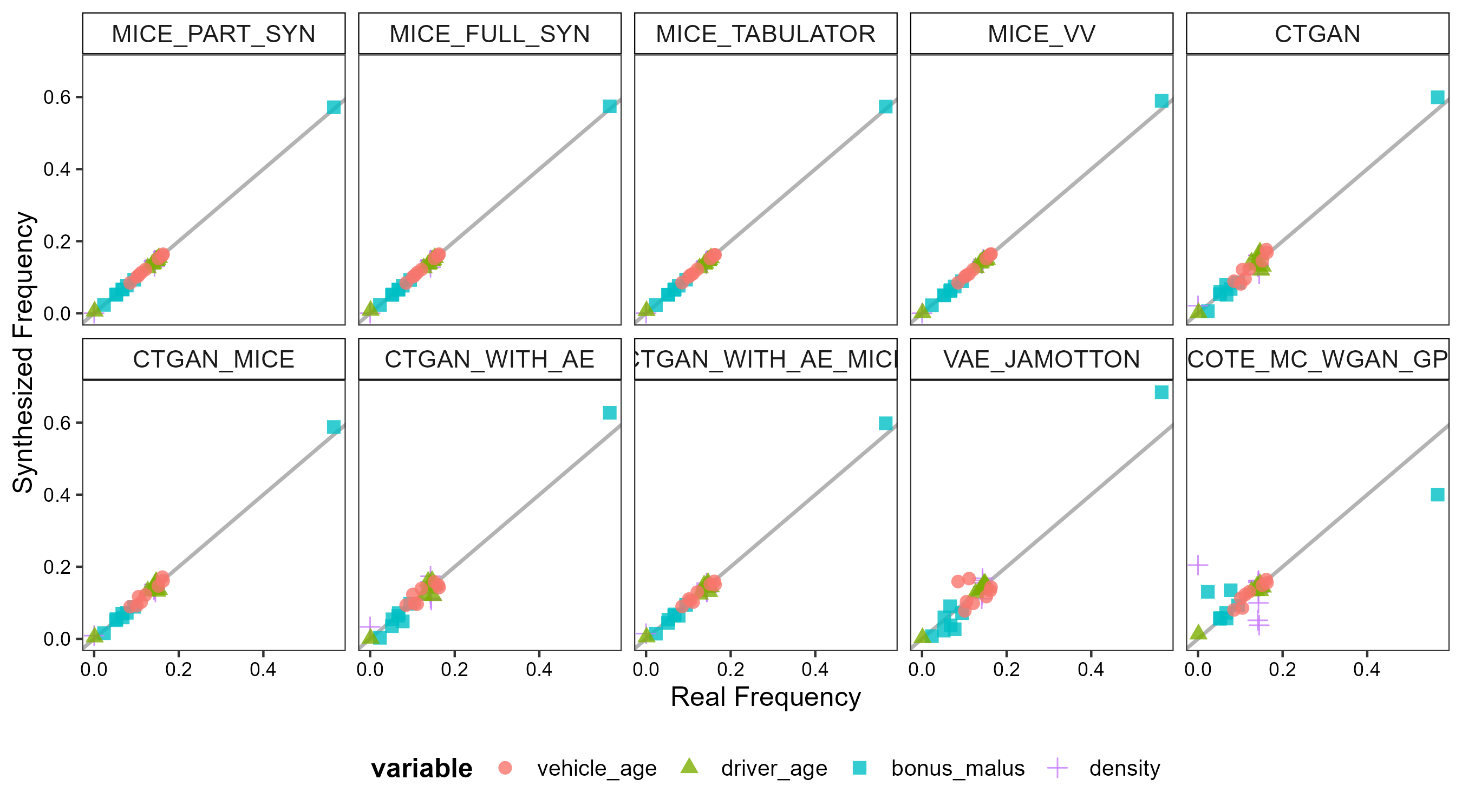}
    \caption{\small Frequencies of real versus synthetic binned (8 bins) numerical variables for all synthetic data generation methods trained using the linear formula \eqref{eval:linear_formula} training data}
    \label{fig:lin_cat_dist}
\end{figure}
\begin{figure}
    \centering
    \includegraphics[width=0.95\linewidth]{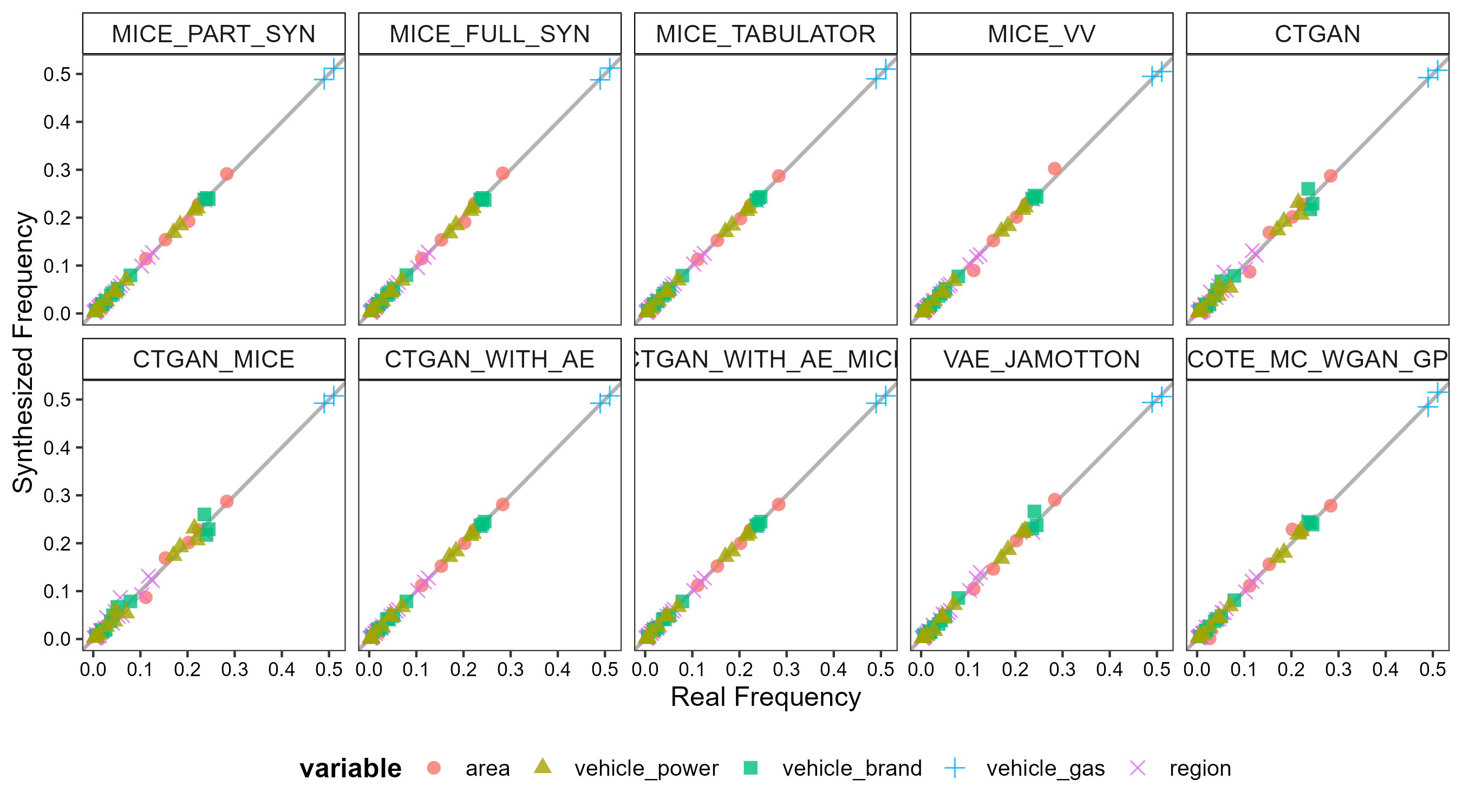}
    \caption{\small Frequencies of real versus synthetic categorical variables for all synthetic data generation methods trained using the linear formula \eqref{eval:linear_formula} training data}
    \label{fig:lin_num_dist}
\end{figure}

\begin{figure}
    \centering
    \includegraphics[width=0.95\linewidth]{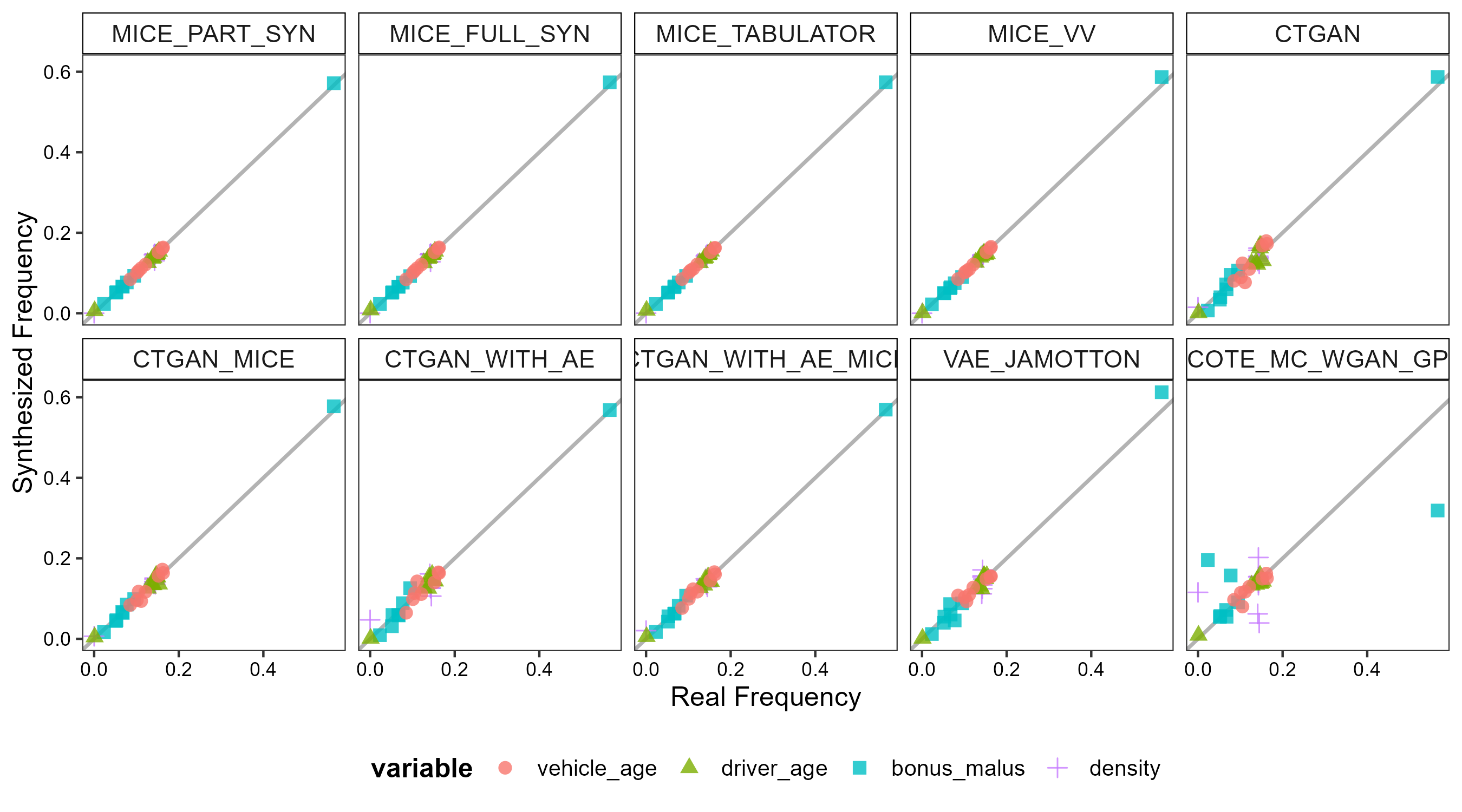}
    \caption{\small Frequencies of real versus synthetic binned (8 bins) numerical variables for all synthetic data generation methods trained using the interaction formula \eqref{eval:interaction_formula} training data}
    \label{fig:int_cat_dist}
\end{figure}
\begin{figure}
    \centering
    \includegraphics[width=0.95\linewidth]{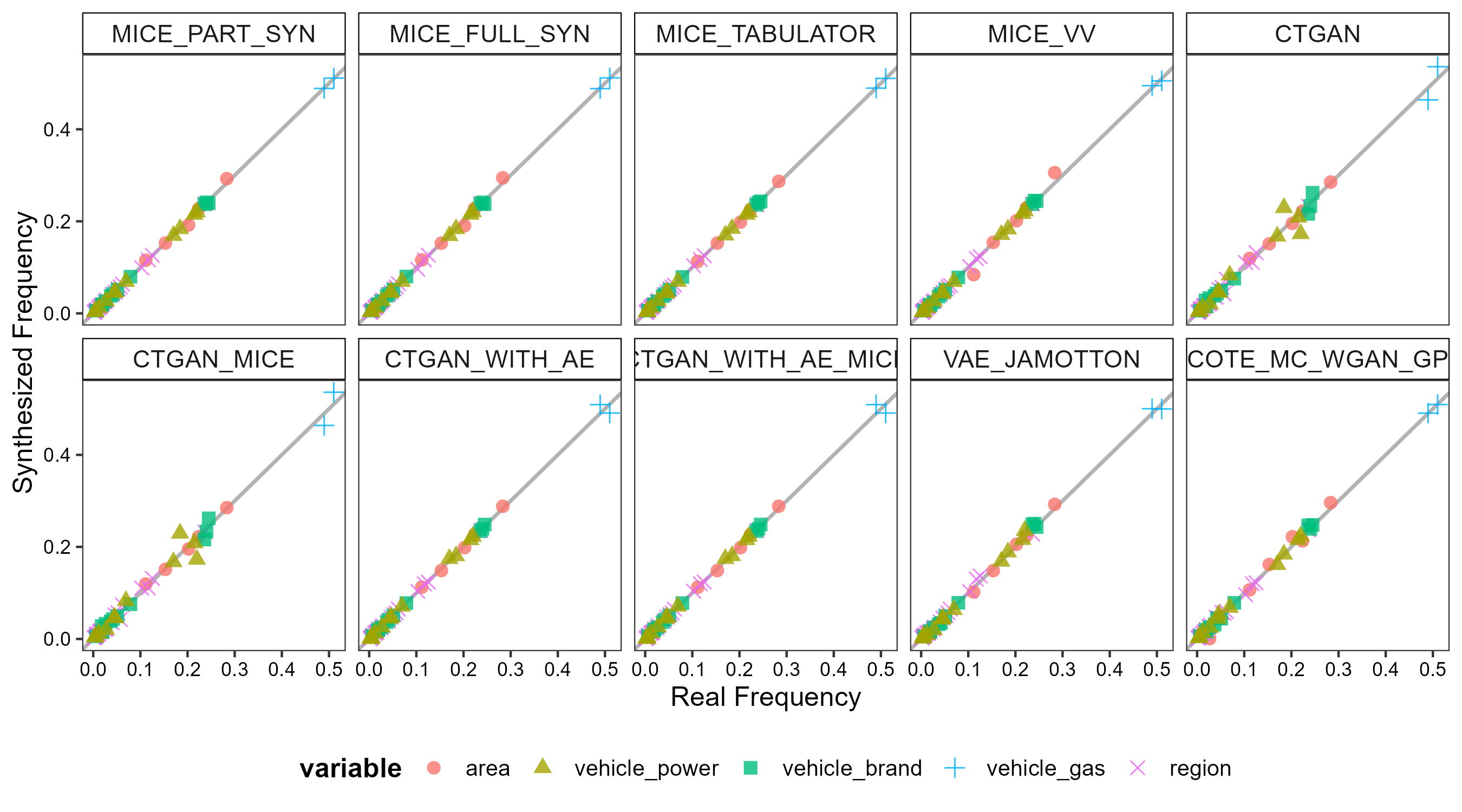}
    \caption{\small Frequencies of real versus synthetic categorical variables for all synthetic data generation methods trained using the interaction formula \eqref{eval:interaction_formula} training data}
    \label{fig:int_num_dist}
\end{figure}

From the plots, similar results and insights are evident, as in the case study results tables (Table \ref{results:linear_table} and Table \ref{results:interaction_table}). MICE-based methods perform well in both settings and both variable types. Methods that gave worse results with respect to numeric variable accuracy can clearly be seen diverging from the bisector line. For both the linear and interaction cases, the effect of using autoencoders can be seen in the marginal distributions of categorical variables. Similarly, the effect of using CTGAN with MICE is evident here, as the frequencies of binned numeric variables are closer to the bisector line after combining the methods.
}

\section{Summary of the second case study}\label{app:case_study_2}
{
In this appendix, the results of the same case study evaluated on a Swedish motorcycle dataset used in \cite{Jamotton2024} are presented. The setup of the case study, training of the generator models, and their evaluation using model and data metrics are handled the same way as in the main case study.

The Swedish motorcycle dataset is about $10$ times smaller than the \textit{freMTPL2freq} dataset. Readers interested in the details of this smaller dataset can find the results of the respective exploratory analysis in Section 13.2 in \cite{Wuethrich2023}.

As described in Subsection \ref{subsec:experiment_evaluation}, our case study uses two relationships to calculate the mean of Poisson random variables used to generate the claim counts. The linear relationship used is
\begin{align}\label{eval:mccase_linear_formula}
\begin{split}
     f_{lin\_alt}(\xbold) &= -3 + 0.0075\cdot (\text{vehicleAge}-5) - 0.0075\cdot (\text{ownerAge}-35) \\
     &+ 0.15\cdot \mathds{1}_{\fixparen{\text{geoZone} \in \text{\{1,2\}}}} - 0.5\cdot \mathds{1}_{\fixparen{\text{geoZone} \in \text{\{5,6,7\}}}}\\ 
     &- 0.3\cdot \mathds{1}_{\fixparen{\text{vehicleClass} \in \text{\{1,2\}}}} + 0.15\cdot \mathds{1}_{\fixparen{\text{vehicleClass} \in \text{\{4\}}}} \\
     &+ 0.35\cdot \mathds{1}_{\fixparen{\text{vehicleClass} \in \text{\{5\}}}} + 0.55\cdot \mathds{1}_{\fixparen{\text{vehicleClass} \in \text{\{6,7\}}}}\\
     &+ 0.25\cdot \mathds{1}_{\fixparen{\text{bonusMalus} \in \text{\{1,2\}}}} + 0.125\cdot \mathds{1}_{\fixparen{\text{bonusMalus} \in \text{\{3,4\}}}},
\end{split}
\end{align}
where the variable \textit{gender} was not used in the model at all.

The interaction relationship used is
\begin{align}\label{eval:mccase_int_formula}
\begin{split}
     f_{int\_alt}(\xbold) &= f_{lin\_alt}(\xbold) - 0.25\cdot \mathds{1}_{\fixparen{\text{geoZone} \in \text{\{1,2\}}}}\cdot\mathds{1}_{\fixparen{\text{bonusMalus} \in \text{\{5,6,7\}}}}\\
     &+ 0.25\cdot \mathds{1}_{\fixparen{\text{geoZone} \in \text{\{1,2\}}}}\cdot\mathds{1}_{\fixparen{\text{bonusMalus} \in \text{\{1,2\}}}}\\
     &+ 0.25\cdot \mathds{1}_{\fixparen{\text{geoZone} \in \text{\{5,6,7\}}}}\cdot\mathds{1}_{\fixparen{\text{bonusMalus} \in \text{\{3,4\}}}}\\
     &+ 0.015\cdot \text{vehicleAge}\cdot \mathds{1}_{\fixparen{\text{vehicleClass} \in \text{\{5,6,7\}}}}\\
     &-0.01\cdot \text{vehicleAge}\cdot \mathds{1}_{\fixparen{\text{vehicleClass} \in \text{\{1,2,3\}}}}\\
\end{split}
\end{align}

The same generation methods as in the main study are evaluated and $n_E = 5$ datasets are generated for each of the methods.

\begin{table}[!ht]
\centering
\resizebox{\textwidth}{!}{%
\begin{tabular}{|l|r|r|r|r|r|r|r|r|r|r|r|r|r|r|r|r|l|}
\hline
\textbf{Generation   method}   & \multicolumn{1}{l|}{\textbf{M1}}                                                      & \multicolumn{1}{l|}{\textbf{M2}}                                                      & \multicolumn{1}{l|}{\textbf{\begin{tabular}[c]{@{}l@{}}Correct \\ variables \\ selected\end{tabular}}} & \multicolumn{1}{l|}{\textbf{\begin{tabular}[c]{@{}l@{}}Incorrect \\ variables \\ selected\end{tabular}}} & \multicolumn{1}{l|}{\textbf{\begin{tabular}[c]{@{}l@{}}Fit \\ Poisson \\ deviance\end{tabular}}} & \multicolumn{1}{l|}{\textbf{\begin{tabular}[c]{@{}l@{}}Fit\\ RMSE\end{tabular}}}    & \multicolumn{1}{l|}{\textbf{\begin{tabular}[c]{@{}l@{}}Pair-wise \\ MAE\end{tabular}}} & \multicolumn{1}{l|}{\textbf{\begin{tabular}[c]{@{}l@{}}Pair-Wise \\ MAPE\end{tabular}}} & \multicolumn{1}{l|}{\textbf{\begin{tabular}[c]{@{}l@{}}Correlation \\ MAE\end{tabular}}} & \multicolumn{1}{l|}{\textbf{\begin{tabular}[c]{@{}l@{}}Correlation \\ MAPE\end{tabular}}} & \multicolumn{1}{l|}{\textbf{\begin{tabular}[c]{@{}l@{}}Categorical \\ column \\ MAE\end{tabular}}} & \multicolumn{1}{l|}{\textbf{\begin{tabular}[c]{@{}l@{}}Categorical \\ column \\ MAPE\end{tabular}}} & \multicolumn{1}{l|}{\textbf{\begin{tabular}[c]{@{}l@{}}Numeric \\ column \\ MAE\end{tabular}}} & \multicolumn{1}{l|}{\textbf{\begin{tabular}[c]{@{}l@{}}Numeric \\ column \\ MAPE\end{tabular}}} & \multicolumn{1}{l|}{\textbf{\begin{tabular}[c]{@{}l@{}}Cramér's V\\ MAE\end{tabular}}} & \multicolumn{1}{l|}{\textbf{\begin{tabular}[c]{@{}l@{}}Cramér's V \\ MAPE\end{tabular}}} & \textbf{\begin{tabular}[c]{@{}l@{}}Ease \\ of use \\ ranking\end{tabular}} \\ \hline
\textbf{training}              & \cellcolor[HTML]{63BE7B}\begin{tabular}[c]{@{}r@{}}1\\      (1.12949)\end{tabular}    & \cellcolor[HTML]{63BE7B}\begin{tabular}[c]{@{}r@{}}1\\      (1)\end{tabular}          & \cellcolor[HTML]{63BE7B}\begin{tabular}[c]{@{}r@{}}1\\      (5)\end{tabular}                           & \cellcolor[HTML]{63BE7B}\begin{tabular}[c]{@{}r@{}}1\\      (0)\end{tabular}                             & \cellcolor[HTML]{63BE7B}\begin{tabular}[c]{@{}r@{}}1\\      (0.34792)\end{tabular}               & \cellcolor[HTML]{A1D07E}\begin{tabular}[c]{@{}r@{}}3\\      (0.24992)\end{tabular}  & \cellcolor[HTML]{63BE7B}\begin{tabular}[c]{@{}r@{}}1\\      (0.03307)\end{tabular}     & \cellcolor[HTML]{63BE7B}\begin{tabular}[c]{@{}r@{}}1\\      (1.07851)\end{tabular}      & \cellcolor[HTML]{63BE7B}\begin{tabular}[c]{@{}r@{}}1\\      (0)\end{tabular}             & \cellcolor[HTML]{63BE7B}\begin{tabular}[c]{@{}r@{}}1\\      (0)\end{tabular}              & \cellcolor[HTML]{63BE7B}\begin{tabular}[c]{@{}r@{}}1\\      (0)\end{tabular}                       & \cellcolor[HTML]{63BE7B}\begin{tabular}[c]{@{}r@{}}1\\      (0)\end{tabular}                        & \cellcolor[HTML]{63BE7B}\begin{tabular}[c]{@{}r@{}}1\\      (0)\end{tabular}                   & \cellcolor[HTML]{63BE7B}\begin{tabular}[c]{@{}r@{}}1\\      (0)\end{tabular}                    & \cellcolor[HTML]{63BE7B}\begin{tabular}[c]{@{}r@{}}1\\      (0)\end{tabular}           & \cellcolor[HTML]{63BE7B}\begin{tabular}[c]{@{}r@{}}1\\      (0)\end{tabular}             & \cellcolor[HTML]{63BE7B}1                                                  \\ \hline
\textbf{MICE\_PART\_SYN}       & \cellcolor[HTML]{82C77C}\begin{tabular}[c]{@{}r@{}}2\\      (5.30839)\end{tabular}    & \cellcolor[HTML]{82C77C}\begin{tabular}[c]{@{}r@{}}2\\      (6.71966)\end{tabular}    & \cellcolor[HTML]{DFE282}\begin{tabular}[c]{@{}r@{}}5\\      (3.8)\end{tabular}                         & \cellcolor[HTML]{63BE7B}\begin{tabular}[c]{@{}r@{}}1\\      (0)\end{tabular}                             & \cellcolor[HTML]{A1D07E}\begin{tabular}[c]{@{}r@{}}3\\      (0.3483)\end{tabular}                & \cellcolor[HTML]{82C77C}\begin{tabular}[c]{@{}r@{}}2\\      (0.2499)\end{tabular}   & \cellcolor[HTML]{A1D07E}\begin{tabular}[c]{@{}r@{}}3\\      (0.03325)\end{tabular}     & \cellcolor[HTML]{A1D07E}\begin{tabular}[c]{@{}r@{}}3\\      (1.1367)\end{tabular}       & \cellcolor[HTML]{A1D07E}\begin{tabular}[c]{@{}r@{}}3\\      (0.00532)\end{tabular}       & \cellcolor[HTML]{A1D07E}\begin{tabular}[c]{@{}r@{}}3\\      (0.29012)\end{tabular}        & \cellcolor[HTML]{A1D07E}\begin{tabular}[c]{@{}r@{}}3\\      (0.00133)\end{tabular}                 & \cellcolor[HTML]{A1D07E}\begin{tabular}[c]{@{}r@{}}3\\      (0.01161)\end{tabular}                  & \cellcolor[HTML]{A1D07E}\begin{tabular}[c]{@{}r@{}}3\\      (0.00083)\end{tabular}             & \cellcolor[HTML]{DFE282}\begin{tabular}[c]{@{}r@{}}5\\      (0.11737)\end{tabular}              & \cellcolor[HTML]{82C77C}\begin{tabular}[c]{@{}r@{}}2\\      (0.00547)\end{tabular}     & \cellcolor[HTML]{82C77C}\begin{tabular}[c]{@{}r@{}}2\\      (0.10007)\end{tabular}       & \cellcolor[HTML]{85C87D}2                                                  \\ \hline
\textbf{MICE\_FULL\_SYN}       & \cellcolor[HTML]{A1D07E}\begin{tabular}[c]{@{}r@{}}3\\      (5.91315)\end{tabular}    & \cellcolor[HTML]{A1D07E}\begin{tabular}[c]{@{}r@{}}3\\      (7.60619)\end{tabular}    & \cellcolor[HTML]{DFE282}\begin{tabular}[c]{@{}r@{}}5\\      (3.8)\end{tabular}                         & \cellcolor[HTML]{C0D980}\begin{tabular}[c]{@{}r@{}}4\\      (0.2)\end{tabular}                           & \cellcolor[HTML]{82C77C}\begin{tabular}[c]{@{}r@{}}2\\      (0.34827)\end{tabular}               & \cellcolor[HTML]{63BE7B}\begin{tabular}[c]{@{}r@{}}1\\      (0.24989)\end{tabular}  & \cellcolor[HTML]{C0D980}\begin{tabular}[c]{@{}r@{}}4\\      (0.03325)\end{tabular}     & \cellcolor[HTML]{82C77C}\begin{tabular}[c]{@{}r@{}}2\\      (1.13308)\end{tabular}      & \cellcolor[HTML]{82C77C}\begin{tabular}[c]{@{}r@{}}2\\      (0.0046)\end{tabular}        & \cellcolor[HTML]{82C77C}\begin{tabular}[c]{@{}r@{}}2\\      (0.17619)\end{tabular}        & \cellcolor[HTML]{C0D980}\begin{tabular}[c]{@{}r@{}}4\\      (0.00155)\end{tabular}                 & \cellcolor[HTML]{C0D980}\begin{tabular}[c]{@{}r@{}}4\\      (0.01251)\end{tabular}                  & \cellcolor[HTML]{C0D980}\begin{tabular}[c]{@{}r@{}}4\\      (0.00091)\end{tabular}             & \cellcolor[HTML]{C0D980}\begin{tabular}[c]{@{}r@{}}4\\      (0.1152)\end{tabular}               & \cellcolor[HTML]{A1D07E}\begin{tabular}[c]{@{}r@{}}3\\      (0.00676)\end{tabular}     & \cellcolor[HTML]{A1D07E}\begin{tabular}[c]{@{}r@{}}3\\      (0.12505)\end{tabular}       & \multicolumn{1}{l|}{\cellcolor[HTML]{C0D980}4}                             \\ \hline
\textbf{MICE\_VV}              & \cellcolor[HTML]{FB9D75}\begin{tabular}[c]{@{}r@{}}9\\      (28.82229)\end{tabular}   & \cellcolor[HTML]{FB9D75}\begin{tabular}[c]{@{}r@{}}9\\      (34.79883)\end{tabular}   & \cellcolor[HTML]{F8696B}\begin{tabular}[c]{@{}r@{}}11\\      (1.2)\end{tabular}                        & \cellcolor[HTML]{FDB87B}\begin{tabular}[c]{@{}r@{}}8\\      (0.6)\end{tabular}                           & \cellcolor[HTML]{FDB87B}\begin{tabular}[c]{@{}r@{}}8\\      (0.35388)\end{tabular}               & \cellcolor[HTML]{FDB87B}\begin{tabular}[c]{@{}r@{}}8\\      (0.25062)\end{tabular}  & \cellcolor[HTML]{82C77C}\begin{tabular}[c]{@{}r@{}}2\\      (0.03309)\end{tabular}     & \cellcolor[HTML]{FDB87B}\begin{tabular}[c]{@{}r@{}}8\\      (1.19556)\end{tabular}      & \cellcolor[HTML]{FB9D75}\begin{tabular}[c]{@{}r@{}}9\\      (0.01708)\end{tabular}       & \cellcolor[HTML]{FED280}\begin{tabular}[c]{@{}r@{}}7\\      (0.57358)\end{tabular}        & \cellcolor[HTML]{DFE282}\begin{tabular}[c]{@{}r@{}}5\\      (0.0051)\end{tabular}                  & \cellcolor[HTML]{DFE282}\begin{tabular}[c]{@{}r@{}}5\\      (0.03325)\end{tabular}                  & \cellcolor[HTML]{DFE282}\begin{tabular}[c]{@{}r@{}}5\\      (0.00159)\end{tabular}             & \cellcolor[HTML]{A1D07E}\begin{tabular}[c]{@{}r@{}}3\\      (0.1068)\end{tabular}               & \cellcolor[HTML]{FECB7E}\begin{tabular}[c]{@{}r@{}}7\\      (0.02563)\end{tabular}     & \cellcolor[HTML]{FECB7E}\begin{tabular}[c]{@{}r@{}}7\\      (0.44568)\end{tabular}       & \cellcolor[HTML]{85C87D}2                                                  \\ \hline
\textbf{MICE\_TABULATOR}       & \cellcolor[HTML]{FED280}\begin{tabular}[c]{@{}r@{}}7\\      (21.97599)\end{tabular}   & \cellcolor[HTML]{FDB87B}\begin{tabular}[c]{@{}r@{}}8\\      (27.19137)\end{tabular}   & \cellcolor[HTML]{FA8370}\begin{tabular}[c]{@{}r@{}}10\\      (1.6)\end{tabular}                        & \cellcolor[HTML]{C0D980}\begin{tabular}[c]{@{}r@{}}4\\      (0.2)\end{tabular}                           & \cellcolor[HTML]{FED280}\begin{tabular}[c]{@{}r@{}}7\\      (0.35223)\end{tabular}               & \cellcolor[HTML]{FFEB84}\begin{tabular}[c]{@{}r@{}}6\\      (0.2504)\end{tabular}   & \cellcolor[HTML]{DFE282}\begin{tabular}[c]{@{}r@{}}5\\      (0.0334)\end{tabular}      & \cellcolor[HTML]{FED280}\begin{tabular}[c]{@{}r@{}}7\\      (1.19525)\end{tabular}      & \cellcolor[HTML]{FDB87B}\begin{tabular}[c]{@{}r@{}}8\\      (0.01403)\end{tabular}       & \cellcolor[HTML]{DFE282}\begin{tabular}[c]{@{}r@{}}5\\      (0.37046)\end{tabular}        & \cellcolor[HTML]{82C77C}\begin{tabular}[c]{@{}r@{}}2\\      (0.00103)\end{tabular}                 & \cellcolor[HTML]{82C77C}\begin{tabular}[c]{@{}r@{}}2\\      (0.00977)\end{tabular}                  & \cellcolor[HTML]{82C77C}\begin{tabular}[c]{@{}r@{}}2\\      (0.0007)\end{tabular}              & \cellcolor[HTML]{82C77C}\begin{tabular}[c]{@{}r@{}}2\\      (0.07307)\end{tabular}              & \cellcolor[HTML]{FFEB84}\begin{tabular}[c]{@{}r@{}}6\\      (0.02374)\end{tabular}     & \cellcolor[HTML]{FFEB84}\begin{tabular}[c]{@{}r@{}}6\\      (0.42172)\end{tabular}       & \cellcolor[HTML]{CBDC81}4                                                  \\ \hline
\textbf{CTGAN}                 & \cellcolor[HTML]{FA8370}\begin{tabular}[c]{@{}r@{}}10\\      (58.59928)\end{tabular}  & \cellcolor[HTML]{FA8370}\begin{tabular}[c]{@{}r@{}}10\\      (52.52411)\end{tabular}  & \cellcolor[HTML]{63BE7B}\begin{tabular}[c]{@{}r@{}}1\\      (5)\end{tabular}                           & \cellcolor[HTML]{63BE7B}\begin{tabular}[c]{@{}r@{}}1\\      (0)\end{tabular}                             & \cellcolor[HTML]{FA8370}\begin{tabular}[c]{@{}r@{}}10\\      (0.36651)\end{tabular}              & \cellcolor[HTML]{FA8370}\begin{tabular}[c]{@{}r@{}}10\\      (0.25406)\end{tabular} & \cellcolor[HTML]{FDB87B}\begin{tabular}[c]{@{}r@{}}8\\      (0.03433)\end{tabular}     & \cellcolor[HTML]{FFEB84}\begin{tabular}[c]{@{}r@{}}6\\      (1.18245)\end{tabular}      & \cellcolor[HTML]{FA8370}\begin{tabular}[c]{@{}r@{}}10\\      (0.01831)\end{tabular}      & \cellcolor[HTML]{FA8370}\begin{tabular}[c]{@{}r@{}}10\\      (3.84642)\end{tabular}       & \cellcolor[HTML]{FA8370}\begin{tabular}[c]{@{}r@{}}10\\      (0.02037)\end{tabular}                & \cellcolor[HTML]{FB9D75}\begin{tabular}[c]{@{}r@{}}9\\      (0.14895)\end{tabular}                  & \cellcolor[HTML]{FB9D75}\begin{tabular}[c]{@{}r@{}}9\\      (0.01123)\end{tabular}             & \cellcolor[HTML]{FB9D75}\begin{tabular}[c]{@{}r@{}}9\\      (0.24925)\end{tabular}              & \cellcolor[HTML]{FCAA78}\begin{tabular}[c]{@{}r@{}}8\\      (0.0343)\end{tabular}      & \cellcolor[HTML]{FCAA78}\begin{tabular}[c]{@{}r@{}}8\\      (0.72351)\end{tabular}       & \cellcolor[HTML]{CBDC81}4                                                  \\ \hline
\textbf{CTGAN\_MICE}           & \cellcolor[HTML]{DFE282}\begin{tabular}[c]{@{}r@{}}5\\      (12.40901)\end{tabular}   & \cellcolor[HTML]{DFE282}\begin{tabular}[c]{@{}r@{}}5\\      (10.88513)\end{tabular}   & \cellcolor[HTML]{FDB87B}\begin{tabular}[c]{@{}r@{}}8\\      (3.6)\end{tabular}                         & \cellcolor[HTML]{FED280}\begin{tabular}[c]{@{}r@{}}7\\      (0.4)\end{tabular}                           & \cellcolor[HTML]{DFE282}\begin{tabular}[c]{@{}r@{}}5\\      (0.35058)\end{tabular}               & \cellcolor[HTML]{FED280}\begin{tabular}[c]{@{}r@{}}7\\      (0.25055)\end{tabular}  & \cellcolor[HTML]{FFEB84}\begin{tabular}[c]{@{}r@{}}6\\      (0.03372)\end{tabular}     & \cellcolor[HTML]{C0D980}\begin{tabular}[c]{@{}r@{}}4\\      (1.14013)\end{tabular}      & \cellcolor[HTML]{DFE282}\begin{tabular}[c]{@{}r@{}}5\\      (0.00735)\end{tabular}       & \cellcolor[HTML]{FB9D75}\begin{tabular}[c]{@{}r@{}}9\\      (0.69464)\end{tabular}        & \cellcolor[HTML]{FA8370}\begin{tabular}[c]{@{}r@{}}10\\      (0.02037)\end{tabular}                & \cellcolor[HTML]{FB9D75}\begin{tabular}[c]{@{}r@{}}9\\      (0.14895)\end{tabular}                  & \cellcolor[HTML]{FED280}\begin{tabular}[c]{@{}r@{}}7\\      (0.00461)\end{tabular}             & \cellcolor[HTML]{FFEB84}\begin{tabular}[c]{@{}r@{}}6\\      (0.134)\end{tabular}                & \cellcolor[HTML]{FCAA78}\begin{tabular}[c]{@{}r@{}}8\\      (0.0343)\end{tabular}      & \cellcolor[HTML]{FCAA78}\begin{tabular}[c]{@{}r@{}}8\\      (0.72351)\end{tabular}       & \cellcolor[HTML]{FEC87E}7                                                  \\ \hline
\textbf{CTGAN\_WITH\_AE}       & \cellcolor[HTML]{FDB87B}\begin{tabular}[c]{@{}r@{}}8\\      (23.12505)\end{tabular}   & \cellcolor[HTML]{FED280}\begin{tabular}[c]{@{}r@{}}7\\      (25.73443)\end{tabular}   & \cellcolor[HTML]{DFE282}\begin{tabular}[c]{@{}r@{}}5\\      (3.8)\end{tabular}                         & \cellcolor[HTML]{FA8370}\begin{tabular}[c]{@{}r@{}}10\\      (1)\end{tabular}                            & \cellcolor[HTML]{FB9D75}\begin{tabular}[c]{@{}r@{}}9\\      (0.36133)\end{tabular}               & \cellcolor[HTML]{FB9D75}\begin{tabular}[c]{@{}r@{}}9\\      (0.25122)\end{tabular}  & \cellcolor[HTML]{FA8370}\begin{tabular}[c]{@{}r@{}}10\\      (0.03789)\end{tabular}    & \cellcolor[HTML]{FA8370}\begin{tabular}[c]{@{}r@{}}10\\      (1.3217)\end{tabular}      & \cellcolor[HTML]{FFEB84}\begin{tabular}[c]{@{}r@{}}6\\      (0.01181)\end{tabular}       & \cellcolor[HTML]{FFEB84}\begin{tabular}[c]{@{}r@{}}6\\      (0.4029)\end{tabular}         & \cellcolor[HTML]{FED280}\begin{tabular}[c]{@{}r@{}}7\\      (0.01577)\end{tabular}                 & \cellcolor[HTML]{FED280}\begin{tabular}[c]{@{}r@{}}7\\      (0.08379)\end{tabular}                  & \cellcolor[HTML]{FA8370}\begin{tabular}[c]{@{}r@{}}10\\      (0.0257)\end{tabular}             & \cellcolor[HTML]{FA8370}\begin{tabular}[c]{@{}r@{}}10\\      (0.37753)\end{tabular}             & \cellcolor[HTML]{F8696B}\begin{tabular}[c]{@{}r@{}}10\\      (0.05094)\end{tabular}    & \cellcolor[HTML]{F8696B}\begin{tabular}[c]{@{}r@{}}10\\      (1.20827)\end{tabular}      & \cellcolor[HTML]{FEC87E}7                                                  \\ \hline
\textbf{CTGAN\_WITH\_AE\_MICE} & \cellcolor[HTML]{C0D980}\begin{tabular}[c]{@{}r@{}}4\\      (7.47082)\end{tabular}    & \cellcolor[HTML]{C0D980}\begin{tabular}[c]{@{}r@{}}4\\      (9.42164)\end{tabular}    & \cellcolor[HTML]{C0D980}\begin{tabular}[c]{@{}r@{}}4\\      (4)\end{tabular}                           & \cellcolor[HTML]{FDB87B}\begin{tabular}[c]{@{}r@{}}8\\      (0.6)\end{tabular}                           & \cellcolor[HTML]{C0D980}\begin{tabular}[c]{@{}r@{}}4\\      (0.34927)\end{tabular}               & \cellcolor[HTML]{C0D980}\begin{tabular}[c]{@{}r@{}}4\\      (0.25003)\end{tabular}  & \cellcolor[HTML]{FB9D75}\begin{tabular}[c]{@{}r@{}}9\\      (0.03667)\end{tabular}     & \cellcolor[HTML]{FB9D75}\begin{tabular}[c]{@{}r@{}}9\\      (1.26214)\end{tabular}      & \cellcolor[HTML]{C0D980}\begin{tabular}[c]{@{}r@{}}4\\      (0.00681)\end{tabular}       & \cellcolor[HTML]{C0D980}\begin{tabular}[c]{@{}r@{}}4\\      (0.33708)\end{tabular}        & \cellcolor[HTML]{FED280}\begin{tabular}[c]{@{}r@{}}7\\      (0.01577)\end{tabular}                 & \cellcolor[HTML]{FED280}\begin{tabular}[c]{@{}r@{}}7\\      (0.08379)\end{tabular}                  & \cellcolor[HTML]{FDB87B}\begin{tabular}[c]{@{}r@{}}8\\      (0.01116)\end{tabular}             & \cellcolor[HTML]{FED280}\begin{tabular}[c]{@{}r@{}}7\\      (0.18046)\end{tabular}              & \cellcolor[HTML]{F8696B}\begin{tabular}[c]{@{}r@{}}10\\      (0.05094)\end{tabular}    & \cellcolor[HTML]{F8696B}\begin{tabular}[c]{@{}r@{}}10\\      (1.20827)\end{tabular}      & \cellcolor[HTML]{FB9975}9                                                  \\ \hline
\textbf{VAE\_JAMOTTON}         & \cellcolor[HTML]{FFEB84}\begin{tabular}[c]{@{}r@{}}6\\      (17.43056)\end{tabular}   & \cellcolor[HTML]{FFEB84}\begin{tabular}[c]{@{}r@{}}6\\      (21.82393)\end{tabular}   & \cellcolor[HTML]{FDB87B}\begin{tabular}[c]{@{}r@{}}8\\      (3.6)\end{tabular}                         & \cellcolor[HTML]{C0D980}\begin{tabular}[c]{@{}r@{}}4\\      (0.2)\end{tabular}                           & \cellcolor[HTML]{FFEB84}\begin{tabular}[c]{@{}r@{}}6\\      (0.35125)\end{tabular}               & \cellcolor[HTML]{DFE282}\begin{tabular}[c]{@{}r@{}}5\\      (0.25028)\end{tabular}  & \cellcolor[HTML]{FED280}\begin{tabular}[c]{@{}r@{}}7\\      (0.03397)\end{tabular}     & \cellcolor[HTML]{DFE282}\begin{tabular}[c]{@{}r@{}}5\\      (1.17735)\end{tabular}      & \cellcolor[HTML]{FED280}\begin{tabular}[c]{@{}r@{}}7\\      (0.01359)\end{tabular}       & \cellcolor[HTML]{FDB87B}\begin{tabular}[c]{@{}r@{}}8\\      (0.64918)\end{tabular}        & \cellcolor[HTML]{FFEB84}\begin{tabular}[c]{@{}r@{}}6\\      (0.00531)\end{tabular}                 & \cellcolor[HTML]{FFEB84}\begin{tabular}[c]{@{}r@{}}6\\      (0.06159)\end{tabular}                  & \cellcolor[HTML]{FFEB84}\begin{tabular}[c]{@{}r@{}}6\\      (0.00405)\end{tabular}             & \cellcolor[HTML]{FDB87B}\begin{tabular}[c]{@{}r@{}}8\\      (0.21452)\end{tabular}              & \cellcolor[HTML]{DFE282}\begin{tabular}[c]{@{}r@{}}5\\      (0.01881)\end{tabular}     & \cellcolor[HTML]{C0D980}\begin{tabular}[c]{@{}r@{}}4\\      (0.31386)\end{tabular}       & \cellcolor[HTML]{FA8170}10                                                 \\ \hline
\textbf{COTE\_MC\_WGAN\_GP}    & \cellcolor[HTML]{F8696B}\begin{tabular}[c]{@{}r@{}}11\\      (100.37982)\end{tabular} & \cellcolor[HTML]{F8696B}\begin{tabular}[c]{@{}r@{}}11\\      (102.98341)\end{tabular} & \cellcolor[HTML]{A1D07E}\begin{tabular}[c]{@{}r@{}}3\\      (4.8)\end{tabular}                         & \cellcolor[HTML]{FA8370}\begin{tabular}[c]{@{}r@{}}10\\      (1)\end{tabular}                            & \cellcolor[HTML]{F8696B}\begin{tabular}[c]{@{}r@{}}11\\      (0.39574)\end{tabular}              & \cellcolor[HTML]{F8696B}\begin{tabular}[c]{@{}r@{}}11\\      (0.26078)\end{tabular} & \cellcolor[HTML]{F8696B}\begin{tabular}[c]{@{}r@{}}11\\      (0.03921)\end{tabular}    & \cellcolor[HTML]{F8696B}\begin{tabular}[c]{@{}r@{}}11\\      (5.28639)\end{tabular}     & \cellcolor[HTML]{F8696B}\begin{tabular}[c]{@{}r@{}}11\\      (0.11859)\end{tabular}      & \cellcolor[HTML]{F8696B}\begin{tabular}[c]{@{}r@{}}11\\      (5.02162)\end{tabular}       & \cellcolor[HTML]{FB9D75}\begin{tabular}[c]{@{}r@{}}9\\      (0.01807)\end{tabular}                 & \cellcolor[HTML]{F8696B}\begin{tabular}[c]{@{}r@{}}11\\      (0.27211)\end{tabular}                 & \cellcolor[HTML]{F8696B}\begin{tabular}[c]{@{}r@{}}11\\      (0.05335)\end{tabular}            & \cellcolor[HTML]{F8696B}\begin{tabular}[c]{@{}r@{}}11\\      (31.18191)\end{tabular}            & \cellcolor[HTML]{C0D980}\begin{tabular}[c]{@{}r@{}}4\\      (0.01679)\end{tabular}     & \cellcolor[HTML]{DFE282}\begin{tabular}[c]{@{}r@{}}5\\      (0.37443)\end{tabular}       & \cellcolor[HTML]{F8696B}11                                                 \\ \hline
\end{tabular}%
}
\caption{Summary of evaluation results across all tested algorithms when the variable dependence is based on linear formula \eqref{eval:mccase_linear_formula}.}
\label{results:mccase_linear_table}
\end{table}

The results of the smaller case study follow closely on what we can see in the main case study. Again, it is important to note that the ordering of the methods within the tables is done based on the groupings of similar methods, as it was done in the main case study.

Based on these results, MICE-based methods perform best overall on both model and data metrics. In Table \ref{results:mccase_linear_table}, RMSE for \texttt{MICE\_FULL\_SYN} even outperforms the model trained on original training data when evaluated on the hold-out test dataset, although the difference in the metric values is in the fourth decimal. From Table \ref{results:mccase_linear_table}, we can also see that the distributions of the $M_1$ and $M_2$ metrics are much tighter than those of the same metrics in the main study's linear results (Table \ref{results:linear_table}). By far, for both linear and interaction formulas, \texttt{COTE\_MC\_WGAN\_GP} performed the worst, achieving most of the worst ranks for model and data metrics. Lastly, we can see that, although the autoencoders improved the generation of categorical variables, their dependence structure is the worst in both linear and interaction cases. This possibly indicates a limitation in autoencoder usage (at least in pre-processing capacity) for smaller datasets.
\begin{table}[!ht]
\centering
\resizebox{\textwidth}{!}{%
\begin{tabular}{|l|r|r|r|r|r|r|r|r|r|r|r|r|r|r|r|r|r|l|}
\hline
\textbf{Generation   method}   & \multicolumn{1}{l|}{\textbf{M1}}                                                     & \multicolumn{1}{l|}{\textbf{M2}}                                                     & \multicolumn{1}{l|}{\textbf{\begin{tabular}[c]{@{}l@{}}Correct \\ variables \\ selected\end{tabular}}} & \multicolumn{1}{l|}{\textbf{\begin{tabular}[c]{@{}l@{}}Incorrect \\ variables \\ selected\end{tabular}}} & \multicolumn{1}{l|}{\textbf{\begin{tabular}[c]{@{}l@{}}Missing \\ main effect \\ variable\end{tabular}}}                            & \multicolumn{1}{l|}{\textbf{\begin{tabular}[c]{@{}l@{}}Fit \\ Poisson \\ deviance\end{tabular}}} & \multicolumn{1}{l|}{\textbf{\begin{tabular}[c]{@{}l@{}}Fit\\ RMSE\end{tabular}}}    & \multicolumn{1}{l|}{\textbf{\begin{tabular}[c]{@{}l@{}}Pair-wise \\ MAE\end{tabular}}} & \multicolumn{1}{l|}{\textbf{\begin{tabular}[c]{@{}l@{}}Pair-Wise \\ MAPE\end{tabular}}} & \multicolumn{1}{l|}{\textbf{\begin{tabular}[c]{@{}l@{}}Correlation \\ MAE\end{tabular}}} & \multicolumn{1}{l|}{\textbf{\begin{tabular}[c]{@{}l@{}}Correlation \\ MAPE\end{tabular}}} & \multicolumn{1}{l|}{\textbf{\begin{tabular}[c]{@{}l@{}}Categorical \\ column \\ MAE\end{tabular}}} & \multicolumn{1}{l|}{\textbf{\begin{tabular}[c]{@{}l@{}}Categorical \\ column \\ MAPE\end{tabular}}} & \multicolumn{1}{l|}{\textbf{\begin{tabular}[c]{@{}l@{}}Numeric \\ column \\ MAE\end{tabular}}} & \multicolumn{1}{l|}{\textbf{\begin{tabular}[c]{@{}l@{}}Numeric \\ column \\ MAPE\end{tabular}}} & \multicolumn{1}{l|}{\textbf{\begin{tabular}[c]{@{}l@{}}Cramér's V\\ MAE\end{tabular}}} & \multicolumn{1}{l|}{\textbf{\begin{tabular}[c]{@{}l@{}}Cramér's V \\ MAPE\end{tabular}}} & \textbf{\begin{tabular}[c]{@{}l@{}}Ease \\ of use \\ ranking\end{tabular}} \\ \hline
\textbf{training}              & \cellcolor[HTML]{63BE7B}\begin{tabular}[c]{@{}r@{}}1\\      (0.97424)\end{tabular}   & \cellcolor[HTML]{63BE7B}\begin{tabular}[c]{@{}r@{}}1\\      (1)\end{tabular}         & \cellcolor[HTML]{FDB87B}\begin{tabular}[c]{@{}r@{}}8\\      (6)\end{tabular}                           & \cellcolor[HTML]{63BE7B}\begin{tabular}[c]{@{}r@{}}1\\      (0)\end{tabular}                             & \multicolumn{1}{l|}{\cellcolor[HTML]{FDB87B}{\color[HTML]{333333} \textbf{\begin{tabular}[c]{@{}l@{}}8\\      (1)\end{tabular}}}}   & \cellcolor[HTML]{63BE7B}\begin{tabular}[c]{@{}r@{}}1\\      (0.34416)\end{tabular}               & \cellcolor[HTML]{63BE7B}\begin{tabular}[c]{@{}r@{}}1\\      (0.25078)\end{tabular}  & \cellcolor[HTML]{82C77C}\begin{tabular}[c]{@{}r@{}}2\\      (0.03302)\end{tabular}     & \cellcolor[HTML]{63BE7B}\begin{tabular}[c]{@{}r@{}}1\\      (1.06924)\end{tabular}      & \cellcolor[HTML]{63BE7B}\begin{tabular}[c]{@{}r@{}}1\\      (0)\end{tabular}             & \cellcolor[HTML]{63BE7B}\begin{tabular}[c]{@{}r@{}}1\\      (0)\end{tabular}              & \cellcolor[HTML]{63BE7B}\begin{tabular}[c]{@{}r@{}}1\\      (0)\end{tabular}                       & \cellcolor[HTML]{63BE7B}\begin{tabular}[c]{@{}r@{}}1\\      (0)\end{tabular}                        & \cellcolor[HTML]{63BE7B}\begin{tabular}[c]{@{}r@{}}1\\      (0)\end{tabular}                   & \cellcolor[HTML]{63BE7B}\begin{tabular}[c]{@{}r@{}}1\\      (0)\end{tabular}                    & \cellcolor[HTML]{63BE7B}\begin{tabular}[c]{@{}r@{}}1\\      (0)\end{tabular}           & \cellcolor[HTML]{63BE7B}\begin{tabular}[c]{@{}r@{}}1\\      (0)\end{tabular}             & \cellcolor[HTML]{63BE7B}1                                                  \\ \hline
\textbf{MICE\_PART\_SYN}       & \cellcolor[HTML]{82C77C}\begin{tabular}[c]{@{}r@{}}2\\      (3.38923)\end{tabular}   & \cellcolor[HTML]{82C77C}\begin{tabular}[c]{@{}r@{}}2\\      (4.33722)\end{tabular}   & \cellcolor[HTML]{DFE282}\begin{tabular}[c]{@{}r@{}}5\\      (6.2)\end{tabular}                         & \cellcolor[HTML]{DFE282}\begin{tabular}[c]{@{}r@{}}5\\      (0.4)\end{tabular}                           & \multicolumn{1}{l|}{\cellcolor[HTML]{FFEB84}{\color[HTML]{333333} \textbf{\begin{tabular}[c]{@{}l@{}}6\\      (0.6)\end{tabular}}}} & \cellcolor[HTML]{82C77C}\begin{tabular}[c]{@{}r@{}}2\\      (0.34702)\end{tabular}               & \cellcolor[HTML]{82C77C}\begin{tabular}[c]{@{}r@{}}2\\      (0.25132)\end{tabular}  & \cellcolor[HTML]{C0D980}\begin{tabular}[c]{@{}r@{}}4\\      (0.03317)\end{tabular}     & \cellcolor[HTML]{82C77C}\begin{tabular}[c]{@{}r@{}}2\\      (1.12737)\end{tabular}      & \cellcolor[HTML]{82C77C}\begin{tabular}[c]{@{}r@{}}2\\      (0.00396)\end{tabular}       & \cellcolor[HTML]{82C77C}\begin{tabular}[c]{@{}r@{}}2\\      (0.13892)\end{tabular}        & \cellcolor[HTML]{A1D07E}\begin{tabular}[c]{@{}r@{}}3\\      (0.00128)\end{tabular}                 & \cellcolor[HTML]{C0D980}\begin{tabular}[c]{@{}r@{}}4\\      (0.01079)\end{tabular}                  & \cellcolor[HTML]{A1D07E}\begin{tabular}[c]{@{}r@{}}3\\      (0.00086)\end{tabular}             & \cellcolor[HTML]{C0D980}\begin{tabular}[c]{@{}r@{}}4\\      (0.06943)\end{tabular}              & \cellcolor[HTML]{82C77C}\begin{tabular}[c]{@{}r@{}}2\\      (0.00531)\end{tabular}     & \cellcolor[HTML]{82C77C}\begin{tabular}[c]{@{}r@{}}2\\      (0.09782)\end{tabular}       & \cellcolor[HTML]{85C87D}2                                                  \\ \hline
\textbf{MICE\_FULL\_SYN}       & \cellcolor[HTML]{A1D07E}\begin{tabular}[c]{@{}r@{}}3\\      (4.85001)\end{tabular}   & \cellcolor[HTML]{A1D07E}\begin{tabular}[c]{@{}r@{}}3\\      (5.87237)\end{tabular}   & \cellcolor[HTML]{82C77C}\begin{tabular}[c]{@{}r@{}}2\\      (6.8)\end{tabular}                         & \cellcolor[HTML]{82C77C}\begin{tabular}[c]{@{}r@{}}2\\      (0.2)\end{tabular}                           & \cellcolor[HTML]{A1D07E}\begin{tabular}[c]{@{}r@{}}3\\      (0.2)\end{tabular}                                                      & \cellcolor[HTML]{C0D980}\begin{tabular}[c]{@{}r@{}}4\\      (0.3483)\end{tabular}                & \cellcolor[HTML]{C0D980}\begin{tabular}[c]{@{}r@{}}4\\      (0.25153)\end{tabular}  & \cellcolor[HTML]{A1D07E}\begin{tabular}[c]{@{}r@{}}3\\      (0.03315)\end{tabular}     & \cellcolor[HTML]{A1D07E}\begin{tabular}[c]{@{}r@{}}3\\      (1.1275)\end{tabular}       & \cellcolor[HTML]{C0D980}\begin{tabular}[c]{@{}r@{}}4\\      (0.00478)\end{tabular}       & \cellcolor[HTML]{A1D07E}\begin{tabular}[c]{@{}r@{}}3\\      (0.14765)\end{tabular}        & \cellcolor[HTML]{C0D980}\begin{tabular}[c]{@{}r@{}}4\\      (0.00131)\end{tabular}                 & \cellcolor[HTML]{A1D07E}\begin{tabular}[c]{@{}r@{}}3\\      (0.01069)\end{tabular}                  & \cellcolor[HTML]{C0D980}\begin{tabular}[c]{@{}r@{}}4\\      (0.00087)\end{tabular}             & \cellcolor[HTML]{82C77C}\begin{tabular}[c]{@{}r@{}}2\\      (0.05977)\end{tabular}              & \cellcolor[HTML]{A1D07E}\begin{tabular}[c]{@{}r@{}}3\\      (0.00594)\end{tabular}     & \cellcolor[HTML]{A1D07E}\begin{tabular}[c]{@{}r@{}}3\\      (0.10978)\end{tabular}       & \multicolumn{1}{l|}{\cellcolor[HTML]{C0D980}4}                             \\ \hline
\textbf{MICE\_VV}              & \cellcolor[HTML]{FDB87B}\begin{tabular}[c]{@{}r@{}}8\\      (16.54448)\end{tabular}  & \cellcolor[HTML]{FB9D75}\begin{tabular}[c]{@{}r@{}}9\\      (19.83362)\end{tabular}  & \cellcolor[HTML]{FB9D75}\begin{tabular}[c]{@{}r@{}}9\\      (4.2)\end{tabular}                         & \cellcolor[HTML]{DFE282}\begin{tabular}[c]{@{}r@{}}5\\      (0.4)\end{tabular}                           & \cellcolor[HTML]{F8696B}\begin{tabular}[c]{@{}r@{}}11\\      (2.8)\end{tabular}                                                     & \cellcolor[HTML]{FB9D75}\begin{tabular}[c]{@{}r@{}}9\\      (0.35908)\end{tabular}               & \cellcolor[HTML]{FB9D75}\begin{tabular}[c]{@{}r@{}}9\\      (0.25309)\end{tabular}  & \cellcolor[HTML]{63BE7B}\begin{tabular}[c]{@{}r@{}}1\\      (0.03302)\end{tabular}     & \cellcolor[HTML]{DFE282}\begin{tabular}[c]{@{}r@{}}5\\      (1.18562)\end{tabular}      & \cellcolor[HTML]{FED280}\begin{tabular}[c]{@{}r@{}}7\\      (0.01766)\end{tabular}       & \cellcolor[HTML]{FFEB84}\begin{tabular}[c]{@{}r@{}}6\\      (0.32904)\end{tabular}        & \cellcolor[HTML]{DFE282}\begin{tabular}[c]{@{}r@{}}5\\      (0.00509)\end{tabular}                 & \cellcolor[HTML]{DFE282}\begin{tabular}[c]{@{}r@{}}5\\      (0.03342)\end{tabular}                  & \cellcolor[HTML]{DFE282}\begin{tabular}[c]{@{}r@{}}5\\      (0.00161)\end{tabular}             & \cellcolor[HTML]{DFE282}\begin{tabular}[c]{@{}r@{}}5\\      (0.08691)\end{tabular}              & \cellcolor[HTML]{FA8A72}\begin{tabular}[c]{@{}r@{}}9\\      (0.02559)\end{tabular}     & \cellcolor[HTML]{FECB7E}\begin{tabular}[c]{@{}r@{}}7\\      (0.43685)\end{tabular}       & \cellcolor[HTML]{85C87D}2                                                  \\ \hline
\textbf{MICE\_TABULATOR}       & \cellcolor[HTML]{FFEB84}\begin{tabular}[c]{@{}r@{}}6\\      (12.26688)\end{tabular}  & \cellcolor[HTML]{FFEB84}\begin{tabular}[c]{@{}r@{}}6\\      (15.16141)\end{tabular}  & \cellcolor[HTML]{F8696B}\begin{tabular}[c]{@{}r@{}}11\\      (3.4)\end{tabular}                        & \cellcolor[HTML]{82C77C}\begin{tabular}[c]{@{}r@{}}2\\      (0.2)\end{tabular}                           & \cellcolor[HTML]{FB9D75}\begin{tabular}[c]{@{}r@{}}9\\      (1.6)\end{tabular}                                                      & \cellcolor[HTML]{FDB87B}\begin{tabular}[c]{@{}r@{}}8\\      (0.35529)\end{tabular}               & \cellcolor[HTML]{FDB87B}\begin{tabular}[c]{@{}r@{}}8\\      (0.25256)\end{tabular}  & \cellcolor[HTML]{DFE282}\begin{tabular}[c]{@{}r@{}}5\\      (0.03342)\end{tabular}     & \cellcolor[HTML]{C0D980}\begin{tabular}[c]{@{}r@{}}4\\      (1.18158)\end{tabular}      & \cellcolor[HTML]{FFEB84}\begin{tabular}[c]{@{}r@{}}6\\      (0.01457)\end{tabular}       & \cellcolor[HTML]{DFE282}\begin{tabular}[c]{@{}r@{}}5\\      (0.28725)\end{tabular}        & \cellcolor[HTML]{82C77C}\begin{tabular}[c]{@{}r@{}}2\\      (0.00106)\end{tabular}                 & \cellcolor[HTML]{82C77C}\begin{tabular}[c]{@{}r@{}}2\\      (0.00869)\end{tabular}                  & \cellcolor[HTML]{82C77C}\begin{tabular}[c]{@{}r@{}}2\\      (0.00075)\end{tabular}             & \cellcolor[HTML]{A1D07E}\begin{tabular}[c]{@{}r@{}}3\\      (0.06922)\end{tabular}              & \cellcolor[HTML]{FFEB84}\begin{tabular}[c]{@{}r@{}}6\\      (0.02384)\end{tabular}     & \cellcolor[HTML]{FFEB84}\begin{tabular}[c]{@{}r@{}}6\\      (0.41736)\end{tabular}       & \cellcolor[HTML]{CBDC81}4                                                  \\ \hline
\textbf{CTGAN}                 & \cellcolor[HTML]{FA8370}\begin{tabular}[c]{@{}r@{}}10\\      (30.29716)\end{tabular} & \cellcolor[HTML]{FA8370}\begin{tabular}[c]{@{}r@{}}10\\      (27.72298)\end{tabular} & \cellcolor[HTML]{A1D07E}\begin{tabular}[c]{@{}r@{}}3\\      (6.6)\end{tabular}                         & \cellcolor[HTML]{FED280}\begin{tabular}[c]{@{}r@{}}7\\      (1)\end{tabular}                             & \cellcolor[HTML]{63BE7B}\begin{tabular}[c]{@{}r@{}}1\\      (0)\end{tabular}                                                        & \cellcolor[HTML]{FA8370}\begin{tabular}[c]{@{}r@{}}10\\      (0.3639)\end{tabular}               & \cellcolor[HTML]{FA8370}\begin{tabular}[c]{@{}r@{}}10\\      (0.2531)\end{tabular}  & \cellcolor[HTML]{FA8370}\begin{tabular}[c]{@{}r@{}}10\\      (0.03729)\end{tabular}    & \cellcolor[HTML]{FB9D75}\begin{tabular}[c]{@{}r@{}}9\\      (1.29088)\end{tabular}      & \cellcolor[HTML]{FDB87B}\begin{tabular}[c]{@{}r@{}}8\\      (0.01779)\end{tabular}       & \cellcolor[HTML]{FA8370}\begin{tabular}[c]{@{}r@{}}10\\      (0.8713)\end{tabular}        & \cellcolor[HTML]{FB9D75}\begin{tabular}[c]{@{}r@{}}9\\      (0.02537)\end{tabular}                 & \cellcolor[HTML]{FB9D75}\begin{tabular}[c]{@{}r@{}}9\\      (0.16148)\end{tabular}                  & \cellcolor[HTML]{FB9D75}\begin{tabular}[c]{@{}r@{}}9\\      (0.01553)\end{tabular}             & \cellcolor[HTML]{FDB87B}\begin{tabular}[c]{@{}r@{}}8\\      (0.26626)\end{tabular}              & \cellcolor[HTML]{FECB7E}\begin{tabular}[c]{@{}r@{}}7\\      (0.0254)\end{tabular}      & \cellcolor[HTML]{FCAA78}\begin{tabular}[c]{@{}r@{}}8\\      (0.58729)\end{tabular}       & \cellcolor[HTML]{CBDC81}4                                                  \\ \hline
\textbf{CTGAN\_MICE}           & \cellcolor[HTML]{C0D980}\begin{tabular}[c]{@{}r@{}}4\\      (7.57989)\end{tabular}   & \cellcolor[HTML]{C0D980}\begin{tabular}[c]{@{}r@{}}4\\      (7.15009)\end{tabular}   & \cellcolor[HTML]{DFE282}\begin{tabular}[c]{@{}r@{}}5\\      (6.2)\end{tabular}                         & \cellcolor[HTML]{FED280}\begin{tabular}[c]{@{}r@{}}7\\      (1)\end{tabular}                             & \cellcolor[HTML]{A1D07E}\begin{tabular}[c]{@{}r@{}}3\\      (0.2)\end{tabular}                                                      & \cellcolor[HTML]{A1D07E}\begin{tabular}[c]{@{}r@{}}3\\      (0.34717)\end{tabular}               & \cellcolor[HTML]{A1D07E}\begin{tabular}[c]{@{}r@{}}3\\      (0.25138)\end{tabular}  & \cellcolor[HTML]{FED280}\begin{tabular}[c]{@{}r@{}}7\\      (0.03659)\end{tabular}     & \cellcolor[HTML]{FFEB84}\begin{tabular}[c]{@{}r@{}}6\\      (1.24565)\end{tabular}      & \cellcolor[HTML]{A1D07E}\begin{tabular}[c]{@{}r@{}}3\\      (0.00421)\end{tabular}       & \cellcolor[HTML]{C0D980}\begin{tabular}[c]{@{}r@{}}4\\      (0.16246)\end{tabular}        & \cellcolor[HTML]{FB9D75}\begin{tabular}[c]{@{}r@{}}9\\      (0.02537)\end{tabular}                 & \cellcolor[HTML]{FB9D75}\begin{tabular}[c]{@{}r@{}}9\\      (0.16148)\end{tabular}                  & \cellcolor[HTML]{FED280}\begin{tabular}[c]{@{}r@{}}7\\      (0.00634)\end{tabular}             & \cellcolor[HTML]{FFEB84}\begin{tabular}[c]{@{}r@{}}6\\      (0.15504)\end{tabular}              & \cellcolor[HTML]{FECB7E}\begin{tabular}[c]{@{}r@{}}7\\      (0.0254)\end{tabular}      & \cellcolor[HTML]{FCAA78}\begin{tabular}[c]{@{}r@{}}8\\      (0.58729)\end{tabular}       & \cellcolor[HTML]{FEC87E}7                                                  \\ \hline
\textbf{CTGAN\_WITH\_AE}       & \cellcolor[HTML]{FB9D75}\begin{tabular}[c]{@{}r@{}}9\\      (16.84151)\end{tabular}  & \cellcolor[HTML]{FDB87B}\begin{tabular}[c]{@{}r@{}}8\\      (18.41142)\end{tabular}  & \cellcolor[HTML]{C0D980}\begin{tabular}[c]{@{}r@{}}4\\      (6.4)\end{tabular}                         & \cellcolor[HTML]{FED280}\begin{tabular}[c]{@{}r@{}}7\\      (1)\end{tabular}                             & \cellcolor[HTML]{FFEB84}\begin{tabular}[c]{@{}r@{}}6\\      (0.6)\end{tabular}                                                      & \cellcolor[HTML]{FFEB84}\begin{tabular}[c]{@{}r@{}}6\\      (0.35433)\end{tabular}               & \cellcolor[HTML]{DFE282}\begin{tabular}[c]{@{}r@{}}5\\      (0.25234)\end{tabular}  & \cellcolor[HTML]{FB9D75}\begin{tabular}[c]{@{}r@{}}9\\      (0.03722)\end{tabular}     & \cellcolor[HTML]{FA8370}\begin{tabular}[c]{@{}r@{}}10\\      (1.2974)\end{tabular}      & \cellcolor[HTML]{FA8370}\begin{tabular}[c]{@{}r@{}}10\\      (0.02093)\end{tabular}      & \cellcolor[HTML]{FB9D75}\begin{tabular}[c]{@{}r@{}}9\\      (0.37777)\end{tabular}        & \cellcolor[HTML]{FED280}\begin{tabular}[c]{@{}r@{}}7\\      (0.01467)\end{tabular}                 & \cellcolor[HTML]{FED280}\begin{tabular}[c]{@{}r@{}}7\\      (0.08222)\end{tabular}                  & \cellcolor[HTML]{FA8370}\begin{tabular}[c]{@{}r@{}}10\\      (0.01796)\end{tabular}            & \cellcolor[HTML]{FB9D75}\begin{tabular}[c]{@{}r@{}}9\\      (0.28743)\end{tabular}              & \cellcolor[HTML]{F8696B}\begin{tabular}[c]{@{}r@{}}10\\      (0.04644)\end{tabular}    & \cellcolor[HTML]{F8696B}\begin{tabular}[c]{@{}r@{}}10\\      (1.06894)\end{tabular}      & \cellcolor[HTML]{FEC87E}7                                                  \\ \hline
\textbf{CTGAN\_WITH\_AE\_MICE} & \cellcolor[HTML]{FED280}\begin{tabular}[c]{@{}r@{}}7\\      (16.42044)\end{tabular}  & \cellcolor[HTML]{FED280}\begin{tabular}[c]{@{}r@{}}7\\      (16.80113)\end{tabular}  & \cellcolor[HTML]{63BE7B}\begin{tabular}[c]{@{}r@{}}1\\      (7)\end{tabular}                           & \cellcolor[HTML]{FED280}\begin{tabular}[c]{@{}r@{}}7\\      (1)\end{tabular}                             & \cellcolor[HTML]{A1D07E}\begin{tabular}[c]{@{}r@{}}3\\      (0.2)\end{tabular}                                                      & \cellcolor[HTML]{FED280}\begin{tabular}[c]{@{}r@{}}7\\      (0.35506)\end{tabular}               & \cellcolor[HTML]{FED280}\begin{tabular}[c]{@{}r@{}}7\\      (0.25246)\end{tabular}  & \cellcolor[HTML]{FDB87B}\begin{tabular}[c]{@{}r@{}}8\\      (0.03692)\end{tabular}     & \cellcolor[HTML]{FDB87B}\begin{tabular}[c]{@{}r@{}}8\\      (1.28876)\end{tabular}      & \cellcolor[HTML]{FB9D75}\begin{tabular}[c]{@{}r@{}}9\\      (0.02014)\end{tabular}       & \cellcolor[HTML]{FDB87B}\begin{tabular}[c]{@{}r@{}}8\\      (0.36025)\end{tabular}        & \cellcolor[HTML]{FED280}\begin{tabular}[c]{@{}r@{}}7\\      (0.01467)\end{tabular}                 & \cellcolor[HTML]{FED280}\begin{tabular}[c]{@{}r@{}}7\\      (0.08222)\end{tabular}                  & \cellcolor[HTML]{FDB87B}\begin{tabular}[c]{@{}r@{}}8\\      (0.01544)\end{tabular}             & \cellcolor[HTML]{FA8370}\begin{tabular}[c]{@{}r@{}}10\\      (0.29874)\end{tabular}             & \cellcolor[HTML]{F8696B}\begin{tabular}[c]{@{}r@{}}10\\      (0.04644)\end{tabular}    & \cellcolor[HTML]{F8696B}\begin{tabular}[c]{@{}r@{}}10\\      (1.06894)\end{tabular}      & \cellcolor[HTML]{FB9975}9                                                  \\ \hline
\textbf{VAE\_JAMOTTON}         & \cellcolor[HTML]{DFE282}\begin{tabular}[c]{@{}r@{}}5\\      (9.78169)\end{tabular}   & \cellcolor[HTML]{DFE282}\begin{tabular}[c]{@{}r@{}}5\\      (12.4332)\end{tabular}   & \cellcolor[HTML]{FA8370}\begin{tabular}[c]{@{}r@{}}10\\      (4)\end{tabular}                          & \cellcolor[HTML]{82C77C}\begin{tabular}[c]{@{}r@{}}2\\      (0.2)\end{tabular}                           & \cellcolor[HTML]{FA8370}\begin{tabular}[c]{@{}r@{}}10\\      (2.2)\end{tabular}                                                     & \cellcolor[HTML]{DFE282}\begin{tabular}[c]{@{}r@{}}5\\      (0.35431)\end{tabular}               & \cellcolor[HTML]{FFEB84}\begin{tabular}[c]{@{}r@{}}6\\      (0.25241)\end{tabular}  & \cellcolor[HTML]{FFEB84}\begin{tabular}[c]{@{}r@{}}6\\      (0.03421)\end{tabular}     & \cellcolor[HTML]{FED280}\begin{tabular}[c]{@{}r@{}}7\\      (1.25233)\end{tabular}      & \cellcolor[HTML]{DFE282}\begin{tabular}[c]{@{}r@{}}5\\      (0.01414)\end{tabular}       & \cellcolor[HTML]{FED280}\begin{tabular}[c]{@{}r@{}}7\\      (0.33957)\end{tabular}        & \cellcolor[HTML]{FFEB84}\begin{tabular}[c]{@{}r@{}}6\\      (0.00696)\end{tabular}                 & \cellcolor[HTML]{FFEB84}\begin{tabular}[c]{@{}r@{}}6\\      (0.07085)\end{tabular}                  & \cellcolor[HTML]{FFEB84}\begin{tabular}[c]{@{}r@{}}6\\      (0.00466)\end{tabular}             & \cellcolor[HTML]{FED280}\begin{tabular}[c]{@{}r@{}}7\\      (0.2304)\end{tabular}               & \cellcolor[HTML]{DFE282}\begin{tabular}[c]{@{}r@{}}5\\      (0.0179)\end{tabular}      & \cellcolor[HTML]{DFE282}\begin{tabular}[c]{@{}r@{}}5\\      (0.31)\end{tabular}          & \cellcolor[HTML]{FA8170}10                                                 \\ \hline
\textbf{COTE\_MC\_WGAN\_GP}    & \cellcolor[HTML]{F8696B}\begin{tabular}[c]{@{}r@{}}11\\      (43.21455)\end{tabular} & \cellcolor[HTML]{F8696B}\begin{tabular}[c]{@{}r@{}}11\\      (47.24069)\end{tabular} & \cellcolor[HTML]{DFE282}\begin{tabular}[c]{@{}r@{}}5\\      (6.2)\end{tabular}                         & \cellcolor[HTML]{FED280}\begin{tabular}[c]{@{}r@{}}7\\      (1)\end{tabular}                             & \cellcolor[HTML]{63BE7B}\begin{tabular}[c]{@{}r@{}}1\\      (0)\end{tabular}                                                        & \cellcolor[HTML]{F8696B}\begin{tabular}[c]{@{}r@{}}11\\      (0.38874)\end{tabular}              & \cellcolor[HTML]{F8696B}\begin{tabular}[c]{@{}r@{}}11\\      (0.26008)\end{tabular} & \cellcolor[HTML]{F8696B}\begin{tabular}[c]{@{}r@{}}11\\      (0.04278)\end{tabular}    & \cellcolor[HTML]{F8696B}\begin{tabular}[c]{@{}r@{}}11\\      (4.69448)\end{tabular}     & \cellcolor[HTML]{F8696B}\begin{tabular}[c]{@{}r@{}}11\\      (0.10096)\end{tabular}      & \cellcolor[HTML]{F8696B}\begin{tabular}[c]{@{}r@{}}11\\      (2.1778)\end{tabular}        & \cellcolor[HTML]{F8696B}\begin{tabular}[c]{@{}r@{}}11\\      (0.03805)\end{tabular}                & \cellcolor[HTML]{F8696B}\begin{tabular}[c]{@{}r@{}}11\\      (0.32199)\end{tabular}                 & \cellcolor[HTML]{F8696B}\begin{tabular}[c]{@{}r@{}}11\\      (0.06324)\end{tabular}            & \cellcolor[HTML]{F8696B}\begin{tabular}[c]{@{}r@{}}11\\      (14.18713)\end{tabular}            & \cellcolor[HTML]{C0D980}\begin{tabular}[c]{@{}r@{}}4\\      (0.01681)\end{tabular}     & \cellcolor[HTML]{C0D980}\begin{tabular}[c]{@{}r@{}}4\\      (0.23678)\end{tabular}       & \cellcolor[HTML]{F8696B}11                                                 \\ \hline
\end{tabular}%
}
\caption{Summary of evaluation results across all tested algorithms when the variable dependence is based on interaction formula \eqref{eval:mccase_int_formula}.}
\label{results:mccase_interaction_table}
\end{table}

\begin{figure}
    \centering
    \includegraphics[width=0.95\linewidth]{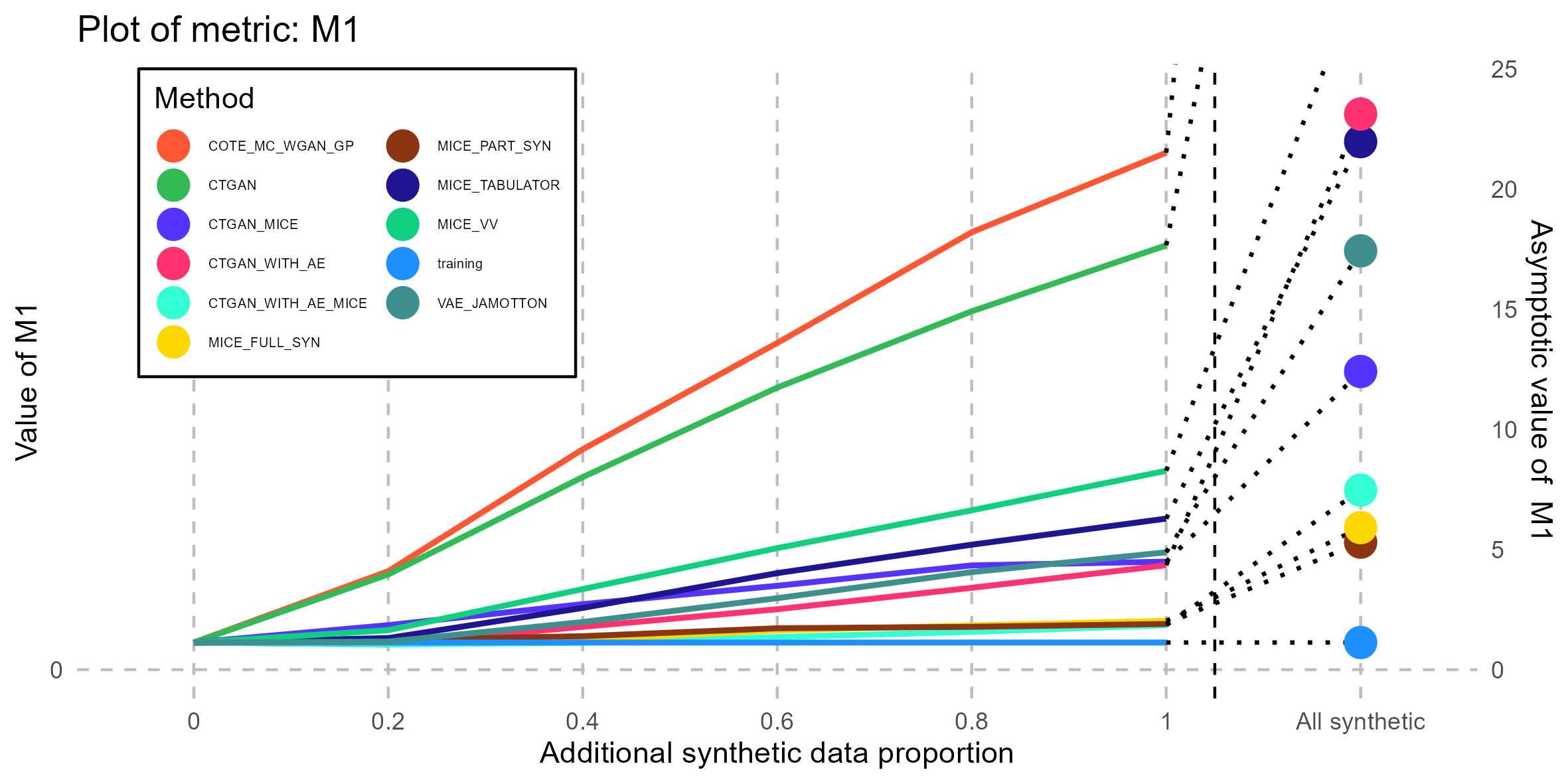}
    \caption{\small $M_1$ metric results as a function of synthetic data proportion. "All synthetic" denotes cases where only synthetic data is used and is treated as an "asymptotic" estimate of the relationship shown. The "All synthetic" metric values are scaled by 1:2 (right-side axis) compared to the other presented values (left-side axis). The dotted lines represent the corresponding asymptotic trends.}
    \label{fig:mccase_M1}
\end{figure}

As with previous metric results, we observe the same phenomenon with generic data augmentation as in the main case study. Namely, as the proportion of synthetic data increases, the $M_1$ metric increases. Again, we make an asymptotic assumption that if this proportion continues to increase (past 1), we would reach the same $M_1$ value as when using only synthetic data. Note that, due to scale differences between the $M_1$ values in this smaller case study and those in the main case study, more of the methods are visible. The most notable difference here is between \texttt{CTGAN\_WITH\_AE\_MICE} and \texttt{CTGAN\_MICE}, which both show a much smaller increase in asymptotic level than we see in the main case study.}

\end{document}